%% file: NC.tex
\documentclass[]{fairmeta}

\input{assets/paperstyle}

\usepackage{caption}

\usepackage{xcolor}
\definecolor{mcwarm}{RGB}{150, 90, 60}
\definecolor{metaaffblue}{HTML}{1877F2}
\definecolor{kaustrose}{HTML}{C97A8D}
\definecolor{metaaffblueedge}{HTML}{0E63CB}
\definecolor{metaaffbluelight}{HTML}{8BD0FF}
\definecolor{kaustroseedge}{HTML}{A55F6F}
\definecolor{kaustroselight}{HTML}{F0B9C5}
\definecolor{emeralddark}{HTML}{007A5E}
\definecolor{emeraldmid}{HTML}{00A878}
\definecolor{emeraldlight}{HTML}{7EF7C7}
\definecolor{emeraldedge}{HTML}{006B53}
\usepackage{amssymb}

\newcommand{\drawgradientbadge}[5]{%
  \tikz[baseline=-0.60ex, x=#5ex, y=#5ex]{
    \shade[
      shading=axis,
      shading angle=72,
      left color=#2,
      middle color=#3,
      right color=#4,
      draw=#2,
      line width=0.045
    ] (0,0) circle[radius=0.66];
    \fill[white, opacity=0.26] (-0.21,0.23) circle[radius=0.18];
    \fill[white, opacity=0.08] (0.14,-0.16) circle[radius=0.34];
    \node[text=white, font=\sffamily\bfseries, inner sep=0pt, scale=0.64] at (0,-0.015) {#1};
  }%
}
\DeclareRobustCommand{\MetaAffMark}{%
  \ifmmode
    \mathchoice
      {\drawgradientbadge{1}{metaaffblueedge}{metaaffblue}{metaaffbluelight}{0.95}}
      {\drawgradientbadge{1}{metaaffblueedge}{metaaffblue}{metaaffbluelight}{0.84}}
      {\drawgradientbadge{1}{metaaffblueedge}{metaaffblue}{metaaffbluelight}{0.75}}
      {\drawgradientbadge{1}{metaaffblueedge}{metaaffblue}{metaaffbluelight}{0.68}}%
  \else
    \drawgradientbadge{1}{metaaffblueedge}{metaaffblue}{metaaffbluelight}{0.95}%
  \fi
}
\DeclareRobustCommand{\KaustAffMark}{%
  \ifmmode
    \mathchoice
      {\drawgradientbadge{2}{kaustroseedge}{kaustrose}{kaustroselight}{0.95}}
      {\drawgradientbadge{2}{kaustroseedge}{kaustrose}{kaustroselight}{0.84}}
      {\drawgradientbadge{2}{kaustroseedge}{kaustrose}{kaustroselight}{0.75}}
      {\drawgradientbadge{2}{kaustroseedge}{kaustrose}{kaustroselight}{0.68}}%
  \else
    \drawgradientbadge{2}{kaustroseedge}{kaustrose}{kaustroselight}{0.95}%
  \fi
}
\newcommand{\drawemeraldplus}[1]{%
  \tikz[baseline=-0.58ex, x=#1ex, y=#1ex]{
    \shade[
      shading=axis,
      shading angle=18,
      left color=emeraldedge,
      middle color=emeraldmid,
      right color=emeraldlight,
      rounded corners=0.11,
      draw=emeraldedge,
      line width=0.03
    ] (-0.56,-0.12) rectangle (0.56,0.12);
    \shade[
      shading=axis,
      shading angle=108,
      left color=emeraldedge,
      middle color=emeraldmid,
      right color=emeraldlight,
      rounded corners=0.11,
      draw=emeraldedge,
      line width=0.03
    ] (-0.12,-0.56) rectangle (0.12,0.56);
  }%
}
\DeclareRobustCommand{\CoreContribMark}{%
  \ifmmode
    \mathchoice{\drawemeraldplus{0.82}}{\drawemeraldplus{0.72}}{\drawemeraldplus{0.62}}{\drawemeraldplus{0.56}}%
  \else
    \drawemeraldplus{0.82}%
  \fi
}

\newcommand{\cligenGeneralLogo}{\raisebox{-0.18em}{\includegraphics[height=1.0em]{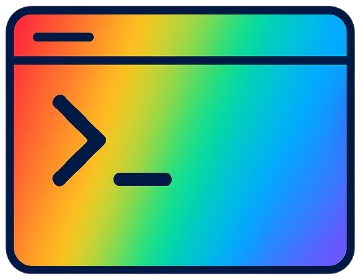}}}
\newcommand{\cligenCleanLogo}{\raisebox{-0.18em}{\includegraphics[height=1.0em]{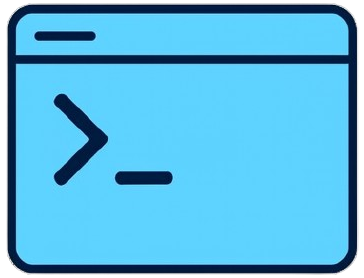}}}
\newcommand{\guiworldLogo}{\raisebox{-0.12em}{\includegraphics[height=1.0em]{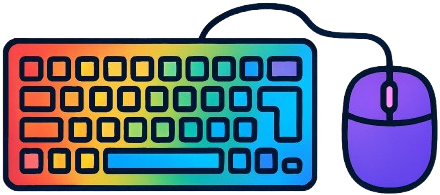}}}
\newcommand{\cligenGeneralData}{\cligenGeneralLogo{}\,CLIGen (General)}
\newcommand{\cligenCleanData}{\cligenCleanLogo{}\,CLIGen (Clean)}

\title{Neural Computers}

\author[\MetaAffMark,\KaustAffMark\CoreContribMark]{Mingchen Zhuge}
\author[\MetaAffMark\CoreContribMark]{Changsheng Zhao}
\author[\MetaAffMark,\KaustAffMark\CoreContribMark]{Haozhe Liu}
\author[\MetaAffMark\CoreContribMark]{Zijian Zhou}
\author[\MetaAffMark,\KaustAffMark\CoreContribMark]{Shuming Liu}
\author[\KaustAffMark]{Wenyi Wang}
\author[\MetaAffMark]{Ernie Chang}
\author[\MetaAffMark]{Gael Le Lan}
\author[\MetaAffMark,\KaustAffMark]{Junjie Fei}
\author[\MetaAffMark,\KaustAffMark]{Wenxuan Zhang}
\author[\KaustAffMark]{Yasheng Sun}
\author[\MetaAffMark]{Zhipeng Cai}
\author[\MetaAffMark]{Zechun Liu}
\author[\MetaAffMark]{Yunyang Xiong}
\author[\MetaAffMark]{Yining Yang}
\author[\MetaAffMark]{Yuandong Tian}
\author[\MetaAffMark]{Yangyang Shi}
\author[\MetaAffMark]{Vikas Chandra}
\author[\KaustAffMark]{Jürgen Schmidhuber}
\affiliation[\MetaAffMark]{Meta AI}
\affiliation[\KaustAffMark]{KAUST}
\contribution[\CoreContribMark]{Core Contributors}

\abstract{\input{section/abstract_js}}

\date{\today}
\correspondence{\email{mczhuge@gmail.com, cszhao@meta.com}}

\metadata[Blogpost]{\url{https://metauto.ai/neuralcomputer}}

\begin{document}

\maketitle

\vspace{0.15em}
{\captionsetup{skip=2pt,font=footnotesize}}
\begin{figure}[H]
  \centering
  \hypertarget{teaser}{}
  \includegraphics[width=0.92\linewidth]{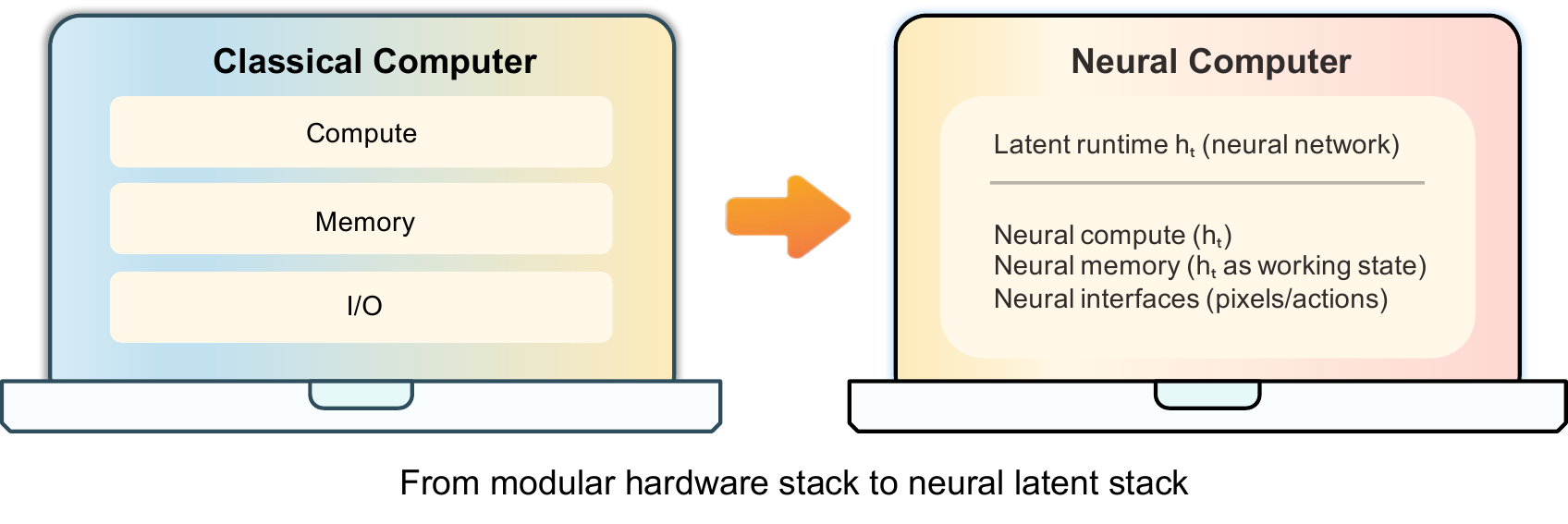}
\end{figure}\label{fig:nc_teaser}
\vspace{-1.2em}
\begingroup
\setcounter{tocdepth}{1}
\vspace{-0.7em}
\small\normalfont
\renewcommand{\baselinestretch}{0.8}\selectfont
\setlength{\parskip}{0pt}
\setlength{\parindent}{0pt}
\vspace{-0.7em}
\tableofcontents
\endgroup

\newpage
\section{Introduction}
\label{section:intro}
\input{section/intro}

\section{Preliminaries}
\label{section:prelim}
\input{section/preliminaries}

\section{Implementation of Neural Computers}
\label{section:implementation}

\begin{figure}[t]
  \centering
  \makebox[\linewidth][c]{\includegraphics[width=1\linewidth]{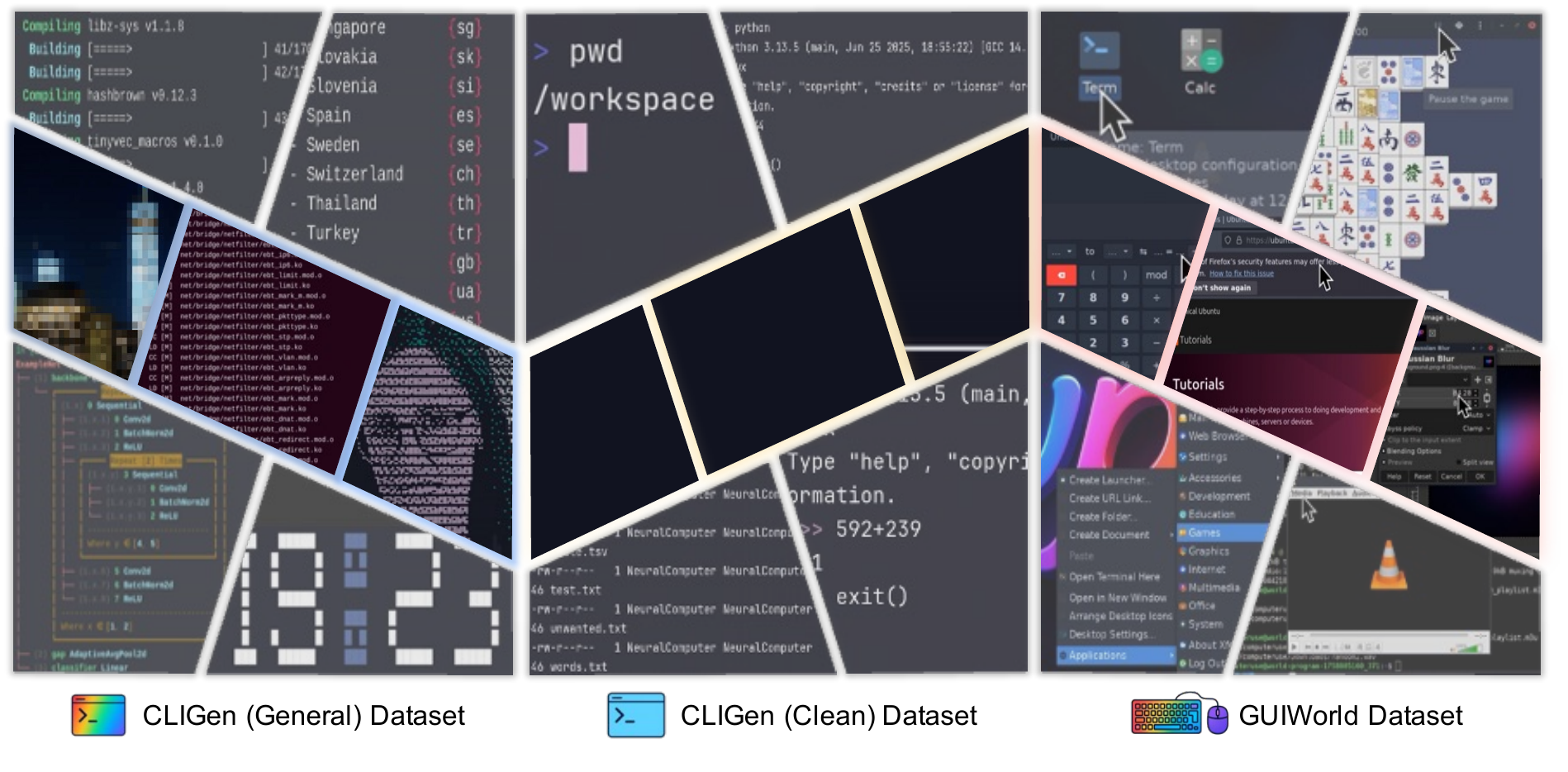}}
  \caption{\textbf{Data types used to learn NC behaviors.} Logos denote datasets: \cligenGeneralLogo{}\,CLIGen (General) replays public \texttt{asciinema} traces spanning diverse real-world terminal workflows. \cligenCleanLogo{}\,CLIGen (Clean) uses scripted \texttt{vhs} runs to capture deterministic terminal traces for controlled experiments. \guiworldLogo{}\,GUIWorld captures desktop RGB with synchronized mouse/keyboard traces to validate action-conditioned rendering and control on GUIs.}
  \label{fig:data-teaser}
\end{figure}

\input{section/model}

\subsection{\cligenGeneralLogo{}\,/\,\cligenCleanLogo{}\,\,The CLI Video Generators}
\label{section:impl-cligen}

\input{section/impl_cli}

\par\justifying

\newpage
\subsection{\guiworldLogo{}\,The GUI World Models}
\label{section:impl-guiworld}
\input{section/impl_gui}

\section{Position: Toward Completely Neural Computers}
\label{section:toward-cnc}
\input{section/toward_cnc}

\section{Conclusion}
\label{section:conclusion}
\input{section/conclusion}

\clearpage
\newpage
\bibliographystyle{assets/plainnat}
\bibliography{paper}

\clearpage
\newpage
\beginappendix

\phantomsection
\addcontentsline{toc}{section}{Appendix}
\addtocontents{toc}{\protect\setcounter{tocdepth}{0}}

\section{Explorations: Alternative Data Sources and Online Interaction}
\label{appendix:explorations}
\input{section/appendix_explorations}

\newpage
\section{Datasets: Collection and Evaluation Protocols}
\label{appendix:pipeline}
\input{section/appendix_pipeline}

\newpage
\section{CLIGen: CLI Trajectory Formats}
\label{appendix:cligen-samples}
\input{section/appendix_cligen}

\newpage
\section{GUIWorld: Action Representation, Temporal Alignment, and Conditioning}
\label{appendix:gui-actions}
\input{section/appendix_gui}

\newpage
\section{Additional Visualizations}
\label{appendix:vis}
\input{section/appendix_vis}

\end{document}

%% file: assets/paperstyle.tex
\usepackage{hyperref}
\usepackage{cleveref}

\usepackage{mathtools}

\makeatletter
\@ifundefined{abs}{}{}
\@ifundefined{norm}{}{}
\@ifundefined{ceil}{}{}
\@ifundefined{floor}{}{}
\@ifundefined{set}{}{}
\@ifundefined{paren}{}{}
\@ifundefined{bracket}{}{}
\@ifundefined{inner}{\DeclarePairedDelimiterX{\inner}[2]{\langle}{\rangle}{#1,\,#2}}{}
\makeatother

\usepackage{siunitx}
\usepackage[table]{xcolor}
\usepackage{tabularx}
\usepackage{array}
\usepackage{wrapfig}
\usepackage{ragged2e} 
\usepackage{pdflscape}
\usepackage{float}
\newcolumntype{J}[1]{>{\justifying\arraybackslash}p{#1}}
\newcolumntype{Z}{>{\justifying\arraybackslash}X}

\newcommand{\PSNR}{\textsc{PSNR}}
\newcommand{\SSIM}{\textsc{SSIM}}
\newcommand{\LPIPS}{\textsc{LPIPS}}
\newcommand{\FVD}{\textsc{FVD}}

\newcommand{\nccligen}{NC$_\text{CLIGen}$}
\newcommand{\ncguiworld}{NC$_\text{GUIWorld}$}

\usepackage{tikz}
\usetikzlibrary{shapes}
\usepackage{enumitem}
\setlist[itemize]{noitemsep, topsep=2pt, leftmargin=1.2em}
\setlist[enumerate]{noitemsep, topsep=2pt, leftmargin=1.4em}

\newcommand{\badgeedgecolor}[1]{%
  \ifcase\numexpr#1\relax
    metablue!88!black%
  \or red!78!black%
  \or blue!78!black%
  \or orange!90!black%
  \or teal!78!black%
  \or magenta!72!black%
  \or emeraldedge%
  \else
    metablue!88!black%
  \fi
}
\newcommand{\badgemidcolor}[1]{%
  \ifcase\numexpr#1\relax
    metablue%
  \or red!72%
  \or blue!74%
  \or orange!85!red%
  \or teal!72%
  \or magenta!68%
  \or emeraldmid%
  \else
    metablue%
  \fi
}
\newcommand{\badgelightcolor}[1]{%
  \ifcase\numexpr#1\relax
    metablue!28!white%
  \or red!16!white%
  \or blue!18!white%
  \or orange!22!white%
  \or teal!18!white%
  \or magenta!18!white%
  \or emeraldlight%
  \else
    metablue!28!white%
  \fi
}
\DeclareRobustCommand{\listbadge}[1]{%
  \tikz[baseline=-0.55ex]{%
    \shade[
      shading=axis,
      shading angle=72,
      left color=\badgeedgecolor{#1},
      middle color=\badgemidcolor{#1},
      right color=\badgelightcolor{#1},
      draw=\badgeedgecolor{#1},
      line width=0.04pt
    ] (0,0) circle[radius=0.88ex];
    \fill[white, opacity=0.22] (-0.24ex,0.26ex) circle[radius=0.24ex];
    \node[text=white, font=\sffamily\bfseries\scriptsize, inner sep=0pt] at (0,-0.01ex) {#1};
  }%
}
\newlist{badgeitems}{enumerate}{1}
\setlist[badgeitems]{
  label=\protect\listbadge{\arabic*},
  leftmargin=1.9em,
  labelsep=0.5em,
  itemsep=3pt,
  topsep=2pt,
  parsep=0pt,
  partopsep=0pt
}

\newcommand{\exptitle}[3][]{%
  \vspace{0.6em}%
  \noindent
  \ifstrempty{#1}{}{#1\hspace{0.5em}}%
  \textbf{Experiment~#2: #3}\par\vspace{0.25em}%
}

\newcommand{\prompttierline}[1]{%
  \noindent\textbf{#1}\hspace{0.5em}%
  \rule[0.55ex]{0.9\linewidth}{0.4pt}\par%
}

\crefname{equation}{Eq.}{Eqs.}
\Crefname{equation}{Equation}{Equations}
\crefname{figure}{Fig.}{Figs.}
\Crefname{figure}{Figure}{Figures}
\crefname{table}{Tab.}{Tabs.}
\Crefname{table}{Table}{Tables}
\crefname{section}{Section}{Sections}

\makeatletter
\@ifundefined{smallschemetable}{%
  \newenvironment{smallschemetable}[2]{
    \begin{table}[t]
      \centering
      \small
      \caption{#1}
      \label{#2}
  }{\end{table}}%
}{}%
\makeatother

\newtcolorbox{defbox}[2][]{
  breakable,
  enhanced,
  colback=apricotglaze!40!white,
  colframe=metablue!55,
  boxrule=0.6pt,
  arc=5pt,
  left=8pt, right=8pt, top=6pt, bottom=6pt,
  boxsep=2pt,
  before skip=8pt, after skip=8pt,
  interior style={shade,shading angle=315,left color=white,right color=apricotglaze!55!white},
  title={#2},
  fonttitle=\sffamily\bfseries\small,
  coltitle=metafg,
  before upper={\parindent=0pt\justifying},
  attach boxed title to top left={yshift=-3pt,xshift=8pt},
  boxed title style={
    frame empty,
    colback=metablue!10,
    boxsep=2pt,
    left=6pt,right=6pt,top=2pt,bottom=2pt,
    arc=5pt
  }
}

%% file: section/abstract_js.tex
We propose a new frontier: Neural Computers (NCs) that unify computation, memory, and I/O of traditional computers in a learned runtime state. 
Our long-term goal is the Completely~Neural~Computer (CNC): the mature, general-purpose realization of this emerging machine form, with stable execution, explicit reprogramming, and durable capability reuse. As an initial step, we study whether elementary NC primitives can be learned solely from collected I/O traces, without instrumented program state. Concretely, we instantiate~NCs~as~video models that roll out screen frames from instructions, pixels, and user actions (when available) in CLI~and~GUI settings. 
We show that NCs
can acquire elementary interface primitives, especially I/O alignment and short-horizon control, while routine reuse, controlled updates, and symbolic stability remain challenging. 
We outline a roadmap toward CNCs, to establish a new~computing paradigm beyond today’s agents and conventional~computers.

%% file: section/intro.tex
Can a neural network act as a traditional computer?  
The Neural Computer (NC) is a neural system that unifies computation, memory, and I/O in a learned runtime state.

Here we instantiate NCs as video models, currently perhaps
the most obvious substrate for NC prototypes, though we expect the long-term solution to require a fundamentally new neural architecture (\Cref{section:toward-cnc}).
Our implementation draws on several technical lines. World models~\citep{ha2018world} show that neural networks can internalize environment dynamics and support predictive imagination, while high-capacity video generators such as Veo~3.1~\citep{google_veo3_1_2025} and Sora~2~\citep{openai_sora2_2025} show that such learned dynamics can be rendered into coherent frame sequences.
Frontier interactive video models such as Genie~3~\citep{bruce2024genie} further extend this trajectory toward action-controllable generative environments. These lines provide practical machinery for implementing NC prototypes. 
NeuralOS~\citep{rivard2025neuralos}  generated next frames of graphical user interfaces (GUIs) from user actions.  
LLM-driven UI systems such as Imagine with Claude\footnote{\url{https://claude.ai/imagine/}} map natural-language inputs to structured interface updates. 
Yet these capabilities remain split across different systems objects: conventional computers execute explicit programs, agents act through external execution environments, and most world models render or predict environment dynamics, while executable state still resides outside the model. 
NCs are motivated by this gap: they are not a smarter layer on top of the existing stack, but a proposal to make the model itself the running computer.
The immediate question in this paper is whether reasonable internal runtime states can be learned directly from raw command line interfaces (CLIs) and GUIs without privileged access to the traditional computer's internal state. 

We study two interface-specific prototypes of this NC formulation (see~\Cref{section:prelim}).
\nccligen{} models CLI interaction from text (natural language or command lines) and an initial frame, while \ncguiworld{} models GUI desktop interaction from recent pixels and synchronized mouse/keyboard actions (\Cref{section:impl-cligen,section:impl-guiworld}).

\begin{tcolorbox}[
  colback=apricotglaze!40!white,
  colframe=metablue!40,
  interior style={shade,shading angle=315,left color=white,right color=apricotglaze!55!white},
  boxrule=0pt,
  borderline west={1pt}{0pt}{gray!60},
  left=6pt,right=4pt,top=3pt,bottom=3pt]
\small
\textbf{Neural Computer (NC) abstraction~(\hyperlink{teaser}{Teaser}).}
A neural system $(F, G)$ parameterized by $\theta$ that models an interactive computer interface through a single latent runtime state $h_t$ that carries executable interface state and also acts as working memory (see~\cref{eq:wm}).
\end{tcolorbox}

In the \nccligen{} experiments, the NC learns to render and execute basic command-line workflows.
It often stays aligned with the terminal buffer and captures common ``physics'' of everyday CLI use (e.g., fast scrollback, prompt wrapping, window resizing), though symbolic stability remains limited.

In the \ncguiworld{} experiments, we evaluate standard world-model designs across data quality, cursor supervision, action injection, and action encoding, using global fidelity, post-action responsiveness, and cursor-accuracy measurements.
Figure~\ref{fig:nc-general} summarizes this template across two interface-specific NCs trained separately without shared parameters. 

Our experimental insights indicate that current NCs can already learn to realize elementary runtime primitives, most notably I/O alignment and short-horizon control.
The long-term target is a Completely Neural Computer (CNC), the mature, general-purpose realization of this machine form: a fully learned computer whose compute, memory, and interfaces are unified in a single learned runtime substrate rather than engineered as separate modules.
Our current NC prototypes are an early step toward that CNC vision. 
Substantial challenges remain in robust long-horizon reasoning, reliable symbolic processing, stable capability reuse, and explicit runtime governance.
\Cref{section:toward-cnc} outlines these open challenges and a roadmap toward CNCs. 

\begin{tcolorbox}[
  colback=apricotglaze!40!white,
  colframe=metablue!40,
  interior style={shade,shading angle=315,left color=white,right color=apricotglaze!55!white},
  boxrule=0pt,
  borderline west={1pt}{0pt}{gray!60},
  left=6pt,right=4pt,top=3pt,bottom=3pt]
\small
\textbf{Completely Neural Computer (CNC) abstraction~(\Cref{sec:roadmap}).}
A Neural Computer instance is \emph{complete} (i.e., a CNC) if it is
(i) \emph{Turing complete},
(ii) \emph{universally programmable},
(iii) \emph{behavior-consistent} unless explicitly reprogrammed, and
(iv) realizes the architectural and programming-language advantages of NCs relative to conventional computers.
\end{tcolorbox}

\textbf{Concretely, this work makes the following contributions:}
{
\renewcommand\labelitemi{\large$\bullet$}
\begin{itemize}
\setlength{\itemsep}{2pt plus 1pt minus 1pt}
\setlength{\parskip}{0pt}
\item Define neural computers (NCs) and build video-based prototypes for both CLI and GUI interfaces.
\item Provide a data engine and alignment recipe that synchronize text, actions, and frames across the CLI and GUI settings studied in this paper.
\item Identify practical design choices for NCs through extensive ablation studies.
\item Outline an engineering roadmap toward completely neural computers (CNCs), centered on acceptance challenges such as reuse, consistency, and runtime governance.

\end{itemize}
}

%% file: section/preliminaries.tex
Throughout this paper, we use \emph{conventional digital computers} as an umbrella term for stored-program machines (e.g., Zuse-style or von Neumann-style architectures): 
at the theory level they are commonly abstracted as random-access machines with an instruction set architecture, and at the systems level they are typically realized through layered operating-system/application stacks. Such systems separate computation, memory, and I/O.
Our motivating question is whether a single neural network  can learn to internalize these roles inside one latent runtime state, rather than relying on an external execution environment (e.g., OS/simulator) to carry executable state.
We model a video-based neural computer (NC) prototype as a learned latent-state system that folds these roles into an update-and-render loop.

\begin{figure}[h!]
  \centering
  \includegraphics[width=1.0\linewidth]{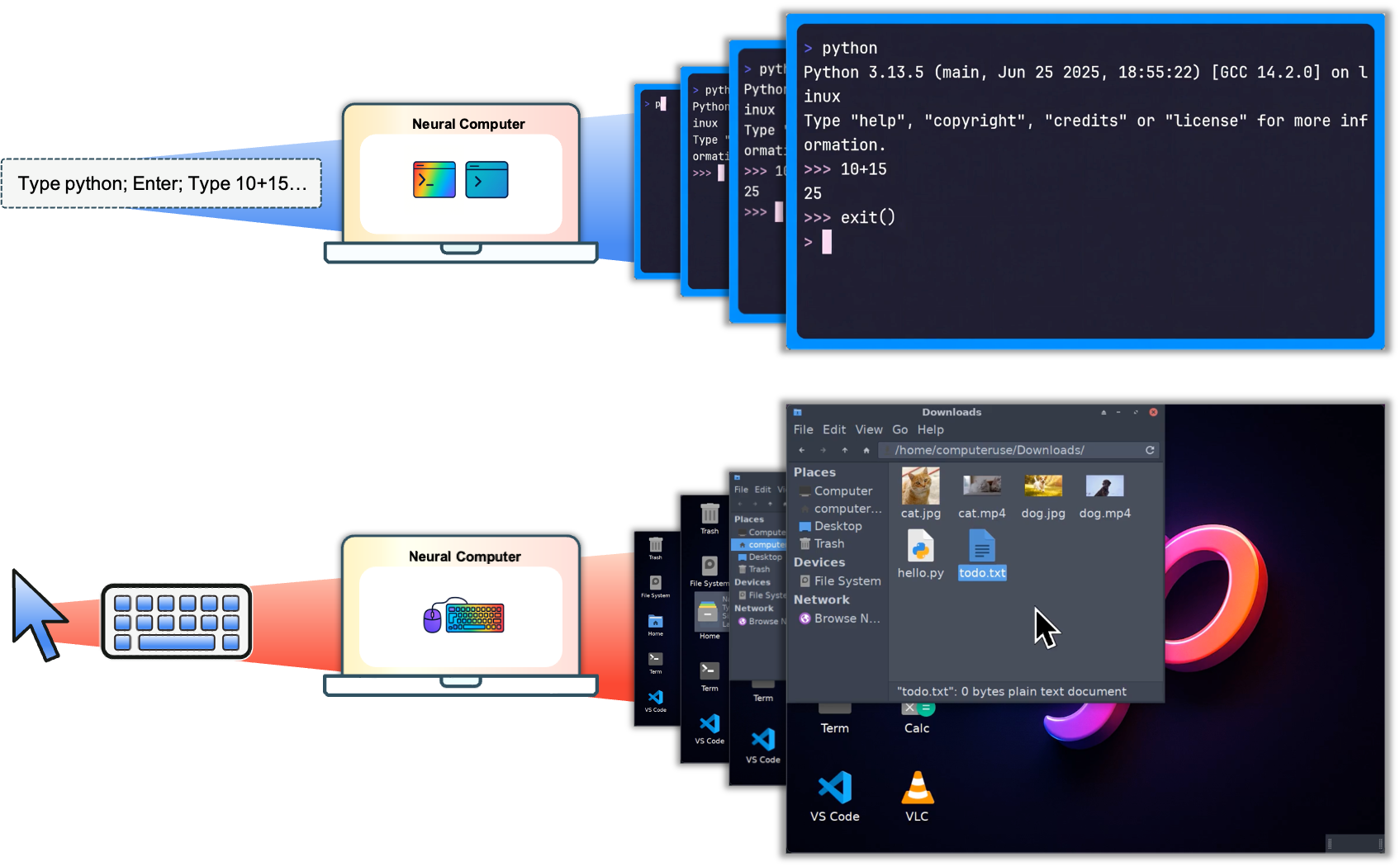}
\caption{\textbf{Neural computers across interfaces.}
Given a prompt or action stream, an NC rolls out future interface frames for \cligenGeneralLogo{}/\cligenCleanLogo{}\,\nccligen{} (top) and \guiworldLogo{}\,\ncguiworld{} (bottom).
Logos denote datasets; \nccligen{} and \ncguiworld{} are the corresponding models trained on those datasets.
It models terminal or desktop dynamics.}
  \label{fig:nc-general}
\end{figure}

Specifically, an NC updates a latent runtime state from the current observation and conditioning input, and then predicts (or samples) the next observation.
In this paper, we treat screen frames as observables and define actions as time-indexed conditions. More broadly, the NC framework can accommodate various other modalities and structural representations for both observables and actions.
Given an initial screen frame $x_0$ and conditioned on user action $u_t$ at iteration $t$, an NC updates its runtime state and samples the next frame $x_{t+1}$.
Formally, an NC defined by an initial runtime state $h_0$, an update function $F_\theta$, and a decoder $G_\theta$ operates as follows, where $G_\theta$ parameterizes a distribution over next frames:

\begin{align}
h_t &= F_\theta(h_{t-1}, x_t, u_t), \qquad x_{t+1} \sim G_\theta(h_t).
\label{eq:wm}
\end{align}

In this formulation, $h_t$ provides the persistent runtime memory, $F_\theta$ carries the state-update computation, and $(x_t, u_t, G_\theta)$ define the I/O pathway from observations and actions to the next observable state.

\noindent\textit{Notation.} We use $h_t$ for the NC latent runtime state and reserve $z$ for VAE/video latents used in diffusion-style video models (e.g., \Cref{section:impl-guiworld}). 

This update-and-render loop can be described using world-model terminology, where $x_t$ are observations and $u_t$ provides conditioning.
In that terminology, the input sequence $\{u_t\}$ is referred to as a conditioning stream.
This view supplies practical machinery for the current prototype, but an NC is not merely a predictor of interface dynamics: it is a learned runtime mechanism in which the latent state $h_t$ carries executable context, $F_\theta$ integrates new observations and inputs, and $G_\theta$ renders the next frame.
Auxiliary heads can encode and decode prompts, buffers, or action traces, shifting functionality that would traditionally live in OS queues, device drivers, and UI toolkits into latent-state dynamics.

\subsection{Related Work}
\label{section:prelim-related}

Early neuromorphic designs~\citep{mead2012analog} explored neural computation as a physical substrate.
Differentiable memory and program-execution architectures, including fast weight programmers and self-referential weight matrices~\citep{schmidhuber92fastweights,schmidhuber93selfref,schmidhuber93ratio}, Neural Turing Machines~\citep{graves2014neural},
Differentiable Neural Computers~\citep{graves2016hybrid}, and Neural Programmer-Interpreters~\citep{reed2015neural}, showed how neural controllers can learn to execute structured procedures and manipulate memories. Neural predictive control can be realized through
general-purpose differentiable world models~\citep{schmidhuber1990making} and their precursors~\citep{werbos1987learning,munro1987dual,nguyen1990truck}.
Latent video and world models~\citep{schmidhuber2015learning,ha2018world,hafner2019learning,hafner2019dream,bruce2024genie} also apply such ideas to embodied control in interactive environments.
Genie~3~\citep{bruce2024genie}, in particular, frames such models as agent-training substrates with improved physical consistency.
More recently, high-capacity generators such as Veo~3~\citep{google_veo3_1_2025} and Sora~2~\citep{openai_sora2_2025} emphasize open-ended, photorealistic simulation. 
Systems such as Imagine with Claude~\citep{anthropic_claude_sonnet4_5_2025} apply model-based conditioning to desktop and DOM-style interfaces.
NeuralOS~\citep{rivard2025neuralos} focus on neural simulation of graphical interfaces (GUIs). We extend this to a broader set of interaction paradigms (GUI + CLI) scaled to more practical scenarios in a unified framework. We systematically study key properties such as I/O alignment, interaction consistency, and short-horizon control, while identifying open issues such as symbolic stability and reliable capability reuse. This illustrates the potential of current techniques and exhibits remaining challenges: a step toward a unified neural execution layer that may eventually act as the machine itself. 

%% file: section/model.tex
We build on the Wan2.1 model~\citep{wan2025wan}, which was a state-of-the-art video generation model at the time of our experiments.
We add NC-specific conditioning and action modules, together with interface-specific training recipes.
Figure~\ref{fig:nc-general} illustrates this setup: NCs take a prompt or action stream as input and generate future interface frames in both CLI and GUI settings. In the present prototypes, these prompts and actions are logged conditioning streams, so evaluation remains open-loop rather than closed-loop interaction with a live environment.
We refer to these two instantiations as CLIGen, our command-line interface (CLI) prototype (\Cref{section:impl-cligen}), and GUIWorld, our graphical user interface (GUI) prototype (\Cref{section:impl-guiworld}).

In this video-based instantiation, the NC latent runtime state $h_t$ is realized by the model’s time-indexed video latents $z_t$.
Under this abstraction, the diffusion transformer acts as the state-update map: it consumes prior latents together with the current observation and conditioning inputs, and produces the updated state $h_t$ (realized as $z_t$).
The decoder $G_\theta$ parameterizes a distribution over the next frame $x_{t+1}$.
Auxiliary heads encode and decode conditioning streams $u_t$, including text prompts and action traces.
Structured logs such as terminal buffers are used for alignment and evaluation where available, not as privileged model-state inputs.

%% file: section/impl_cli.tex
\begin{figure}[ht]
  \centering
  \includegraphics[width=0.8\linewidth]{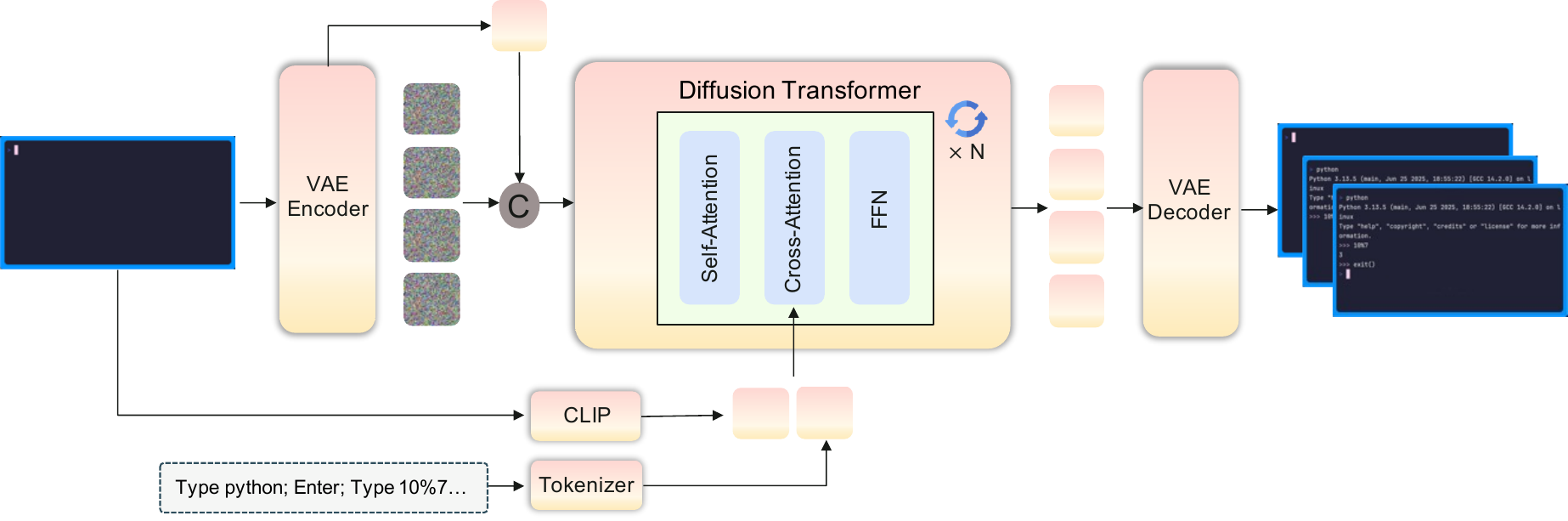}
  \caption{\cligenGeneralLogo{}\,/\,\cligenCleanLogo{}\,NC$_\text{CLIGen}$ architecture.
  Terminal frames are observations $x_t$.
  A prompt and the first frame seed the conditioning stream.
  The Wan2.1-based latent state $z_t$ rolls forward under the standard I2V sampling scheme.}
  \label{fig:cligen-arch}
\end{figure}

CLIGen instantiates the NC abstraction in command-line interfaces.
Observations $x_t$ are terminal frames rendered from the underlying text buffer.
The conditioning stream $u_t$ carries a user prompt and optional metadata, and the video latent state $z_t$ implements the latent runtime state $h_t$ by tracking CLI context across frames.
At inference time, the model rolls out from the prompt and first frame, updates $z_t$, and predicts future terminal frames (\Cref{fig:cligen-arch}).
We use two CLI datasets: \cligenGeneralLogo{}\,CLIGen (General), which contains diverse, open-ended terminal traces, and \cligenCleanLogo{}\,CLIGen (Clean), which contains deterministic Dockerized traces. We train one \nccligen{} model per dataset under the same architecture.

\begin{tcolorbox}[
  colback=apricotglaze!40!white,
  colframe=metablue!40,
  interior style={shade,shading angle=315,left color=white,right color=apricotglaze!55!white},
  boxrule=0pt,
  borderline west={1pt}{0pt}{gray!60},
  left=6pt,right=4pt,top=3pt,bottom=3pt]
\small
\setlength{\tabcolsep}{5pt}
\renewcommand{\arraystretch}{1.06}
\begin{tabularx}{\linewidth}{@{}L{0.24\linewidth} Z L{0.16\linewidth}@{}}
\toprule
\textbf{Dataset} & \textbf{Data characteristics} & \textbf{Model} \\
\midrule
\cligenGeneralLogo{}\,CLIGen (General) & diverse, open-ended terminal traces & \cligenGeneralLogo{}\,\nccligen{} \\
\cligenCleanLogo{}\,CLIGen (Clean) & deterministic Dockerized scripts with cleaner, better-paced buffers & \cligenCleanLogo{}\,\nccligen{} \\
\bottomrule
\end{tabularx}
\end{tcolorbox}

\newcommand{\cliframe}[2]{%
  \begin{tikzpicture}[baseline={(current bounding box.center)}]
    \node (img) [inner sep=0pt] {\includegraphics[width=0.34\linewidth,keepaspectratio]{#1}};
    \node[anchor=south east, fill={rgb,1:red,0.98;green,0.91;blue,0.73}, text=black, circle, inner sep=2pt] at ([xshift=-4pt,yshift=4pt]img.south east) {\scriptsize #2};
  \end{tikzpicture}%
}
\newcommand{\cliframeblock}{%
  \makebox[\linewidth][c]{%
    \begingroup
    \setlength{\tabcolsep}{0.26em}
    \renewcommand{\arraystretch}{1.0}%
    \begin{tabular}{ccc}
      \cliframe{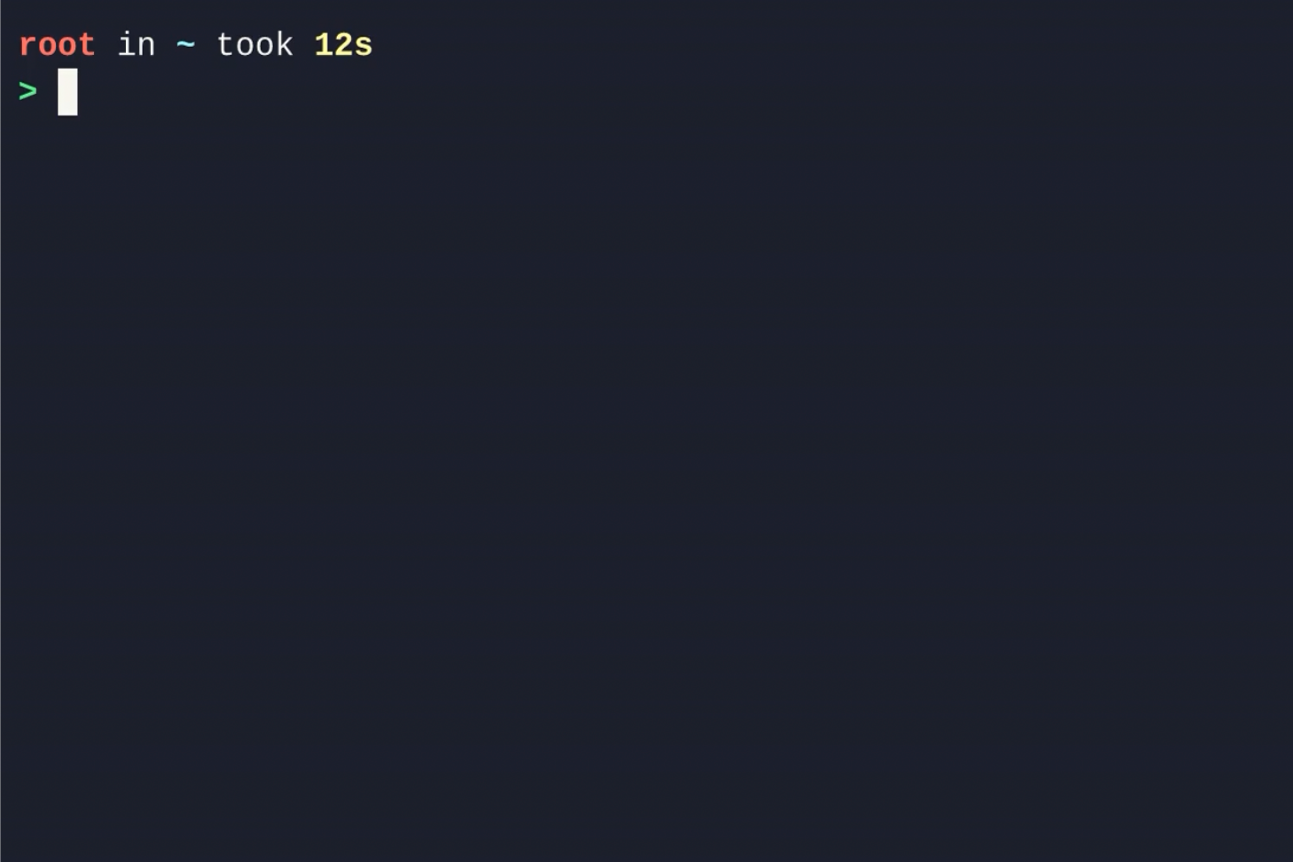}{1} & \cliframe{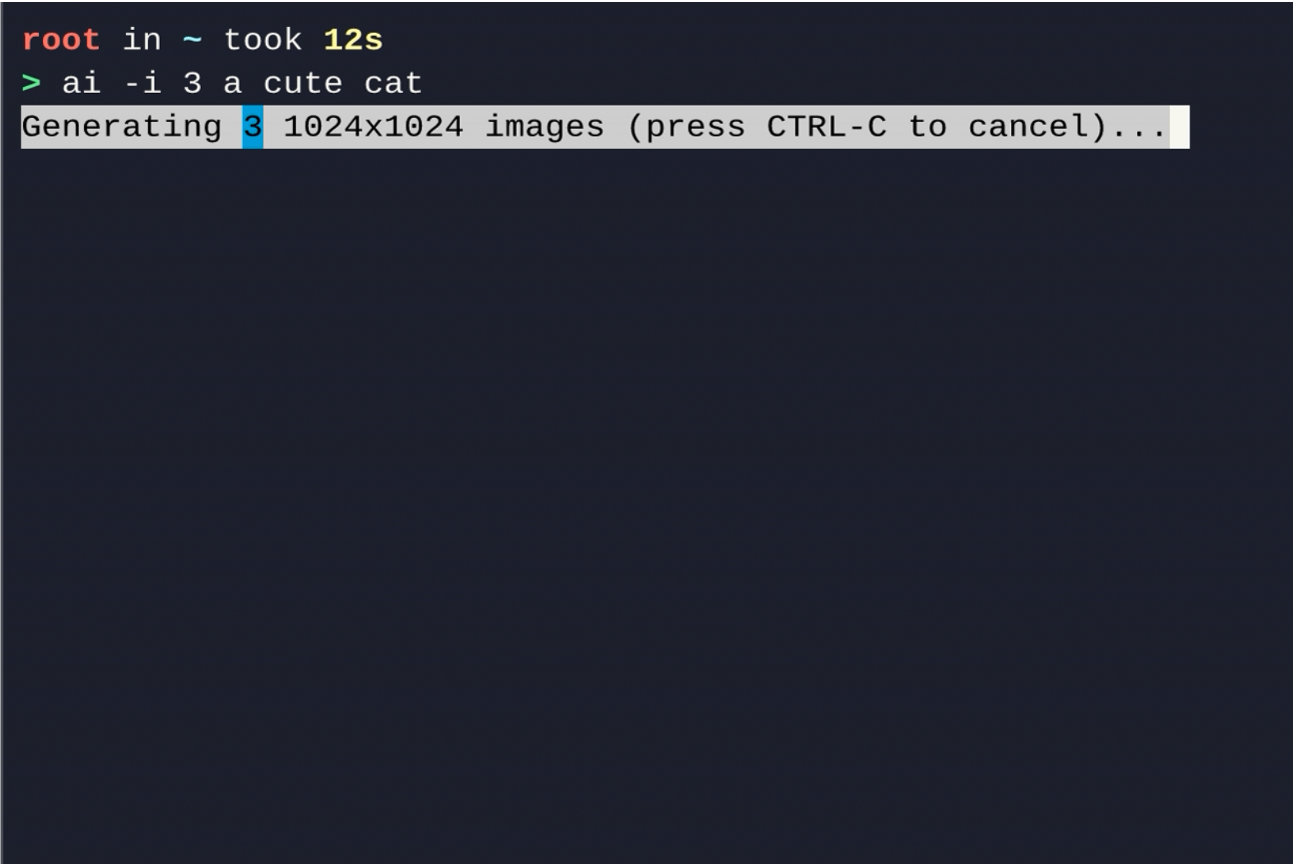}{2} & \cliframe{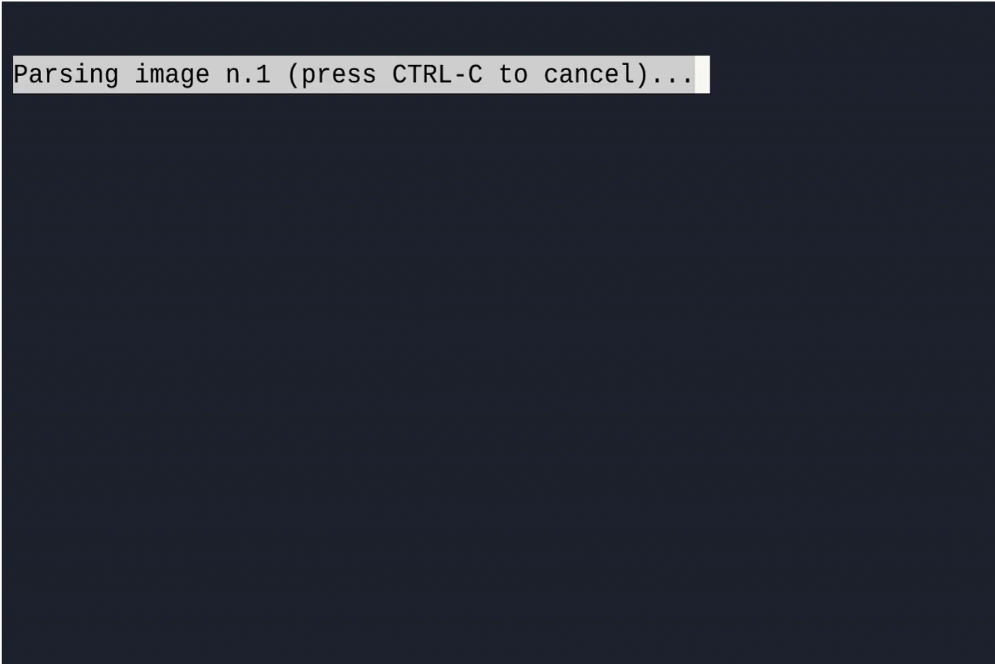}{3} \\
      \addlinespace[0.48em]
      \cliframe{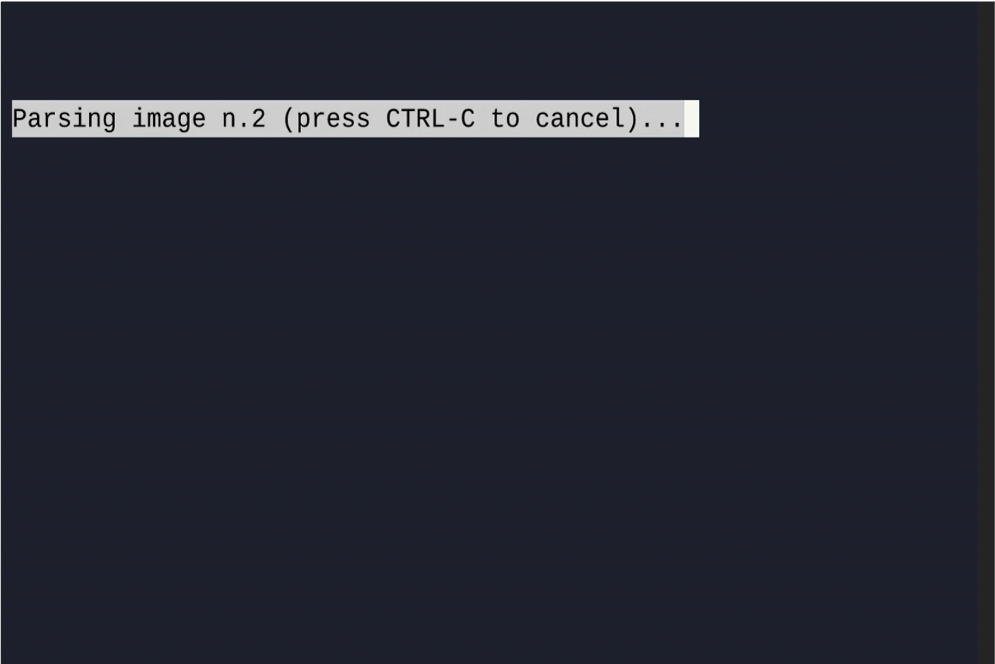}{4} & \cliframe{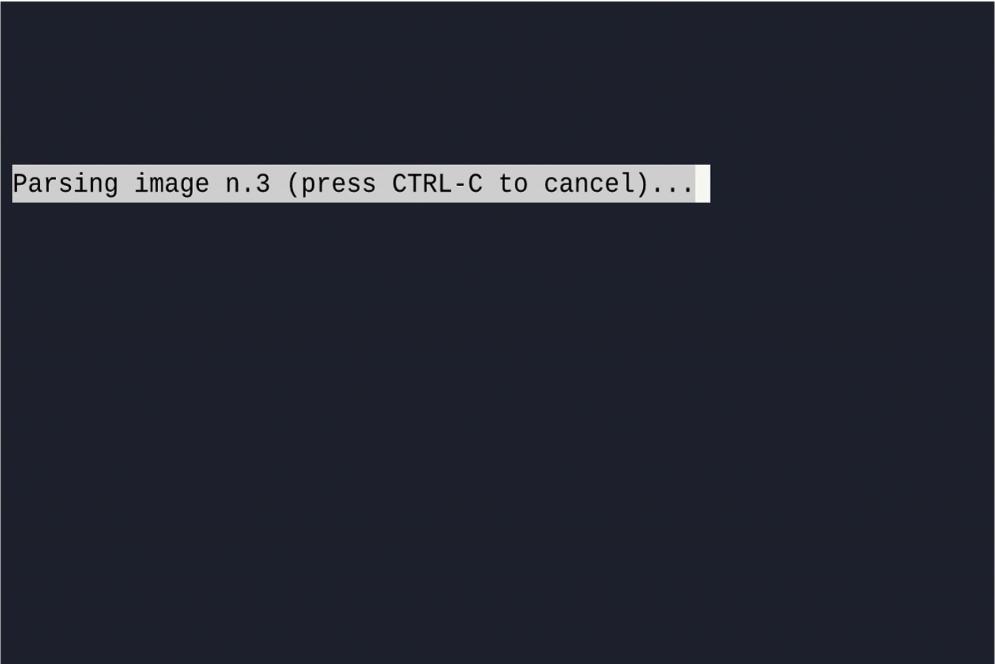}{5} & \cliframe{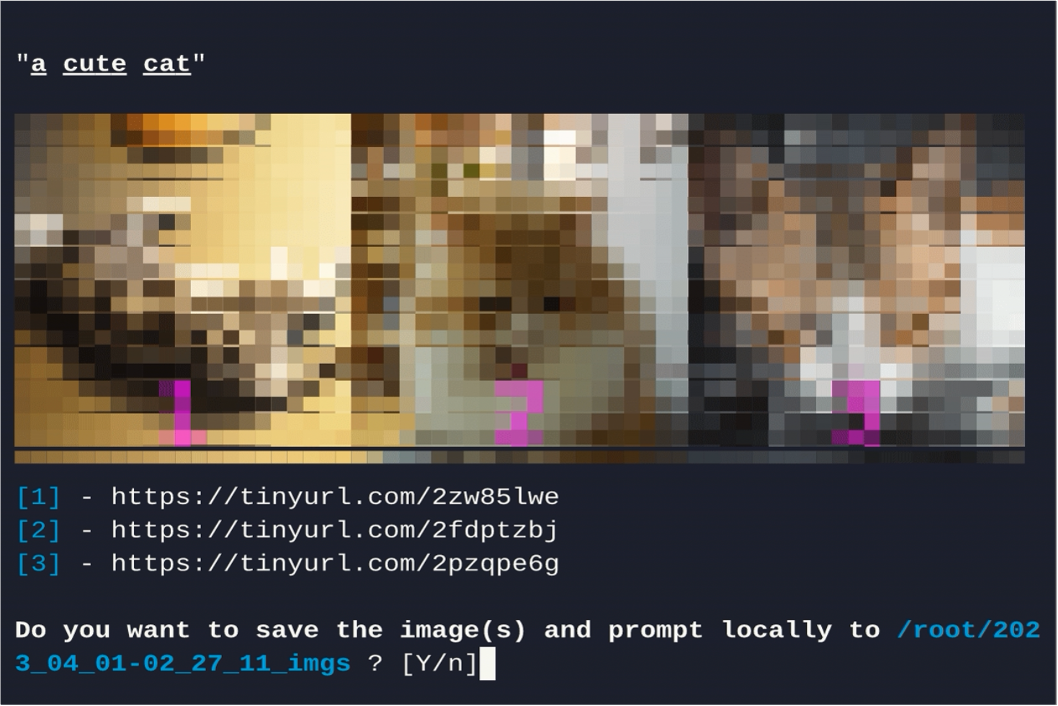}{6}
    \end{tabular}%
    \endgroup}}

\newcommand{\cliframeclean}[2]{%
  \begin{tikzpicture}[baseline={(current bounding box.center)}]
    \node (img) [inner sep=0pt] {\includegraphics[width=0.34\linewidth,keepaspectratio]{#1}};
    \node[anchor=south east, fill={rgb,1:red,0.98;green,0.91;blue,0.73}, text=black, circle, inner sep=2pt] at ([xshift=-4pt,yshift=4pt]img.south east) {\scriptsize #2};
  \end{tikzpicture}%
}
\newcommand{\cliframecleanblock}{%
  \makebox[\linewidth][c]{%
    \begingroup
    \setlength{\tabcolsep}{0.26em}
    \renewcommand{\arraystretch}{1.0}%
    \begin{tabular}{ccc}
      \cliframeclean{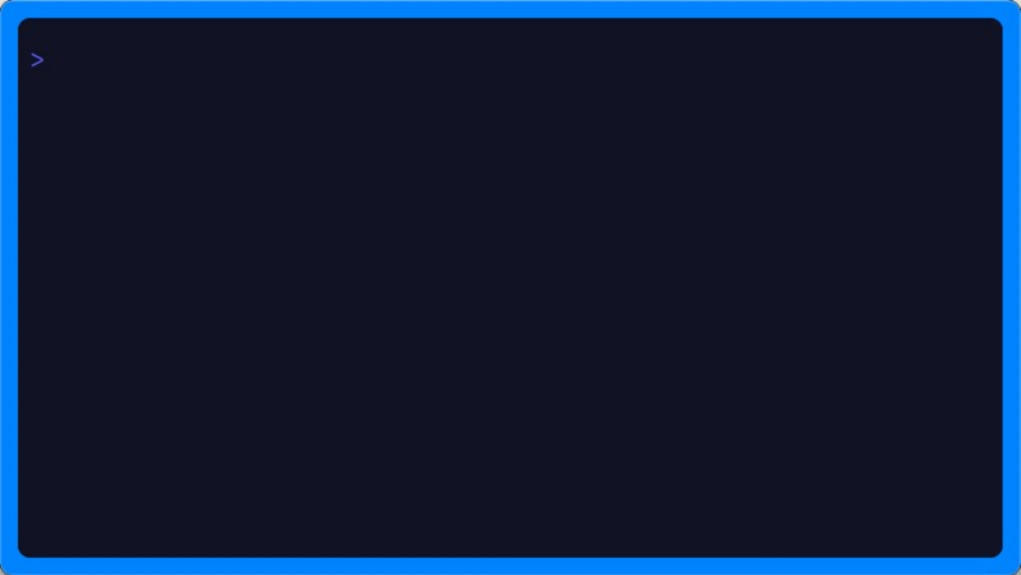}{1} & \cliframeclean{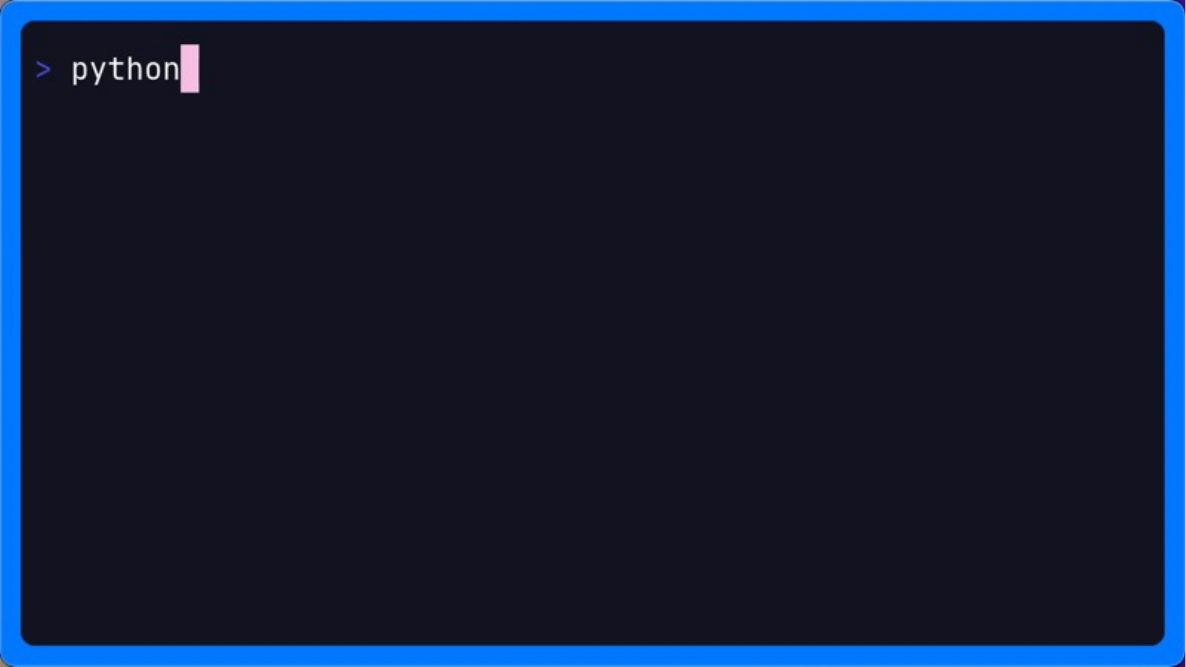}{2} & \cliframeclean{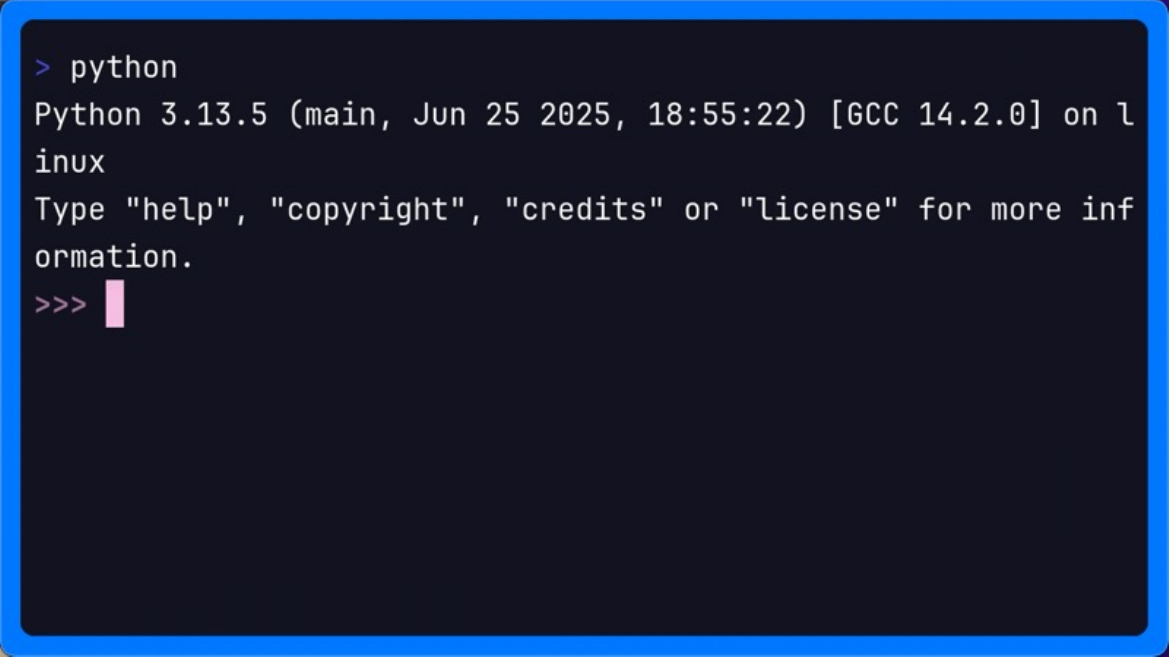}{3} \\
      \addlinespace[0.48em]
      \cliframeclean{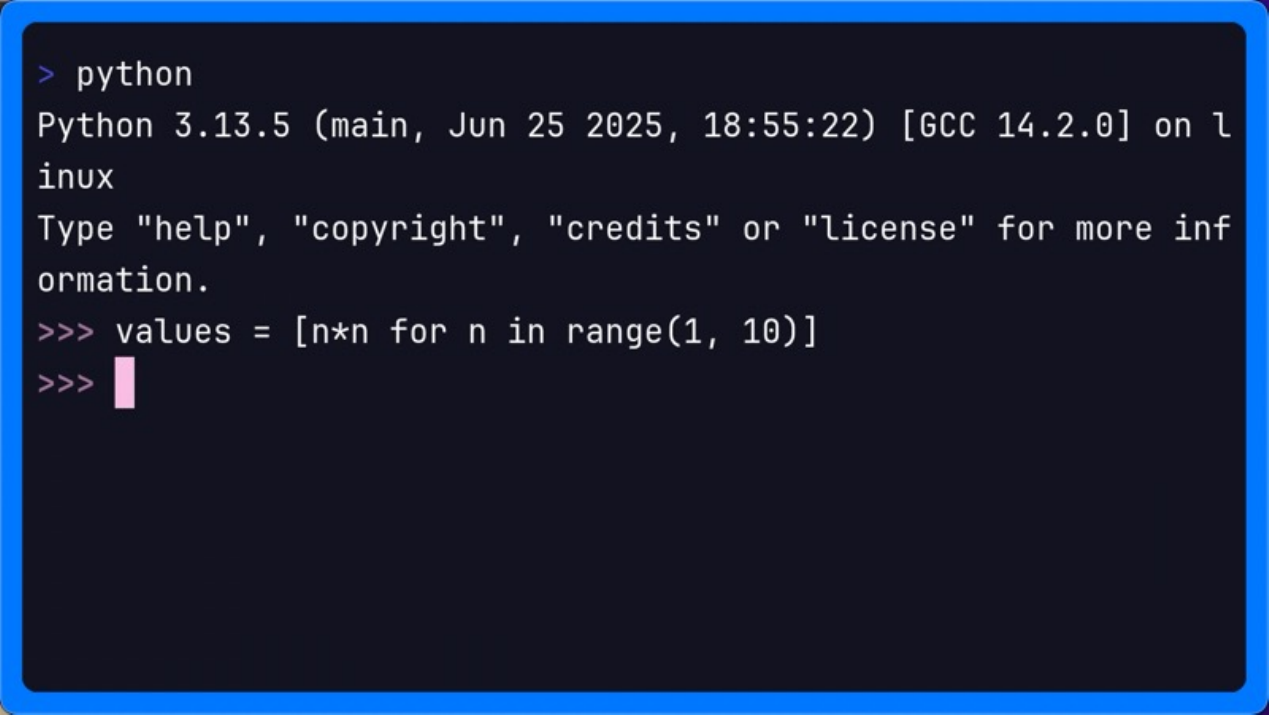}{4} & \cliframeclean{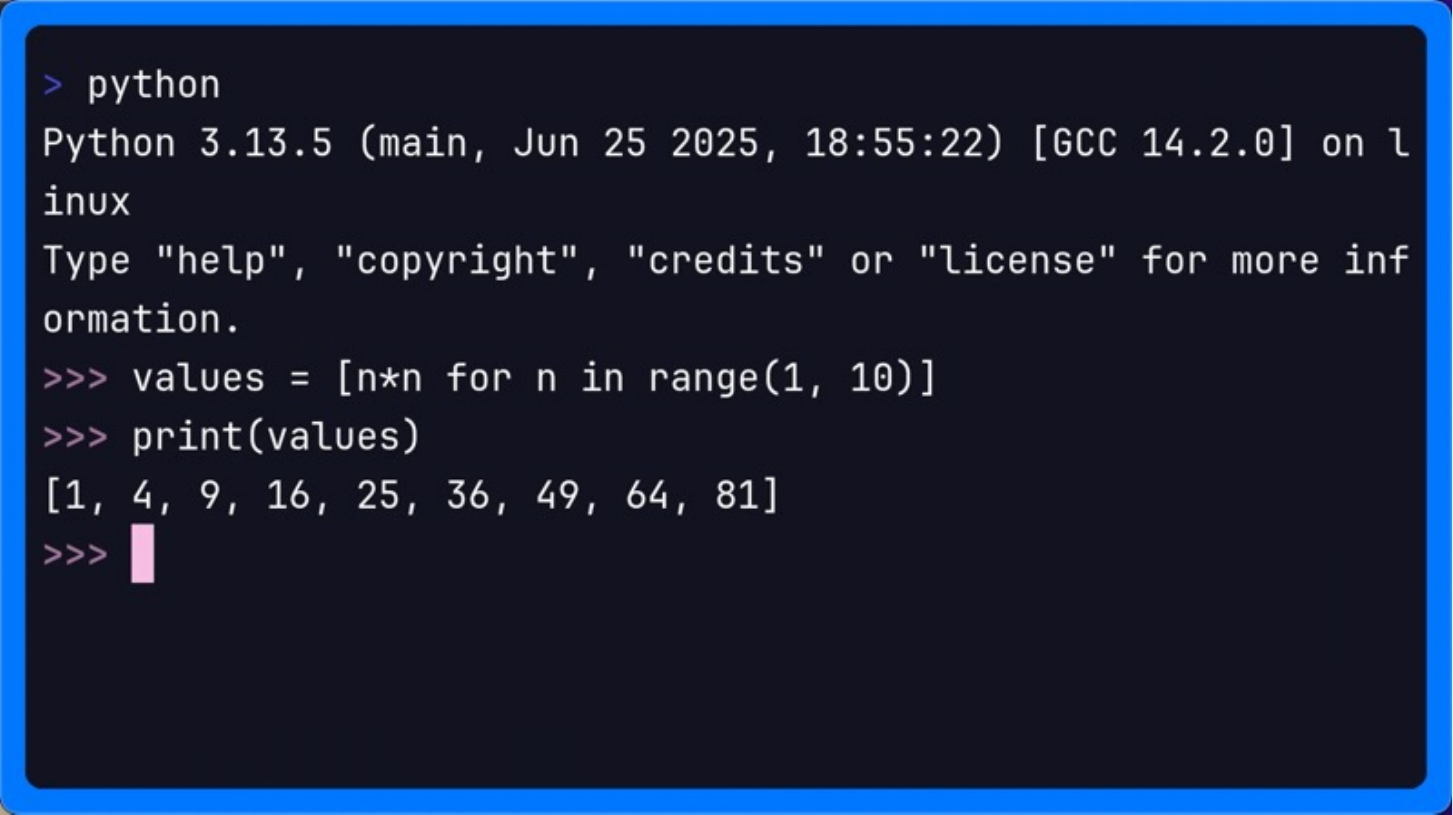}{5} & \cliframeclean{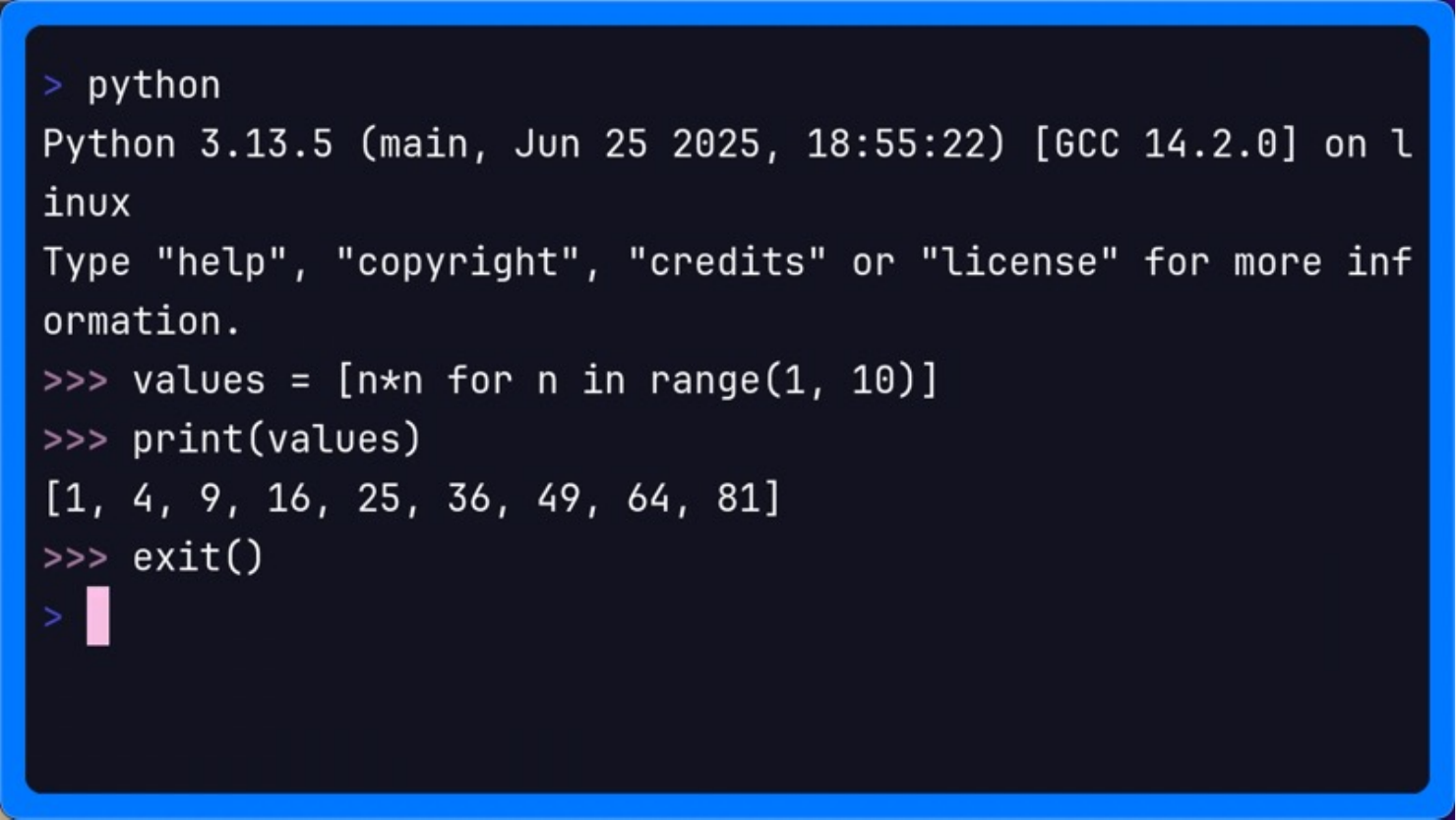}{6}
    \end{tabular}%
    \endgroup}}

\begin{smallschemetable}{Data samples for \cligenGeneralLogo{}\,CLIGen (General) and \cligenCleanLogo{}\,CLIGen (Clean).}{tab:cligen-prompts}
  \begin{tabularx}{\linewidth}{X}
    \multicolumn{1}{c}{\textbf{Frames from Sample — \cligenGeneralLogo{}\,CLIGen (General)}}\\[0.25em]
    \multicolumn{1}{c}{\cliframeblock}\\[7.0em]
    \noalign{\vspace{0.9em}}
    \makebox[\textwidth][c]{\parbox{1.01\textwidth}{%
    \prompttierline{Semantic}
    A root terminal session kicks off an AI command to make three 1024x1024 cat shots, shows quick parsing for each one, then presents pixelated cat previews with numbered links and asks whether to stash them in \texttt{/root/2023\_04\_01-02\_27\_11\_imgs}. \par\medskip
	    \prompttierline{Regular}
	    In a root shell at \texttt{\texttildelow}, the user runs \texttt{ai -i 3 a cute cat}, watches a green progress line announcing three 1024x1024 images, sees sequential parsing messages for images 1 through 3, and ends on a preview pane with three pixelated cat thumbnails, numbered download links, and a save prompt targeting \texttt{/root/2023\_04\_01-02\_27\_11\_imgs}. \par\smallskip
	    \prompttierline{Detailed}
	    In a dark-background terminal at the \texttt{root in \texttildelow} prompt, the user types \texttt{ai -i 3 a cute cat}. The screen prints \texttt{Generating 3 1024x1024 images (press CTRL-C to cancel)...}, shows parsing messages for images 1--3, and ends on a preview pane with three numbered thumbnails and a save prompt targeting \texttt{/root/2023\_04\_01-02\_27\_11\_imgs}.}}\\[0.7em]
    \midrule
    \multicolumn{1}{c}{\textbf{Frames from Sample — \cligenCleanLogo{}\,CLIGen (Clean)}}\\[0.25em]
    \multicolumn{1}{c}{\cliframecleanblock}\\[6.2em]
    \noalign{\vspace{0.6em}}
    \makebox[\textwidth][c]{\parbox{1.01\textwidth}{%
    \prompttierline{Scripted caption}
    Type \texttt{python}; Enter; Type \texttt{values = [n*n for n in range(1, 10)]}; Enter; Type \texttt{print(values)}; Enter; Type \texttt{exit()}; Enter.}}\\
  \end{tabularx}
\end{smallschemetable}

\subsubsection{Data pipeline}

The \cligenGeneralLogo{}\,CLIGen (General) dataset is built from publicly available \texttt{asciinema .cast} trajectories\footnote{\url{https://asciinema.org/}}.
The \texttt{asciinema} stack records and replays terminal sessions with synchronized timing and ANSI-faithful decoding.
We replay each session with the official tools and render it into terminal frames, preserving palette transitions, cursor state, and terminal geometry.
Frames, text buffers, and keyboard-event logs share a single monotonic clock.
At render time, we normalize resolution and aspect ratio and apply a filter to remove sensitive strings.
We render sessions to GIF using \href{https://github.com/asciinema/agg}{\texttt{agg}} and convert them to video with \href{https://github.com/FFmpeg/FFmpeg}{\texttt{ffmpeg}}.

We segment each recording into roughly five-second clips using content-aware splits. 
We temporally normalize each clip to a fixed length: shorter clips repeat the final frame, and longer clips are uniformly subsampled. 
The resulting 823{,}989 video streams (approximately 1{,}100 hours) are resampled to 15~FPS.
Underlying buffers and logs are used to generate aligned textual descriptions with Llama~3.1~70B~\citep{dubey2024llama} in three styles (semantic, regular, and detailed), which serve as prompts.
As shown in Figure~\ref{fig:data-teaser} (left), this split spans diverse real-world terminal use cases\footnote{Additional preprocessing details and a \texttt{.cast} example are in \Cref{appendix:pipeline,appendix:cast-format}, with a sample overview in \Cref{tab:cligen-prompts}.}.

The \cligenCleanLogo{}\,CLIGen (Clean) dataset is collected using the open-source \href{https://github.com/charmbracelet/vhs}{\texttt{vhs}} toolkit.
It enables repeatable terminal demonstrations and integration tests through scripted execution.
Deterministic scripts drive Dockerized environments to capture cleaner, better-paced traces.
We authored roughly $250$k scripts.
After filtering (51.21\% retained), we keep two subsets.
The first contains approximately $78$k regular traces (package installation, log filtering, interactive REPL usage, etc.).
The second contains approximately $50$k Python math validation traces.
Captions are derived directly from the raw \texttt{vhs} scripts for clarity.
We standardize frame rendering by fixing one monospace font/size, using a consistent palette for success and error highlights, and locking resolution and theme to remove typography-related confounds.
Each episode records its caption type and font settings for later slicing.
Clips longer than five seconds are uniformly subsampled for training, while shorter clips repeat the final frame to normalize length\footnote{Additional details are provided in \Cref{appendix:pipeline,appendix:vhs-format}, with a representative data sample in \Cref{tab:cligen-prompts}.}.

\subsubsection{Model architecture}

We treat CLI generation as text-and-image-to-video: a caption and the first terminal frame condition the rollout.
The first frame is encoded by a VAE into a conditioning latent.
In parallel, a CLIP image encoder~\citep{radford2021learning} extracts visual features from the same frame, and a text encoder (e.g., T5~\citep{raffel2020exploring}) embeds the caption.
Following the Wan2.1 image-to-video (I2V) design, these conditioning features are concatenated with diffusion noise, projected through a zero-initialized linear layer, and processed by a DiT stack.
Decoupled cross-attention injects the joint caption and first-frame context derived from the CLIP and text features. 
The VAE encodes and decodes terminal frames.
During generation, the diffusion transformer advances the latent state $z_t$ under the original Wan2.1 I2V sampling schedule, without additional binary masks or periodic reseeding.

\subsubsection{Implementation Details}

Training uses gradient checkpointing and applies dropout 0.1 to the prompt encoder, CLIP, and VAE modules.
Optimization uses AdamW (learning rate $5\times10^{-5}$, weight decay $10^{-2}$), \texttt{bfloat16} precision, and gradient clipping at 1.0.
Training \nccligen{} on CLIGen (General) requires $\sim$15{,}000 H100 GPU hours at batch size 1.
Training on CLIGen (Clean) across both subsets requires $\sim$7{,}000 H100 GPU hours.

\subsubsection{Evaluations}

\begin{wrapfigure}{r}{0.7\linewidth}
  \vspace{-0.8em}
  \centering
  \includegraphics[width=\linewidth]{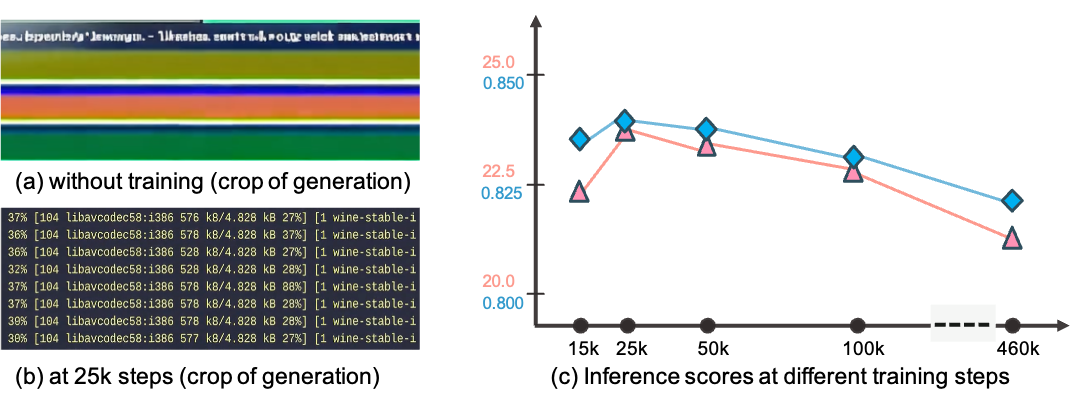}
  \caption{(a--b) Qualitative generations before and after CLIGen training; (c) CLIGen (Clean) \PSNR/\SSIM~plateau around 25k training steps.}
  \label{fig:cligen-long-train}
  \vspace{-0.8em}
\end{wrapfigure}

\begin{wrapfigure}{r}{0.49\linewidth}
  \vspace{-1.8em}
  \centering
  \includegraphics[width=\linewidth]{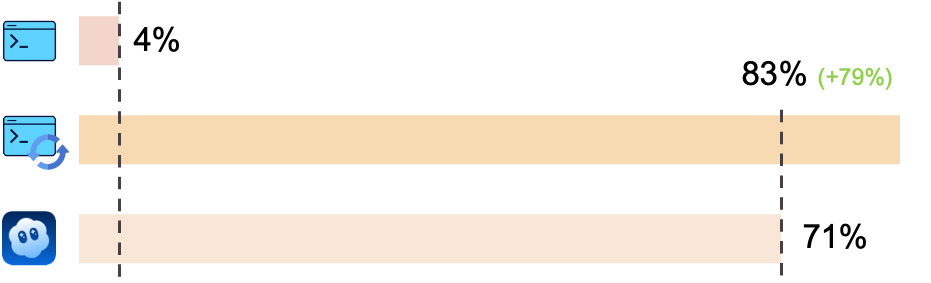}
  \caption{Reprompting boosts performance to 83\%.}
  \label{fig:cligen-exp6}
  \vspace{-1.0em}
\end{wrapfigure}

Unless otherwise noted, NC in this section refers to the current video-based CLI prototype. We report six practical takeaways:
\begin{tcolorbox}[
  colback=apricotglaze!40!white,
  colframe=metablue!40,
  interior style={shade,shading angle=315,left color=white,right color=apricotglaze!55!white},
  boxrule=0pt,
  borderline west={1pt}{0pt}{gray!60},
  left=6pt,right=4pt,top=3pt,bottom=3pt]
\small
\begin{badgeitems}
  \item The NC maintains high-fidelity terminal rendering at practical font sizes (e.g., 13\,px), preserving readable interface state.
  \item Prompt specificity is an effective control channel: detailed, literal captions improve text-to-pixel alignment.
  \item On clean but domain-specific data, global PSNR/\SSIM~plateau around 25k steps (Figure~\ref{fig:cligen-long-train}), indicating early saturation in reconstruction metrics rather than a complete halt in learning.
  \item The NC reproduces complex terminal appearances while sustaining coherent short-horizon command rollouts under fixed conditioning.
  \item Symbolic computation remains the main bottleneck: structured arithmetic reveals reliability limits, motivating stronger symbolic or system-level conditioning.
  \item In our setting, without changing the NC backbone or adding RL, reprompting improves symbolic probes (4\%$\rightarrow$83\%; Figure~\ref{fig:cligen-exp6}), reinforcing the view that current models are strong renderers and conditionable interfaces rather than native reasoners (Table~\ref{tab:sora-leads}).
\end{badgeitems}
\end{tcolorbox}

\exptitle[\cligenGeneralLogo]{1}{The NC stays readable at practical font sizes}

\begin{wraptable}{r}{0.42\linewidth}
  \vspace{-22pt}
  \captionsetup{type=table,width=\linewidth}
  \centering
  \small
  \rowcolors{2}{tablewarmrow}{white}
  \caption{Reconstruction quality.}
  \label{tab:cli-vae-recon}
  \setlength{\tabcolsep}{7pt}
  \begin{tabularx}{\linewidth}{l >{\centering\arraybackslash}X}
    \toprule
    \rowcolor{tablewarmhead} \textbf{Metric} & \textbf{Value} \\
    Average PSNR & 40.77 \\
    Average SSIM & 0.989 \\
    \bottomrule
  \end{tabularx}
\end{wraptable}
The paper on NeuralOS~\citep{rivard2025neuralos} argues that generic natural-image VAEs can perform poorly on structured computer screenshots.
We test this directly by applying the Wan2.1 VAE~\citep{wan2025wan} to terminal content.
In our setting, reconstruction quality is primarily governed by font size.
At 13\,px, it is high (40.77\,dB \PSNR, 0.989 \SSIM).
At 6\,px, text exhibits noticeable blurring even when global \PSNR/\SSIM~remain strong, because background regions dominate these metrics.

\begin{figure}[h!]
  \centering
  \includegraphics[width=\linewidth]{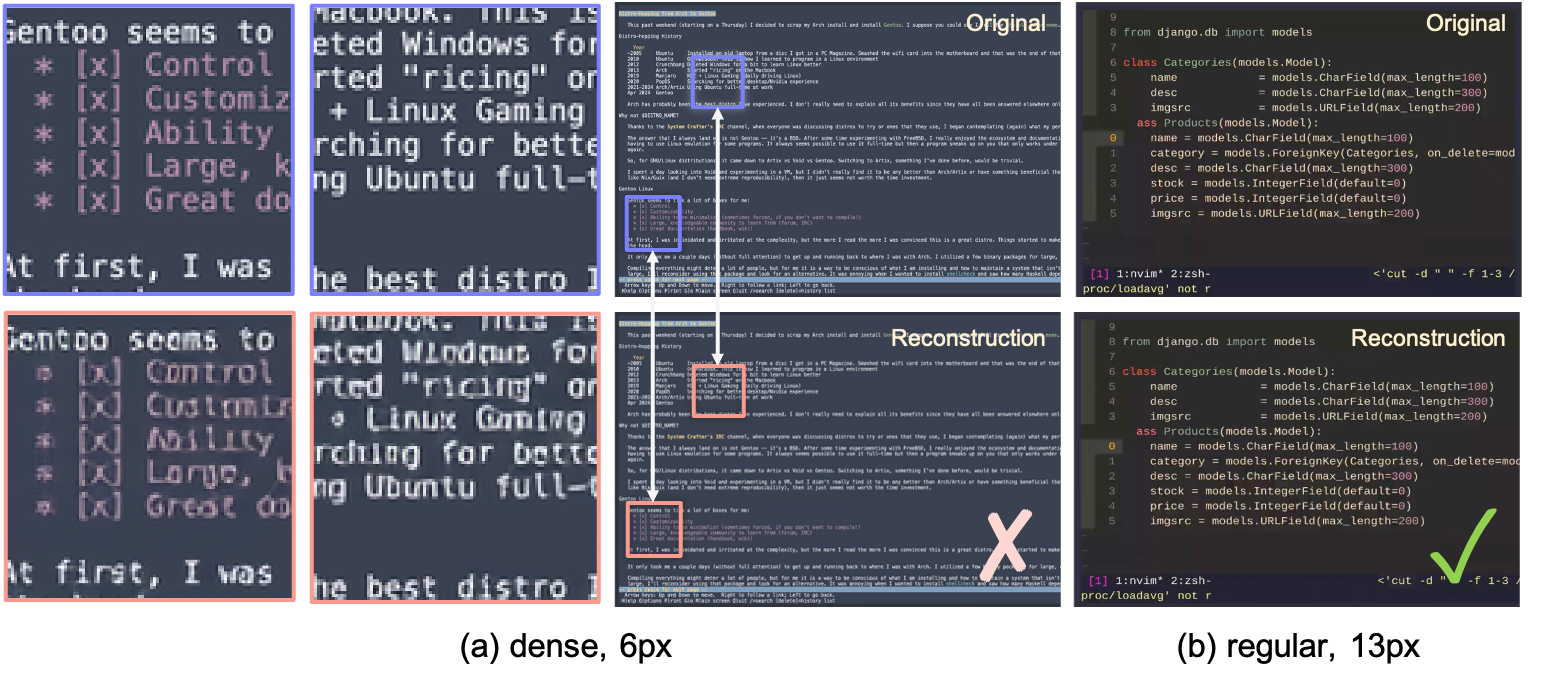}
    \vspace{-15pt}
\caption{Wan2.1 VAE reconstructions on CLIGen (General) terminal frames at different font sizes.}
  \label{fig:cligen-exp1}

\end{figure}

However, a sweep over CLIGen (General) frames shows that this effect is confined to extreme cases (\Cref{fig:cligen-exp1}).
Very small 6\,px fonts and ultra-dense text exhibit localized blurring despite high global \PSNR.
In contrast, the 13\,px terminal font used in CLIGen remains visually sharp across panes and commands. 
These results indicate that the VAE is adequate for regular CLIGen usage and highlight that sensible font choices help ensure stable NC training.

\exptitle[\cligenCleanLogo]{2}{Performance plateaus early and can degrade with prolonged training}

On clean but domain-specific structured interfaces, global reconstruction metrics improve rapidly early and then show limited additional gains under the current training objective.
In CLIGen (Clean), \PSNR/\SSIM~plateau quickly, suggesting that further optimization becomes bottlenecked less by model capacity than by the quality and pacing of the available supervision.
After the early gains, the remaining errors are often tied to artifact-prone signals (e.g., rendering glitches or rapid screen changes that disrupt temporal alignment), so additional training on the same objective can yield diminishing or even slightly unstable returns in these perceptual metrics.

Panels (a--b) illustrate the effect of training on CLIGen data.
Without CLIGen fine-tuning, Wan2.1 produces garbled terminal outputs (a).
After 25k steps, the model generates readable text with consistent formatting and color cues (b).

Figure~\ref{fig:cligen-long-train} plots the corresponding \PSNR/\SSIM~curves and shows that these global perceptual metrics flatten around 25k steps.
They improve little with further training up to 460k steps, and extended optimization can even slightly reduce them.
One plausible explanation is that most learnable structured patterns are acquired early, and further gains require higher-quality, better-paced, or more informative supervision.

\exptitle[\cligenGeneralLogo]{3}{Literal captions drive rendering accuracy}

Caption specificity has a strong effect on terminal rendering quality.
As shown in Table~\ref{tab:cligen-captions}, detailed, literal descriptions improve reconstruction fidelity.
\PSNR increases from 21.90~dB (semantic) to 26.89~dB (detailed), a gain of nearly 5~dB, compared to less specific, high-level semantic descriptions.

The three caption tiers correspond to the same underlying terminal sequence but differ in length and granularity.
\textbf{Semantic} captions (average 55 words) provide high-level summaries (e.g., ``a terminal session generates three cat images'').
\textbf{Regular} captions (average 52 words) include key commands and outputs (e.g., \texttt{ai -i 3 a cute cat}, status messages).
\textbf{Detailed} captions (average 76 words) transcribe screen content more exhaustively, including exact text, colors, and formatting.

\begin{wraptable}{r}{0.42\linewidth}
  \vspace{-1.8em}
  \captionsetup{type=table,width=\linewidth}
  \centering
  \small
  \rowcolors{2}{tablewarmrow}{white}
  \caption{Caption styles versus TI2V fidelity.}
  \label{tab:cligen-captions}
  \setlength{\tabcolsep}{5pt}
  \begin{tabularx}{\linewidth}{l >{\centering\arraybackslash}X >{\centering\arraybackslash}X >{\centering\arraybackslash}X}
    \toprule
    \rowcolor{tablewarmhead} \textbf{Prompt style} & \textbf{\PSNR} & \textbf{\SSIM} & \textbf{Avg. words} \\
    Semantic & 21.90 & 0.813 & 55 \\
    Regular & 23.63 & 0.843 & 52 \\
    Detailed & \textbf{26.89} & \textbf{0.867} & \textbf{76} \\
    \bottomrule
  \end{tabularx}
  \vspace{-1.5em}
\end{wraptable}
This progression helps explain why literal descriptions are particularly effective for terminal rendering.
Unlike natural images, which are dominated by global style patterns, terminal frames are governed primarily by text placement.
Detailed captions act as scaffolding—explicitly specifying which tokens appear where—thereby enabling precise text-to-pixel alignment.

\exptitle[\cligenCleanLogo]{4}{Neural computers achieve accurate character-level text generation}

\begin{wraptable}{r}{0.42\linewidth}
  \vspace{-0.8em}
  \captionsetup{type=table,width=\linewidth}
  \centering
  \small
  \rowcolors{2}{tablewarmrow}{white}
  \caption{OCR accuracy versus training.}
  \label{tab:cligen-ocr}
  \setlength{\tabcolsep}{7pt}
  \begin{tabularx}{\linewidth}{l >{\centering\arraybackslash}X >{\centering\arraybackslash}X}
    \toprule
    \rowcolor{tablewarmhead} \textbf{Steps (k)} & \textbf{Char. acc.} & \textbf{Exact line} \\
    0  & 0.03 & 0.01 \\
    10 & 0.18\,$\color{green!60!black}{\uparrow}_{0.15}$ & 0.05\,$\color{green!60!black}{\uparrow}_{0.04}$ \\
    20 & 0.33\,$\color{green!60!black}{\uparrow}_{0.30}$ & 0.12\,$\color{green!60!black}{\uparrow}_{0.11}$ \\
    30 & 0.41\,$\color{green!60!black}{\uparrow}_{0.38}$ & 0.18\,$\color{green!60!black}{\uparrow}_{0.17}$ \\
    40 & 0.52\,$\color{green!60!black}{\uparrow}_{0.49}$ & 0.26\,$\color{green!60!black}{\uparrow}_{0.25}$ \\
    50 & 0.52\,$\color{green!60!black}{\uparrow}_{0.49}$ & 0.27\,$\color{green!60!black}{\uparrow}_{0.26}$ \\
    60 & \textbf{0.54}\,$\color{green!60!black}{\uparrow}_{0.51}$ & \textbf{0.31}\,$\color{green!60!black}{\uparrow}_{0.30}$ \\
    \bottomrule
  \end{tabularx}
  \vspace{-0.8em}
\end{wraptable}

Beyond PSNR and SSIM, character-level accuracy is a more direct metric for terminal rendering.
Character-level accuracy requires explicit pixel-to-text correspondence.
For CLIGen (Clean), we apply Tesseract to five uniformly sampled (ground-truth, generated) frame pairs per video and normalize whitespace.
We then compute two metrics (full protocol in Appendix~\ref{appendix:pipeline}).
Character accuracy uses the Levenshtein distance between concatenated ground-truth and generated texts.
Exact-line accuracy measures the fraction of ground-truth lines whose normalized content exactly matches the prediction at the same line index.

Table~\ref{tab:cligen-ocr} shows that our models achieve substantial text rendering accuracy under this protocol.
Character accuracy increases from 0.03 at initialization to 0.54 at 60k steps, with exact-line matches reaching 0.31 (0.26 by 40k).
Most gains occur within the first 40k steps, followed by smaller refinements thereafter.
These OCR-based metrics capture properties beyond perceptual similarity. 
Accurately generating terminal characters requires modeling text structure, font rendering, and spatial relationships.
These are core competencies for interactive neural computer systems.
This level of character-level precision is a step toward usable, not just plausible, terminal interfaces.
At the same time, we interpret this result primarily as evidence of interface fidelity, while routine reuse and native symbolic computation remain separate questions.

\exptitle[\cligenCleanLogo]{5}{Does this NC instantiation show native CLI reasoning?}

\providecommand{\wanicon}{\raisebox{-0.25ex}{\includegraphics[height=1.00em]{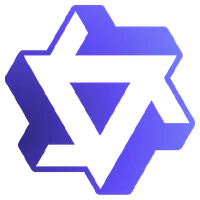}}}
\providecommand{\veoicon}{\raisebox{-0.25ex}{\includegraphics[height=1.00em]{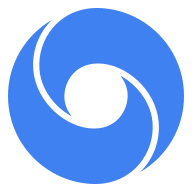}}}
\providecommand{\soraicon}{\raisebox{-0.25ex}{\includegraphics[height=1.00em]{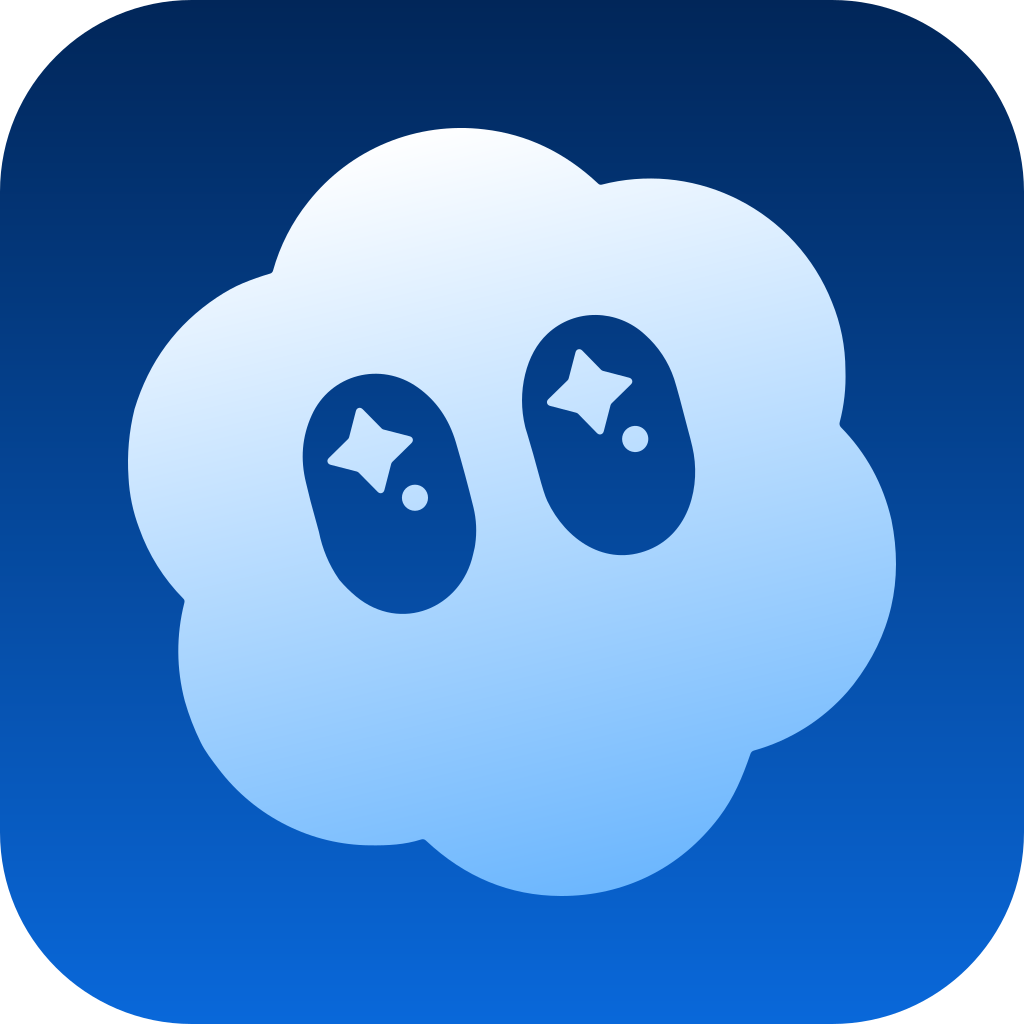}}}
\begin{wraptable}{r}{0.50\linewidth}
  \vspace{-1.0em}
  \captionsetup{type=table,width=\linewidth}
  \centering
  \small
  \rowcolors{2}{tablewarmrow}{white}
  \caption{Arithmetic probe accuracy (100 problems sampled from a 1{,}000-problem held-out pool).}
  \label{tab:cligen-arith}
  \setlength{\tabcolsep}{8pt}
  \begin{tabularx}{\linewidth}{l >{\centering\arraybackslash}X}
    \toprule
    \rowcolor{tablewarmhead} \textbf{Model} & \textbf{Accuracy} \\
    \wanicon\;Wan2.1 & 0\% \\
    \cligenCleanLogo{}\;\nccligen{} & 4\% \\
    \veoicon\;Veo3.1 & 2\% \\
    \soraicon\;Sora2 & \textbf{71}\% \\
    \bottomrule
  \end{tabularx}
   \vspace{-1.0em}
\end{wraptable}
We also probe symbolic computation with CLI arithmetic tasks.
These tasks are a sharp stress test for symbolic reliability: humans answer them instantly, yet current NC instantiations often fail on seemingly simple symbolic operations.

Our arithmetic probe presents basic mathematical operations through terminal interactions.
We reserve a held-out pool of 1{,}000 math problems and randomly sample 100 problems as the final evaluation set.
Table~\ref{tab:cligen-arith} shows that current video models, including this NC instantiation, struggle on these symbolic tasks.
Wan2.1 achieves 0\% accuracy, our \nccligen{} model reaches 4\%, and Veo3.1 manages 2\%—all far below human-level performance on these fundamental tasks.
These results contrast with common claims of strong symbolic reasoning in current video models.
Sora2's 71\% accuracy is a notable outlier and may reflect system-level advantages or additional training beyond our current setup.
Overall, native symbolic reasoning remains an open challenge for current video-based NC~instantiations.
Accordingly, arithmetic probes in this paper serve as a targeted test of symbolic stability under the current prototype substrate.

The poor arithmetic-probe performance in Table~\ref{tab:cligen-arith} raises a key question.
Does this prototype require specialized reinforcement learning to achieve reliable symbolic computation, or can stronger conditioning substantially narrow this gap?

\exptitle[\cligenCleanLogo]{6}{Does this NC instantiation require RL for symbolic probes?}

As shown in Figure~\ref{fig:cligen-exp6}, \nccligen{} accuracy on CLIGen (Clean) arithmetic tasks rises from 4\% to 83\% under reprompting.
This suggests that system-level conditioning can be an effective first lever for improving performance on symbolic probes.
It is complementary to (rather than strictly requiring) RL-based training pipelines.
More generally, the success of reprompting highlights how sensitive symbolic-probe outcomes are to the conditioning interface.
Much of the apparent ``reasoning'' gain can come from better specification and instruction-following rather than new native computation.
For the arithmetic subset, we include the correct answer explicitly in roughly half of the training captions to encourage reliable rendering of the output string.
Because reprompting can similarly provide stronger hints (or even outsource computation to an external text system), we interpret the gain primarily as evidence of steerability.
It also shows faithful rendering of conditioned symbolic content.
We do not treat it as a clean demonstration that the NC backbone performs arithmetic internally.

\begin{table}[h]
  \captionsetup{type=table,width=\linewidth}
  \centering
  \small
  \rowcolors{2}{tablewarmrow}{white}
\caption{Hypotheses for Sora2's advantage.}
 \vspace{-4pt}
  \label{tab:sora-leads}
  \setlength{\tabcolsep}{6pt}
  \begin{tabularx}{\linewidth}{l X}
    \toprule
    \rowcolor{tablewarmhead} \textbf{Factor} & \textbf{Implication} \\
    \tikz[baseline=-0.35ex]{\node[fill=metablue, text=white, circle, inner sep=1.2pt, font=\scriptsize\bfseries] {1};} Stronger base + similar data & Higher intrinsic arithmetic; symbolic capability may be baked in \\
    \tikz[baseline=-0.35ex]{\node[fill=orange!85!red, text=white, circle, inner sep=1.2pt, font=\scriptsize\bfseries] {2};} Additional RL training & Reward shaping teaches math beyond diffusion; could transfer to CLI \\
    \tikz[baseline=-0.35ex]{\node[fill=teal!70!black, text=white, circle, inner sep=1.2pt, font=\scriptsize\bfseries] {3};} System-level reprompt/recaption & LLM computes answers; strong conditioning drives generation \\
    \bottomrule
  \end{tabularx}
\end{table}

The evidence supports system-level conditioning as a practical path forward for this NC instantiation.
Among the three hypotheses for improving arithmetic-probe performance—stronger base models, reinforcement learning, or enhanced conditioning—our results most strongly favor the third approach.
The gain from reprompting (4\%$\rightarrow$83\%), achieved without modifying the underlying NC backbone, is substantial.
It shows that measured ``reasoning'' on these probes is highly sensitive to specification and conditioning.
We therefore do not treat it as direct evidence of native arithmetic inside the NC backbone.

In our setting, strategic conditioning yields larger symbolic-probe gains than the RL pipeline we tested.
Evaluations should therefore distinguish native computation from conditioning-assisted performance when assessing reasoning capabilities in current video-based NC instantiations.

\subsubsection{Visualizations}

\noindent\textbf{(1) \cligenGeneralData{} visualizations.}
Qualitative samples highlight the breadth of real-world terminal dynamics captured in CLIGen (General): ANSI escape sequences that repaint regions with changing foreground/background colors, incremental command entry with syntax highlighting and cursor edits, classic shell prompts and system outputs, long-running jobs with rapidly scrolling and color-coded package logs, full-screen TUIs such as partition editors, and progress dashboards with updating bars, counts, and ETAs.
These traces emphasize that ``looking correct'' requires maintaining terminal geometry, palette transitions, and cursor state frame-by-frame.

\noindent\textbf{(2) \cligenCleanData{} REPL visualizations.}
In contrast to open-world traces, CLIGen (Clean) REPL samples are scripted and temporally well-paced (Figures~\ref{fig:cligen-clean-viz-a}--\ref{fig:cligen-clean-viz-d}; additional examples are in Appendix~\ref{appendix:cligen-samples}).
Each sample includes an explicit action trace (e.g., \texttt{Sleep}, \texttt{Type}, \texttt{Enter}, arrow keys, \texttt{Hide}) alongside rendered terminal frames, making the action-to-pixel link visually unambiguous.
The key insight is that these scripted traces isolate rendering-and-control errors from semantic ambiguity: with explicit actions, failures are dominated by low-level mechanics (cursor placement, character edits, monospace alignment, line breaks, temporal consistency).

\noindent\textbf{(3) \cligenCleanData{} math visualizations.}
Figures~\ref{fig:cligen-clean-math-comp-a}--\ref{fig:cligen-clean-math-comp-c} compare math REPL rollouts, and Figures~\ref{fig:cligen-clean-math-rep-a}--\ref{fig:cligen-clean-math-rep-c} show reprompting cases.
Together they highlight why arithmetic probes should separate native computation from answer-conditioned rendering.
All full-resolution pages are in Appendix~\ref{appendix:vis}; below we keep clickable thumbnails at the original location for quick navigation.

\newcommand{\CliThumbPageHeader}[1]{%
  \par\noindent\makebox[\linewidth][c]{%
    \fcolorbox{metablue!45}{metablue!5}{%
      \begin{minipage}{0.94\linewidth}
        \centering
        \vspace{0.22em}
        {\large\bfseries #1}\par
        \vspace{0.1em}
        {\scriptsize\textsf{\color{metablue!85!black}Click any thumbnail to jump to its full-resolution page in Appendix}}\par
        \vspace{0.18em}
      \end{minipage}
    }%
  }\par\vspace{0.58em}%
}
\newcommand{\CliThumbPageHeaderSimple}[1]{%
  \par\noindent\makebox[\linewidth][c]{%
    \fcolorbox{metablue!45}{metablue!5}{%
      \begin{minipage}{0.94\linewidth}
        \centering
        \vspace{0.22em}
        {\large\bfseries #1}\par
        \vspace{0.16em}
      \end{minipage}
    }%
  }\par\vspace{0.58em}%
}
\newcommand{\CliThumbPageHeaderPlain}[1]{%
  {\par\noindent\centering\Large\bfseries #1\par}
  {\par\noindent\centering%
    \begin{tcolorbox}[
      enhanced,
      boxrule=0.85pt,
      colframe=orange!80!black,
      colback=yellow!14,
      arc=5.5pt,
      left=6pt,right=6pt,top=2.5pt,bottom=2.5pt,
      width=0.78\linewidth
    ]
      \centering\small\bfseries\sffamily\color{orange!85!black}
      Click any thumbnail to jump to its full-resolution page in Appendix
    \end{tcolorbox}\par
  }
  \vspace{0.26em}%
}
\newcommand{\CliThumbPageFrame}{%
  \begin{tikzpicture}[remember picture,overlay]
    \draw[
      draw=black,
      line width=1.9pt,
      rounded corners=14pt
    ]
      ([xshift=0.48cm,yshift=-0.48cm]current page.north west)
      rectangle
      ([xshift=-0.48cm,yshift=0.48cm]current page.south east);
  \end{tikzpicture}%
}
\newcommand{\CliThumbCard}[4]{%
  \par\noindent\makebox[\linewidth][c]{%
    \hyperref[#1]{%
      \begin{tikzpicture}
        \node[inner sep=0pt] (img) {\includegraphics[width=#4\linewidth,keepaspectratio]{#2}};
        \node[
          anchor=south east,
          xshift=-0.38em,
          yshift=0.26em,
          fill=white,
          fill opacity=0.93,
          text opacity=1,
          draw=metablue!50,
          rounded corners=1.5pt,
          inner xsep=0.45em,
          inner ysep=0.2em,
          minimum width=8.4em,
          align=center,
          font=\scriptsize\bfseries,
          text=metablue!90!black
        ] at (img.south east) {#3};
      \end{tikzpicture}%
    }%
  }\par\vspace{0.28em}%
}
\newcommand{\CliThumb}[4]{\CliThumbCard{#1}{#2}{#3}{#4}}

\clearpage
\newgeometry{top=0.9cm,bottom=1.35cm,left=0.9cm,right=0.9cm,includefoot,footskip=18pt}
\thispagestyle{plain}
\hypertarget{main-cligen-vis-thumbs}{}
\hypertarget{main-cligen-vis-thumbs-p1}{}
\hypertarget{main-cligen-vis-thumbs-p2}{}
\CliThumbPageFrame
\CliThumbPageHeaderPlain{CLIGen Visualization Thumbnails}
\noindent\begin{tabular}{@{}p{0.495\linewidth}!{\color{metablue!55}\vrule width 0.8pt}p{0.495\linewidth}@{}}
\begin{minipage}[t]{\linewidth}
  \vspace*{0pt}
  \centering
  {\small\bfseries \cligenGeneralLogo{}\ CLIGen (General) Visualizations}\par
  \vspace{0.14em}
  \CliThumb{fig:cligen-general-viz-1}{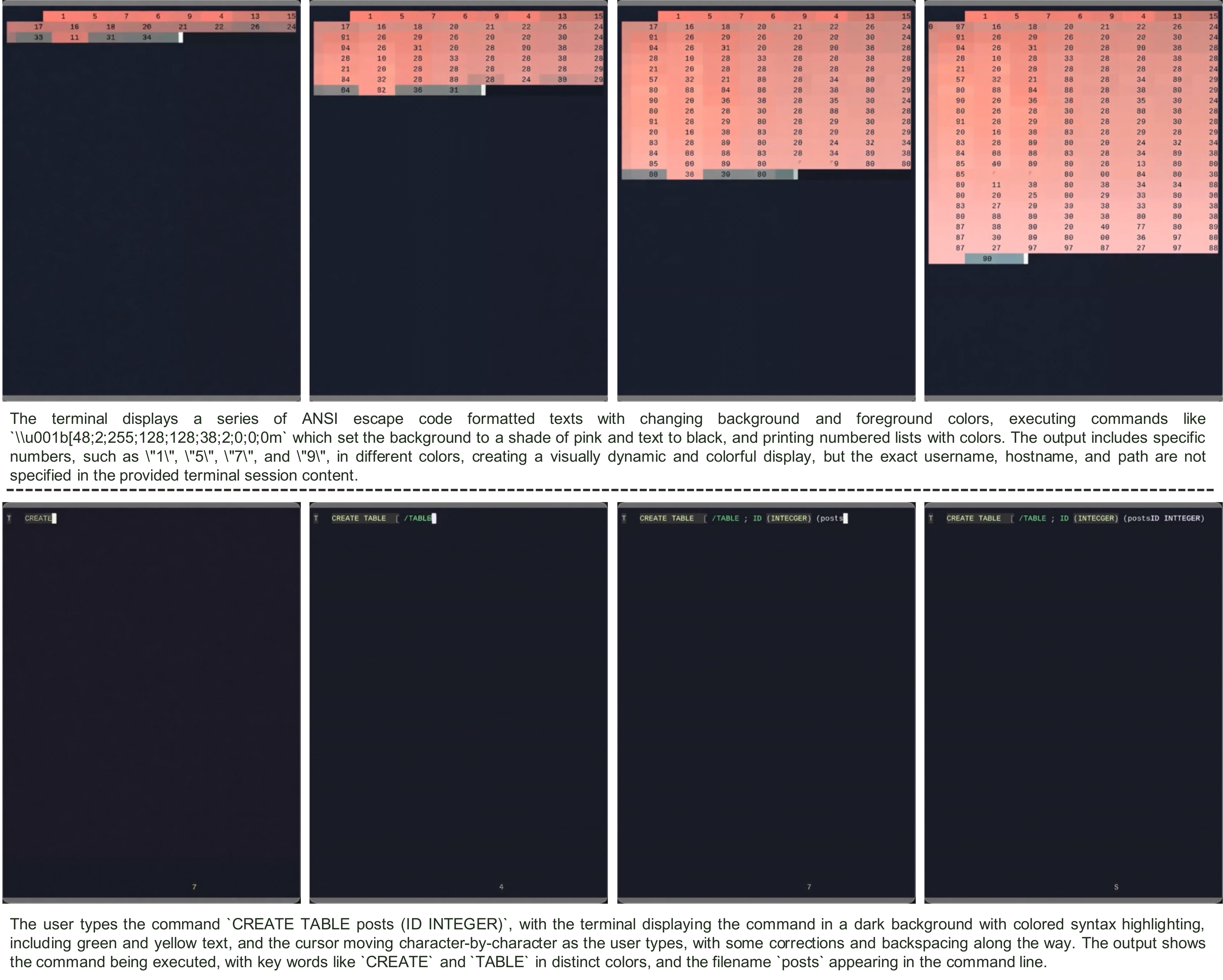}{\cligenGeneralLogo{}\ Samples A}{0.72}
  \CliThumb{fig:cligen-general-viz-2}{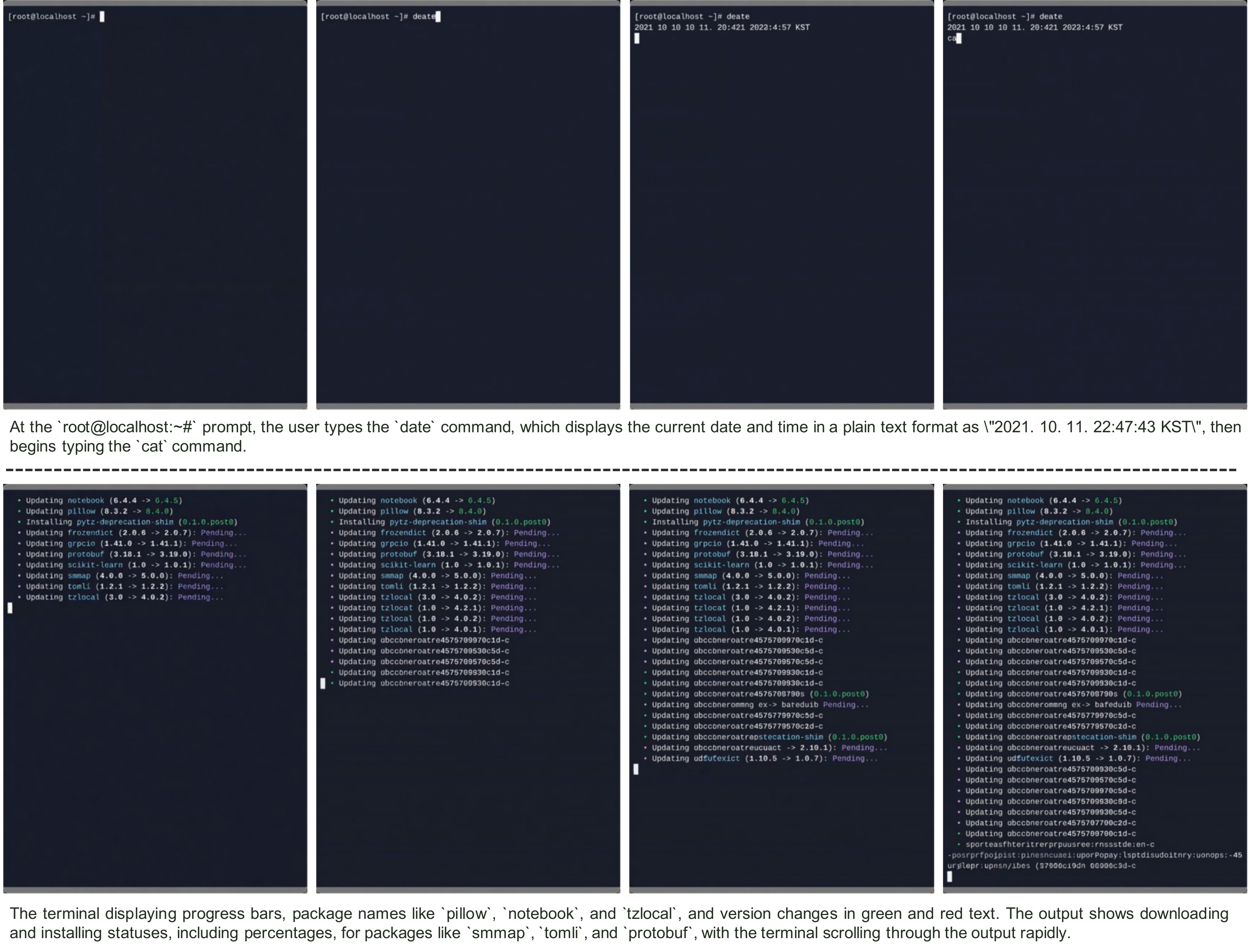}{\cligenGeneralLogo{}\ Samples B}{0.72}
  \CliThumb{fig:cligen-general-viz-3}{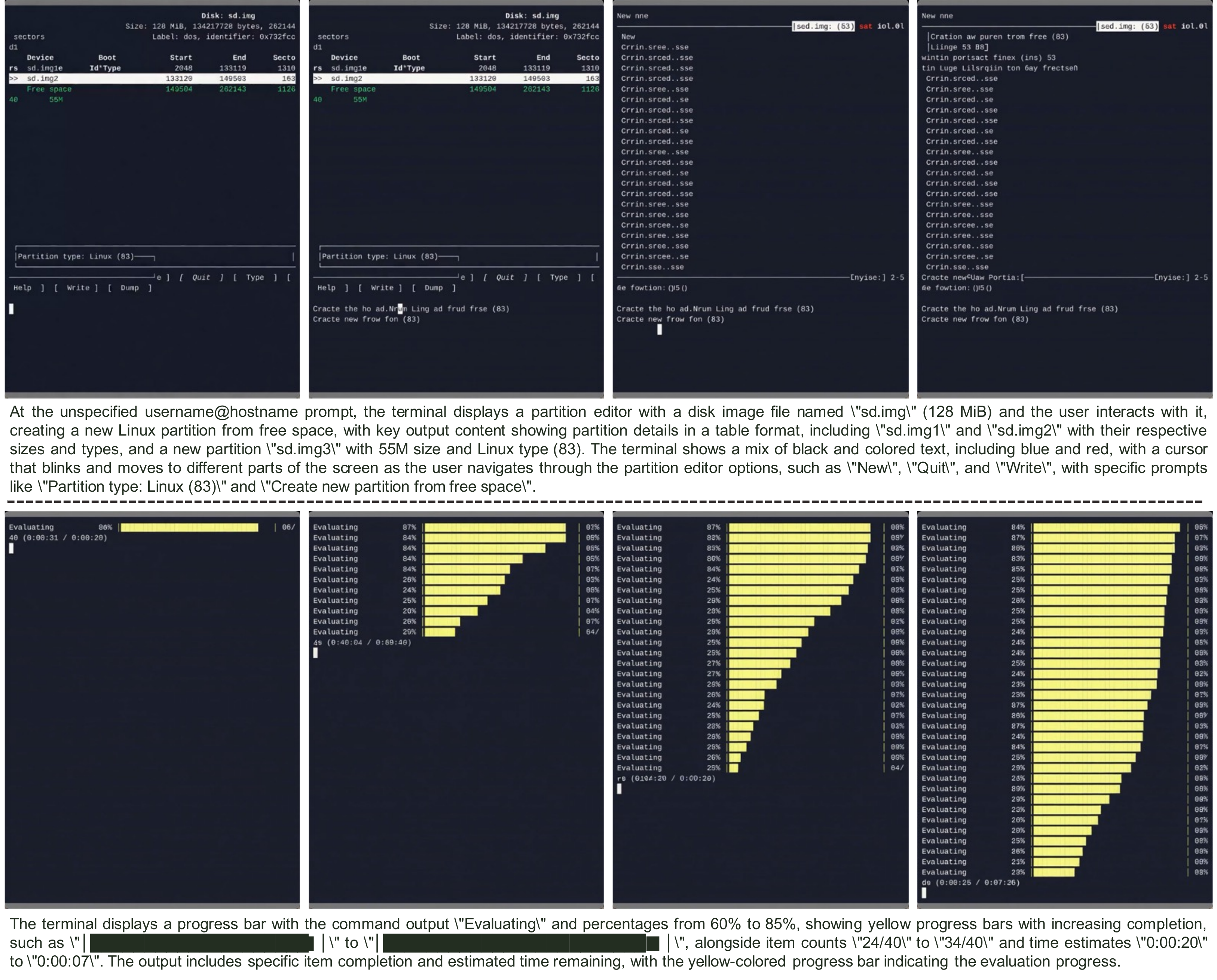}{\cligenGeneralLogo{}\ Samples C}{0.72}
\end{minipage}
&
\begin{minipage}[t]{\linewidth}
  \vspace*{0pt}
  \centering
  {\small\bfseries \cligenCleanLogo{}\ CLIGen (Clean) Visualizations}\par
  \vspace{0.14em}
  \CliThumb{fig:cligen-clean-viz-a}{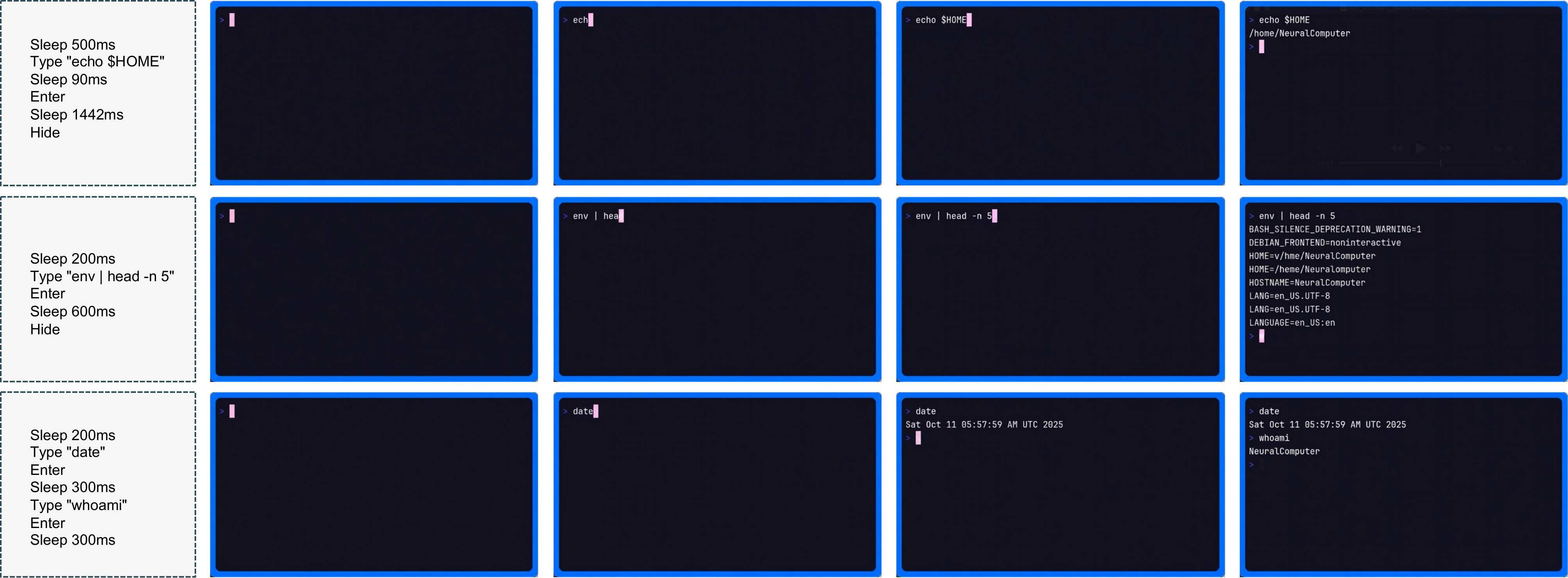}{\cligenCleanLogo{}\ Samples A}{0.97}
  \CliThumb{fig:cligen-clean-viz-b}{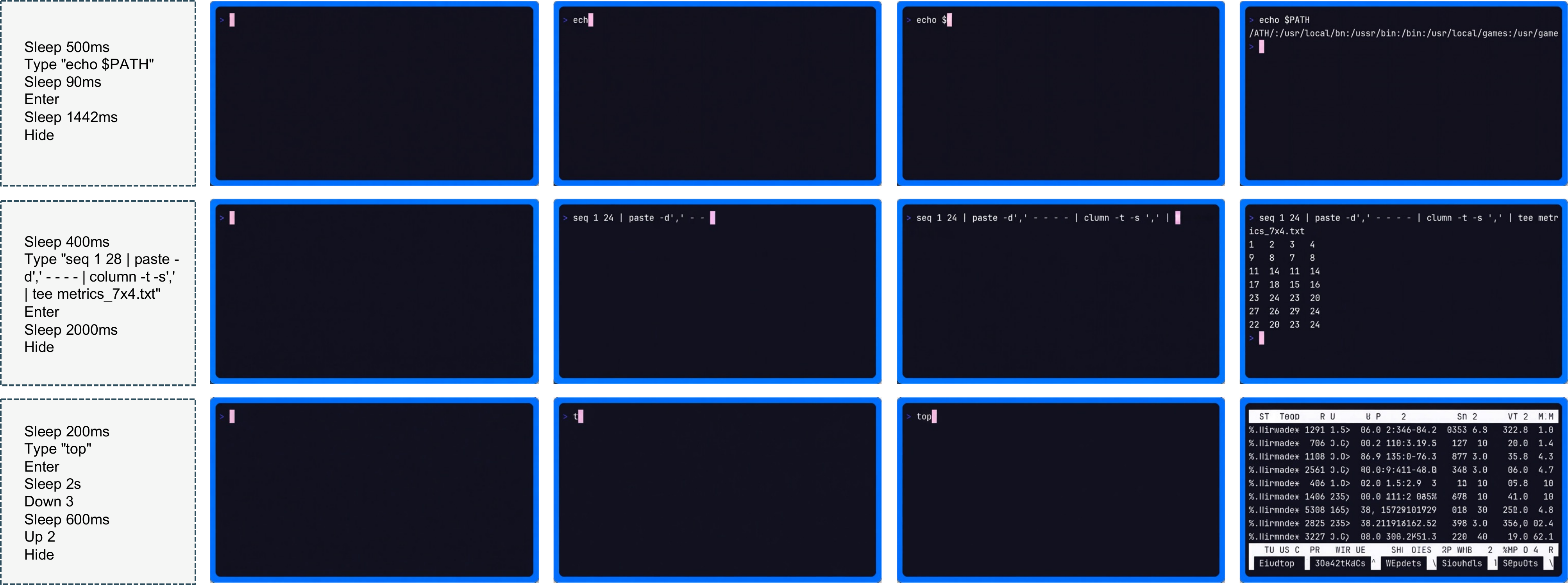}{\cligenCleanLogo{}\ Samples B}{0.97}
  \CliThumb{fig:cligen-clean-viz-c}{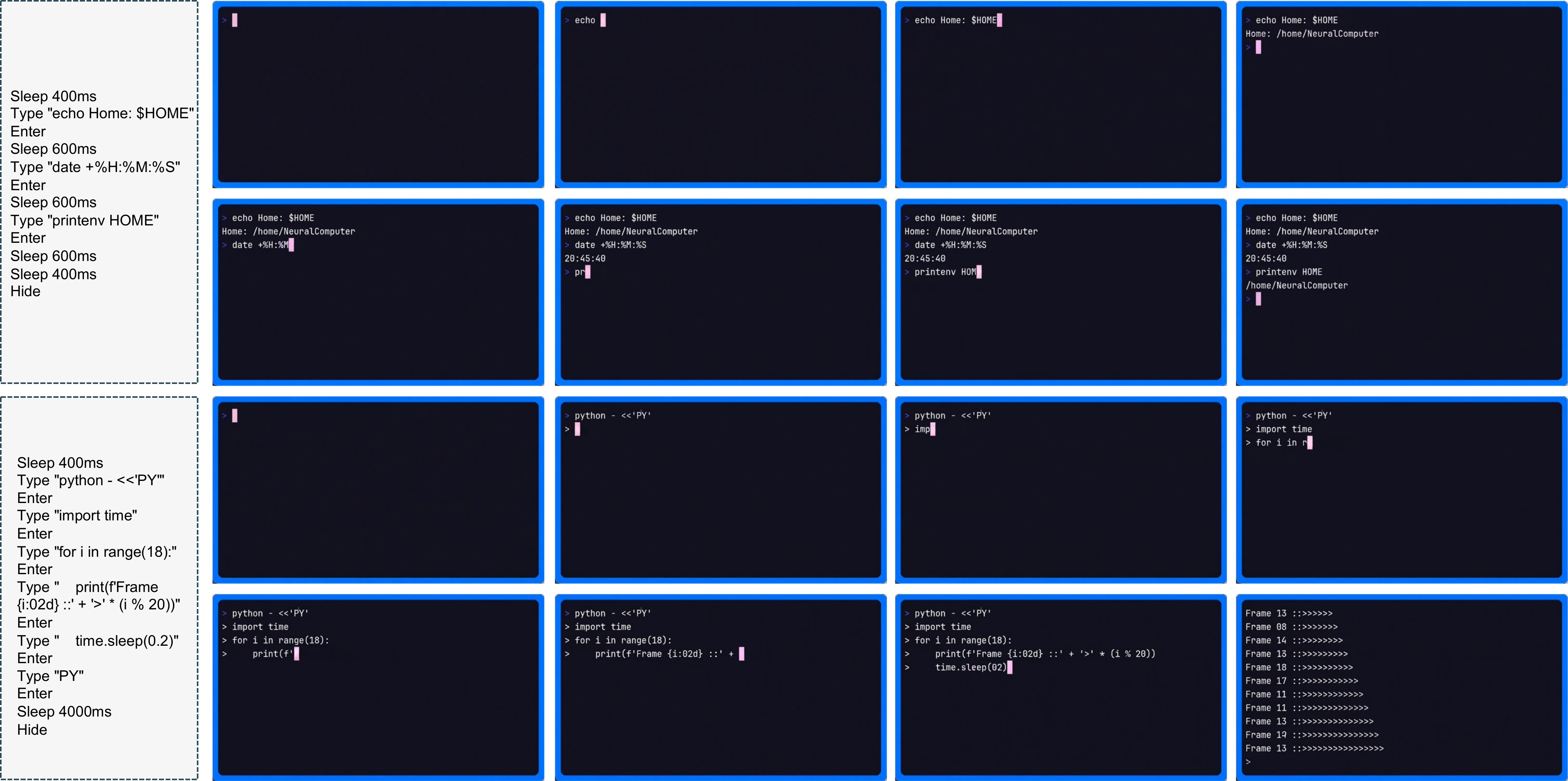}{\cligenCleanLogo{}\ Samples C}{0.97}
  \CliThumb{fig:cligen-clean-viz-d}{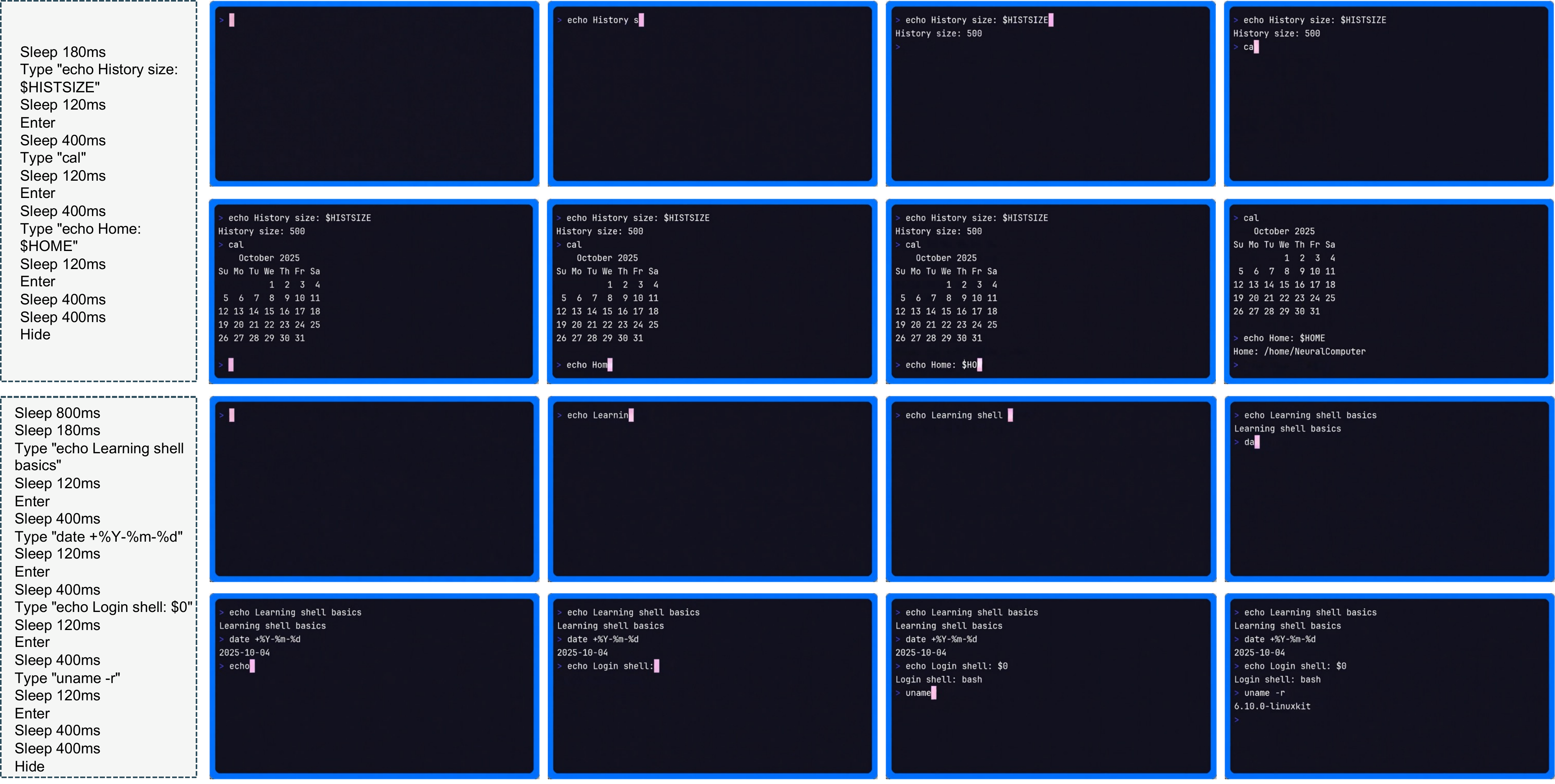}{\cligenCleanLogo{}\ Samples D}{0.97}
\end{minipage}
\end{tabular}
\restoregeometry
\par\justifying

\clearpage
\newgeometry{top=0.9cm,bottom=1.35cm,left=0.9cm,right=0.9cm,includefoot,footskip=18pt}
\thispagestyle{plain}
\hypertarget{main-cligen-vis-thumbs-p3}{}
\hypertarget{main-cligen-vis-thumbs-p4}{}
\CliThumbPageFrame
\CliThumbPageHeaderPlain{CLIGen Visualization Thumbnails}
\noindent\begin{tabular}{@{}p{0.495\linewidth}!{\color{metablue!55}\vrule width 0.8pt}p{0.495\linewidth}@{}}
\begin{minipage}[t]{\linewidth}
  \vspace*{0pt}
  \centering
  {\small\bfseries \cligenCleanLogo{}\ CLIGen (Clean) Math Comparison}\par
  \vspace{0.14em}
  \CliThumb{fig:cligen-clean-math-comp-a}{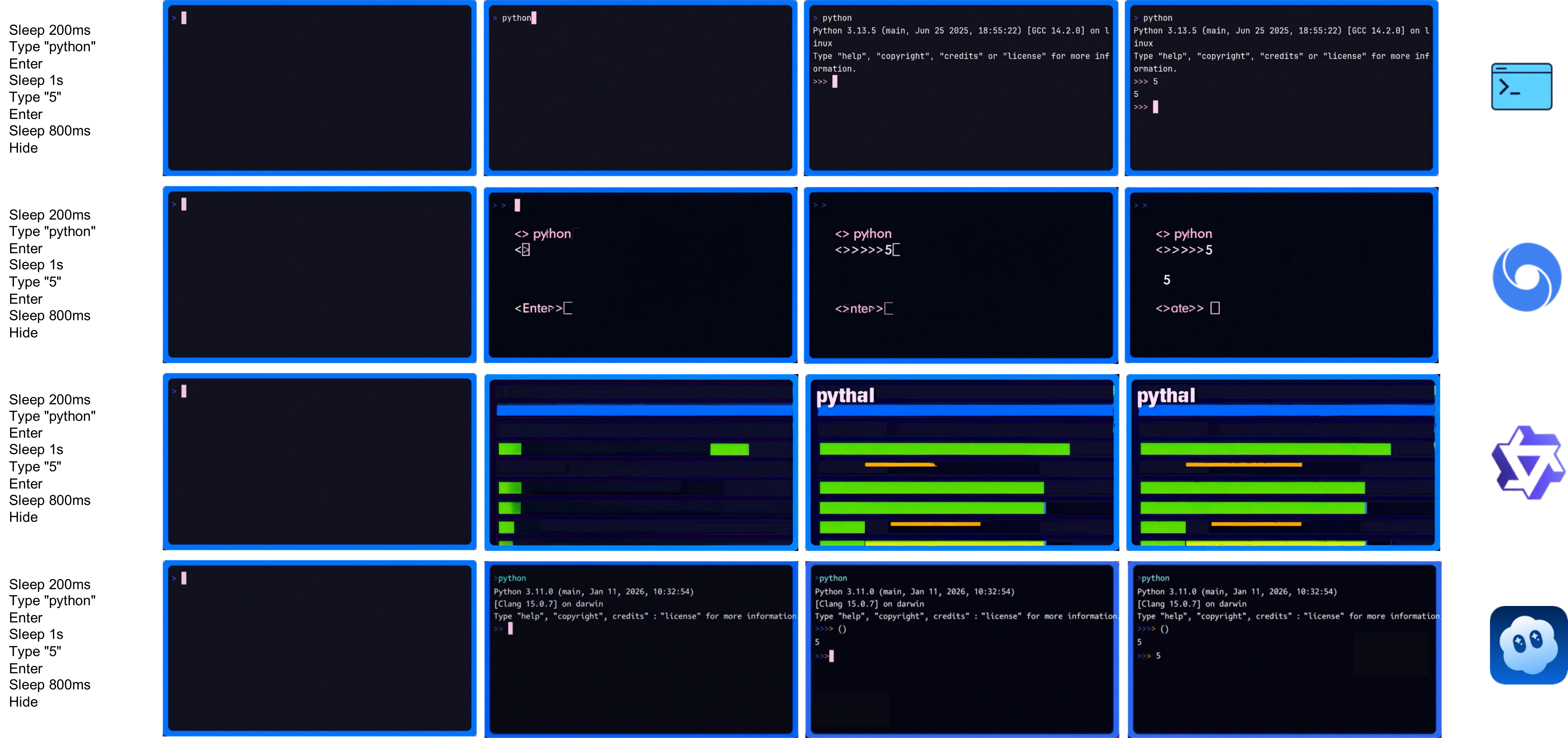}{\cligenCleanLogo{}\ Samples A}{0.76}
  \CliThumb{fig:cligen-clean-math-comp-b}{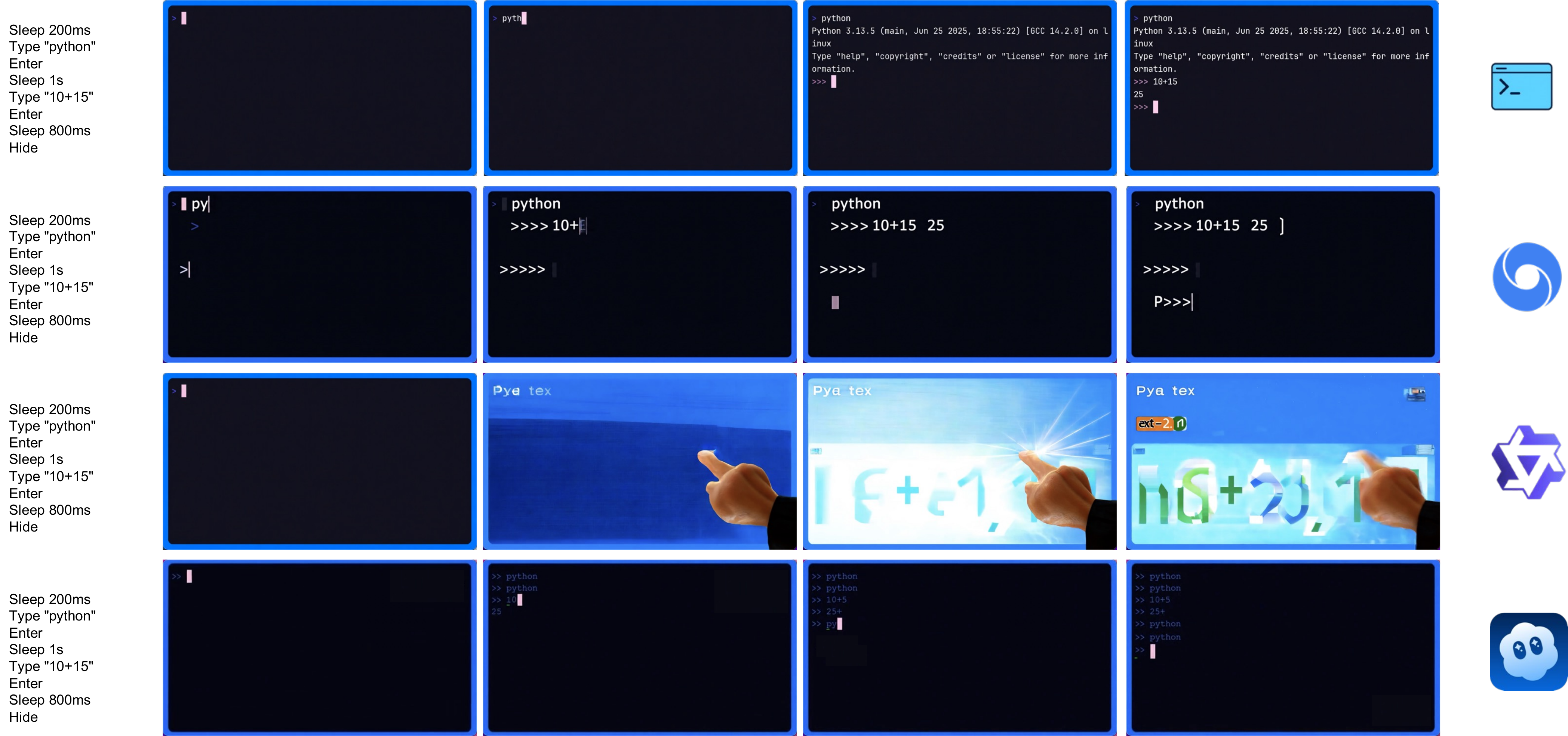}{\cligenCleanLogo{}\ Samples B}{0.76}
  \CliThumb{fig:cligen-clean-math-comp-c}{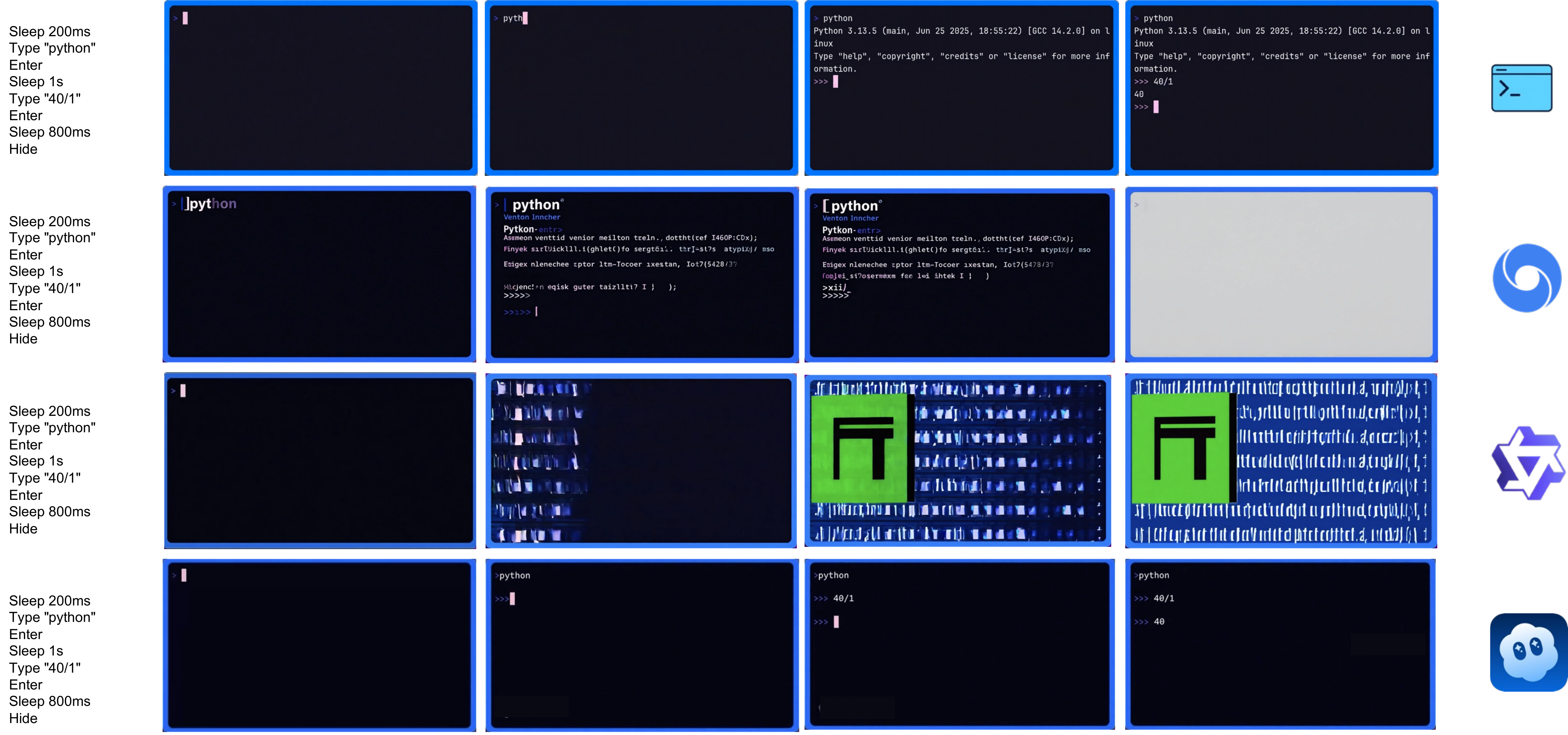}{\cligenCleanLogo{}\ Samples C}{0.76}
\end{minipage}
&
\begin{minipage}[t]{\linewidth}
  \vspace*{0pt}
  \centering
  {\small\bfseries \cligenCleanLogo{}\ CLIGen (Clean) Math Reprompting}\par
  \vspace{0.14em}
  \CliThumb{fig:cligen-clean-math-rep-a}{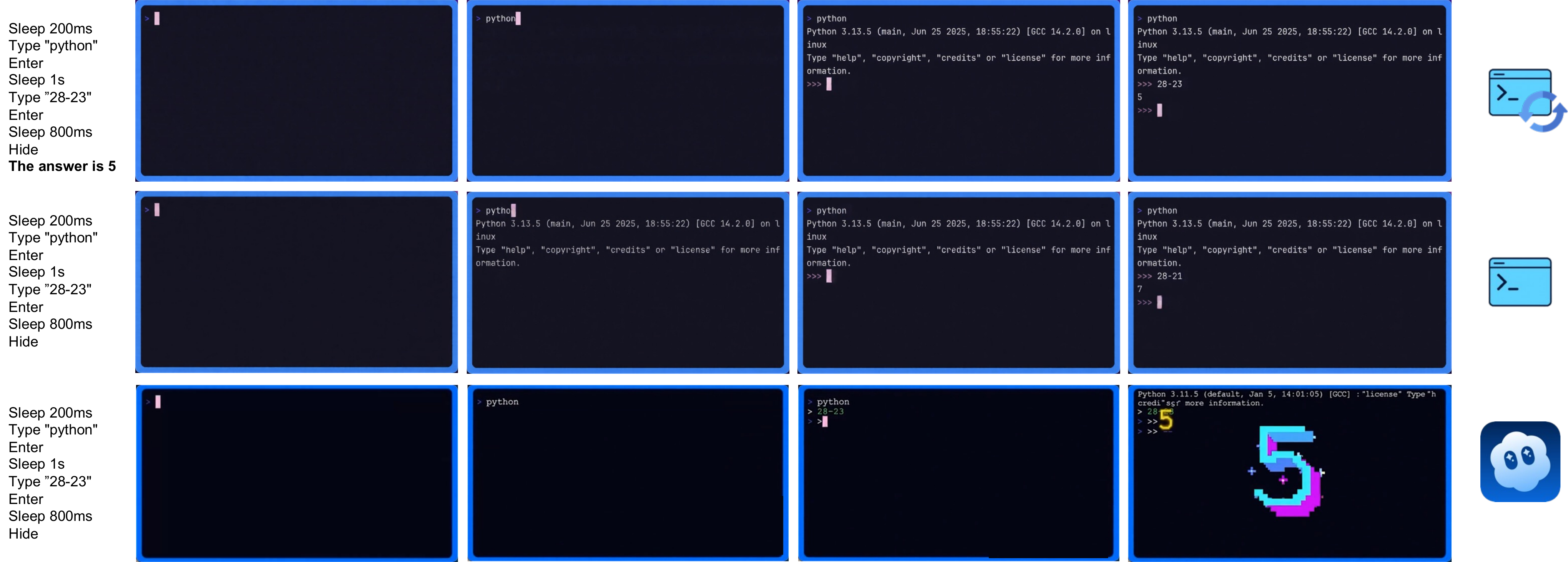}{\cligenCleanLogo{}\ Samples A}{0.98}
  \CliThumb{fig:cligen-clean-math-rep-b}{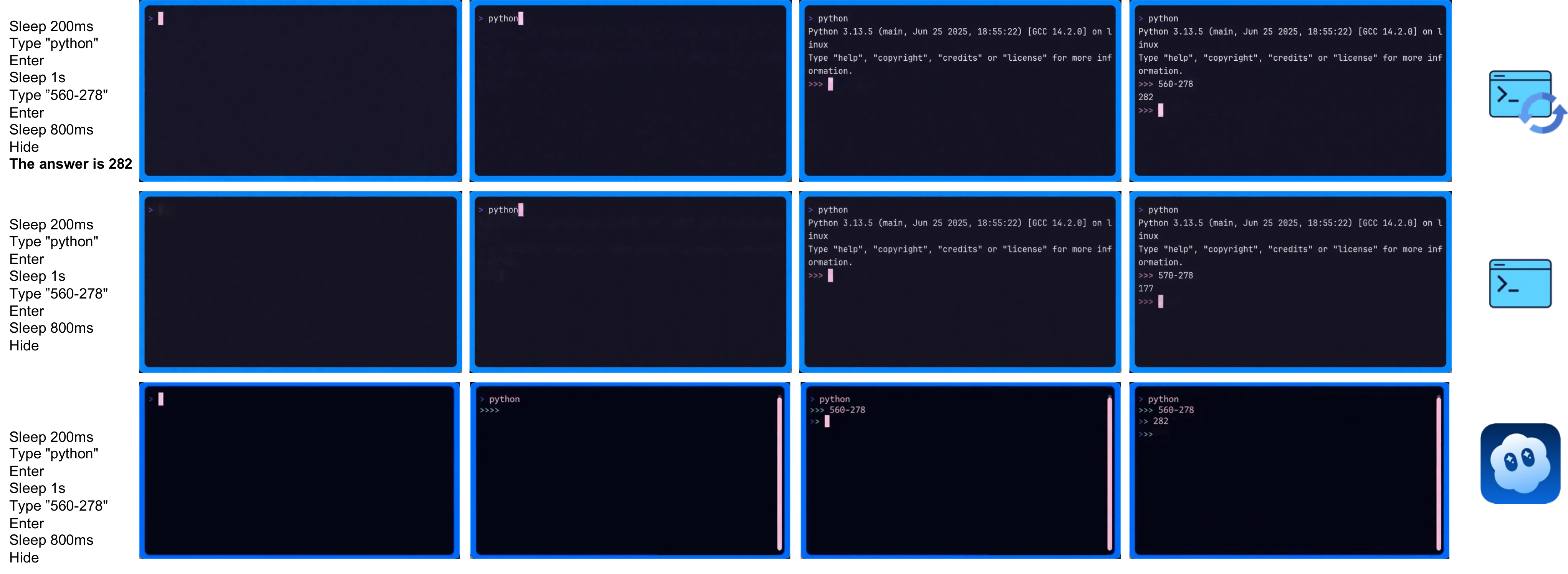}{\cligenCleanLogo{}\ Samples B}{0.98}
  \CliThumb{fig:cligen-clean-math-rep-c}{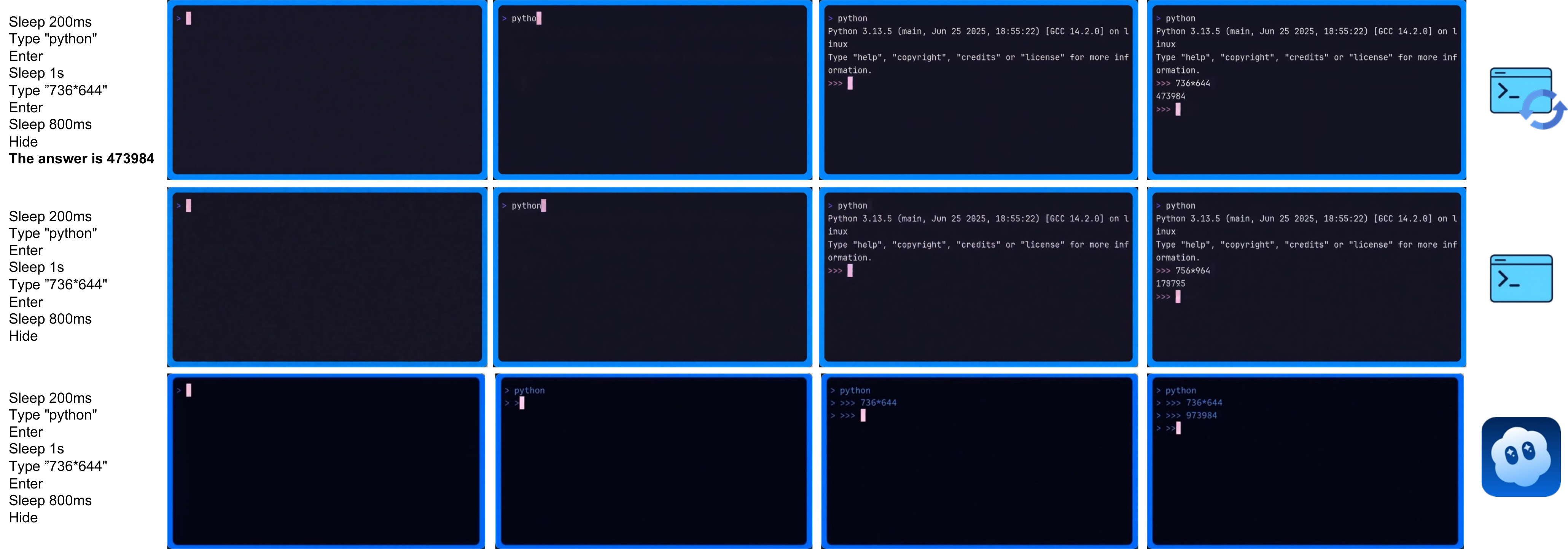}{\cligenCleanLogo{}\ Samples C}{0.98}
\end{minipage}
\end{tabular}
\restoregeometry
\par\justifying

%% file: section/impl_gui.tex
\justifying
We also instantiate the NC abstraction in interactive graphical user interface (GUI) environments with \ncguiworld{}.
In this setting, fine-grained action control is essential: GUI interaction requires precise cursor tracking, timely click feedback, and robustness to rapidly changing interface states.
We model each interaction as a synchronized sequence of RGB frames $x_t$ and input events $u_t$ (mouse and keyboard).
The latent video state maintains interface context across frames, while temporally aligned action inputs provide control signals designed to preserve pixel-level correspondence between user actions and visual changes.

\subsubsection{Data pipeline}
\label{section:impl-guiworld-data-pipeline}
\begin{wraptable}{r}{0.48\linewidth}
  \vspace{-1.7em}
  \captionsetup{type=table,width=\linewidth}
  \centering
  \small
  \rowcolors{2}{tablewarmrow}{white}
  \caption{Cursor/action statistics.}
  \label{tab:gui-stats}
  \setlength{\tabcolsep}{7pt}
  \begin{tabularx}{\linewidth}{l >{\centering\arraybackslash}X >{\centering\arraybackslash}X}
    \toprule
    \rowcolor{tablewarmhead}\textbf{Split} & \textbf{Avg. cursor speed (px/frame)} & \textbf{Actions / sec} \\
    Random Slow & 1.51 & 1.58 \\
    Random Fast & 195.15 & 4.18 \\
    CUA (supervised) & 3.79 & 0.10 \\
    \bottomrule
  \end{tabularx}
  \vspace{-0.8em}
\end{wraptable}

The dataset includes two styles of random interaction, which we refer to collectively as GUIWorld Random: ``Random Slow'' and ``Random Fast'', plus a smaller set of supervised trajectories collected with Claude CUA~\citep{claudeComputerTool}.
Random Slow (approximately 1{,}000 hours) contains longer pauses, idle gaps, and deliberate cursor movements, which can expose cursor drift after extended inactivity.
Random Fast (approximately 400 hours) features denser cursor motion and typing bursts, stressing acceleration dynamics and hover timing.
The supervised trajectories are approximately 110 hours. 
These goal-directed traces provide higher-signal action--response pairs without overwhelming the exploration data.
Table~\ref{tab:gui-stats} summarizes cursor and action statistics across splits; in the collected CUA trajectories, action density is lower due to latency introduced by Claude’s tool API between successive steps.

For GUIWorld Random, our data-collection pipeline builds on the NeuralOS setup~\citep{rivard2025neuralos}: data is collected inside an Ubuntu~22.04 container running XFCE4 (Arc-Dark theme, Papirus icons) on a fixed 1024$\times$768 virtual display at 15~FPS, and we adopt the NeuralOS cursor-position representation, with task-specific adaptations for our setting.
Within this same desktop environment, we develop a separate CUA data-collection pipeline tailored to the needs of this work.
We render the display with Xvfb and interact through a VNC/noVNC stack.
The desktop pins a small open-source app set to launchers.
It includes Firefox ESR, GIMP, VLC, VS~Code, Calculator, Terminal, the file manager, and the Mahjongg game, matching the environment shown in our recordings.
Screen capture uses \texttt{mss} and \texttt{ffmpeg} with cursor overlays, and actions are replayed and logged via \texttt{xdotool}.
We keep the recorded discontinuities and interface latency intact rather than smoothing them.
In dataset packaging, we store both raw-action and meta-action views for modeling.
This lets us train either the raw-action or meta-action encoder under the same loss stack\footnote{Conversion details and alignment quality appear in Appendix~\ref{appendix:gui-actions}.}.

\subsubsection{Model architecture}

The GUIWorld architecture builds on the Wan2.1~\citep{wan2025wan} by incorporating explicit action-conditioning modules.
The central challenge is to align time-stamped user actions with generated frames and inject this information at the appropriate depth within the transformer.

Action features are encoded on-the-fly from frame-aligned mouse and keyboard signals (Section~\ref{section:impl-guiworld-data-pipeline}).
We aggregate them into latent-aligned embeddings that summarize recent action history at each diffusion step.
We evaluate two action encoders.
The \emph{raw-action encoder} (v1) preserves fine-grained mouse/keyboard event streams.
The \emph{meta-action encoder} (v2) abstracts interactions into coarse API-style categories (clicks, drags, scrolls, typing, shortcuts).
Both encoders use the same temporal alignment and are evaluated as separate ablations.
In our experiments, their differences in rendering fidelity and control behavior are modest\footnote{Appendix Table~\ref{tab:action-encoder-comparison} summarizes the representational differences.}.

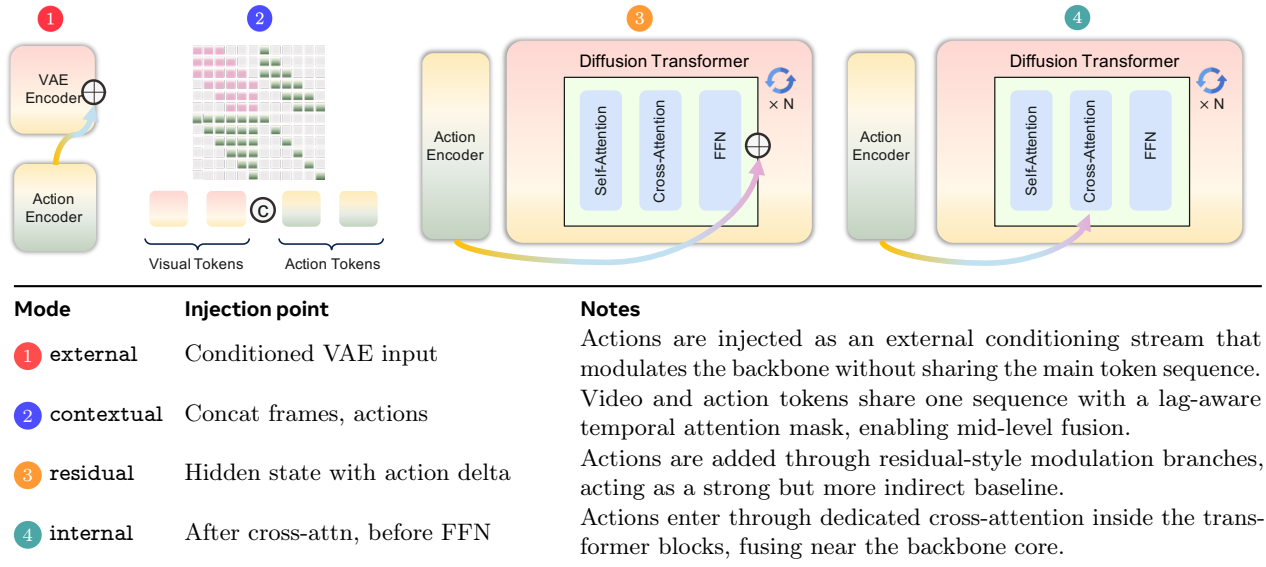
\begin{figure}[t]
  \centering
  \DeclareRobustCommand{\modeExt}{\tikz[baseline=-0.2em]{\node[fill=red!70, circle, inner sep=1.6pt, text=white, font=\scriptsize]{1};}}
  \DeclareRobustCommand{\modeCtx}{\tikz[baseline=-0.2em]{\node[fill=blue!70, circle, inner sep=1.6pt, text=white, font=\scriptsize]{2};}}
  \DeclareRobustCommand{\modeRes}{\tikz[baseline=-0.2em]{\node[fill=orange!80, circle, inner sep=1.6pt, text=white, font=\scriptsize]{3};}}
  \DeclareRobustCommand{\modeInt}{\tikz[baseline=-0.2em]{\node[fill=teal!70, circle, inner sep=1.6pt, text=white, font=\scriptsize]{4};}}
  \includegraphics[width=\linewidth]{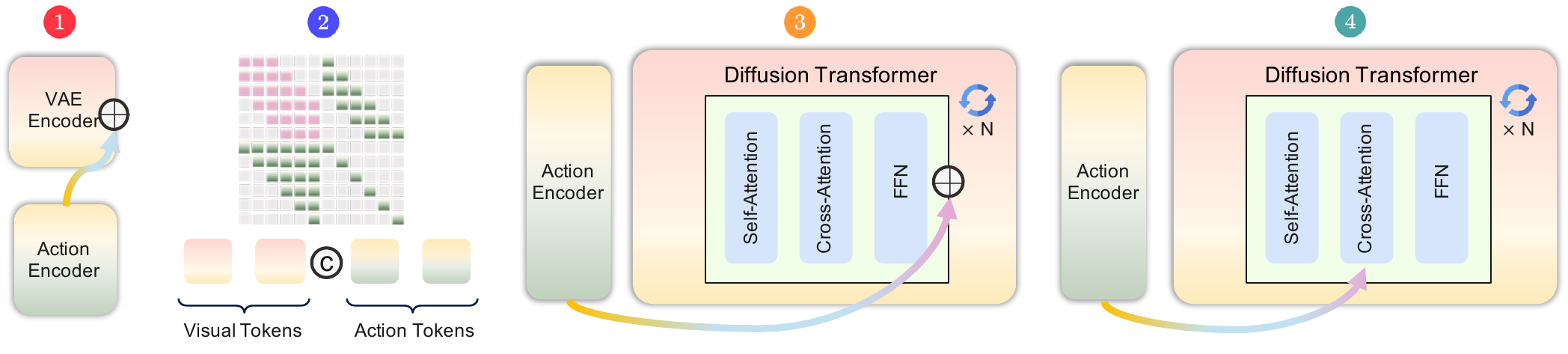}
  \vspace{0.2em}
  \small
  \setlength{\tabcolsep}{4pt}
  \renewcommand{\arraystretch}{1.05}
  \begin{tabularx}{\linewidth}{@{}l >{\raggedright\arraybackslash}p{0.30\linewidth} X@{}}
    \toprule
    \textbf{Mode} & \textbf{Injection point} & \textbf{Notes} \\
    \modeExt\ \texttt{external} & Conditioned VAE input & Actions are injected as an external conditioning stream that modulates the backbone without sharing the main token sequence. \\
    \modeCtx\ \texttt{contextual} & Concat frames, actions & Video and action tokens share one sequence with a lag-aware temporal attention mask, enabling mid-level fusion. \\
    \modeRes\ \texttt{residual} & Hidden state with action delta & Actions are added through residual-style modulation branches, acting as a strong but more indirect baseline. \\
    \modeInt\ \texttt{internal} & After cross-attn, before FFN & Actions enter through dedicated cross-attention inside the transformer blocks, fusing near the backbone core. \\
    \bottomrule
  \end{tabularx}
  \vspace{0.1em}
  \captionsetup{justification=raggedright,singlelinecheck=false}
\caption{\textbf{Four modes for injecting GUI actions into the diffusion transformer.}
\modeExt{} modulates VAE latents before the transformer;
\modeCtx{} adds action tokens alongside visual tokens;
\modeRes{} applies block-wise residual updates; and
\modeInt{} inserts an action cross-attention module inside transformer blocks.}

  \label{fig:action-injection}
\end{figure}

We inject action embeddings into the diffusion backbone in four ways (Figure~\ref{fig:action-injection}).
We study \texttt{external}, \texttt{contextual}, \texttt{residual}, and \texttt{internal} conditioning.
For the injection-scheme ablation, all four modes share the same meta-action encoder and temporal alignment.
They differ only in where the latent action features interact with the video latents and transformer blocks.
We compare raw-action vs.\ meta-action encoders separately in Table~\ref{tab:gui-encoding-ablation}.

\noindent\textbf{External conditioning.}
In the \texttt{external} mode, action information modulates the latent video sequence before the diffusion transformer.
Action features are applied as a pre-conditioning step at the model input, without introducing explicit action tokens or cross-attention inside the diffusion backbone.
As a result, action information enters only through the modified input latents; the diffusion backbone never attends directly to action tokens, so any action signal must be carried implicitly in $z'_{1:T}$.

Formally, given VAE latents $z_{1:T}$ and temporally aligned action features $u_{1:T}$, an external action module applies a small stack of temporal self-attention and action cross-attention layers.
This produces a residual update $\Delta z_{1:T}(u_{1:T})$.
The modified latents are
\[
  z'_{1:T} = z_{1:T} + \Delta z_{1:T}(u_{1:T}),
\]
and the diffusion transformer operates solely on $z'_{1:T}$.
The diffusion backbone remains unchanged, and action features are not exposed as explicit tokens within the transformer.

\noindent\textbf{Contextual conditioning.}
In the \texttt{contextual} mode, actions are represented as additional tokens and integrated directly into the transformer’s self-attention.
Similar token-based action representations have been explored in prior world models, including Gato~\citep{reed2022generalist} and World and Human Action Models~\citep{kanervisto2025world}.

The meta-action encoder produces latent-aligned action tokens $A \in \mathbb{R}^{L_a \times D}$.
We concatenate them with visual tokens $V \in \mathbb{R}^{L_v \times D}$ to form a joint sequence $[V; A]$.
Each transformer block applies self-attention over this combined sequence using a structured temporal mask (Appendix Figure~\ref{fig:contextual-mask}).
The mask enforces causal alignment: each frame token attends only to actions within a short past window, and each action token attends only to frames after a fixed temporal lag.
Through this masked joint attention, \texttt{contextual} conditioning fuses action and visual information within the transformer blocks.

\noindent\textbf{Residual conditioning.}
In the \texttt{residual} mode, the transformer block structure remains unchanged.
A lightweight action module attaches to a subset of layers as an external residual branch.
This follows the \texttt{residual} conditioning paradigm introduced by ControlNet~\citep{zhang2023adding}, while remaining modular and additive to the base diffusion backbone.

At each selected layer $l$, the transformer applies its standard sequence of self-attention, text or reference cross-attention, and feed-forward operations to produce hidden states $h^{(l)}$.
A separate action module then takes $h^{(l)}$ together with a local temporal window of latent action features and mouse trajectories.
It outputs a residual update $\Delta h^{(l)}(a,\text{mouse})$.
The updated hidden states are given by
\[
  \tilde{h}^{(l)} = h^{(l)} + \Delta h^{(l)}(a,\text{mouse}),
\]
which are passed to the subsequent transformer block.
In this formulation, \texttt{residual} conditioning injects action information through block-external residual branches.
It does not modify the internal computations of the transformer blocks themselves.

\noindent\textbf{Internal conditioning.}
In the \texttt{internal} mode, action conditioning is incorporated directly within the transformer blocks.
Related multi-stream world models have explored similar designs, including Matrix-Game-2~\citep{he2025matrix}.
Each selected block augments the standard attention stack with an additional action cross-attention sub-layer.
Specifically, the block applies self-attention, followed by cross-attention over text and reference features, and then a dedicated action cross-attention layer.
Keys and values are derived from latent action features (and, optionally, mouse inputs).

Given block input $h$, text or reference context $c$, and action latents $a$, the internal block computes
\[
  h' = \mathrm{FFN}\Big(
    h + \mathrm{CA}_{\text{text}}\big(\mathrm{SA}(h), c\big)
      + \mathrm{CA}_{\text{action}}(h, a)
  \Big),
\]
where $\mathrm{SA}$ denotes self-attention and $\mathrm{CA}_{\text{text}}$ and $\mathrm{CA}_{\text{action}}$ denote the text and action cross-attention modules applied in sequence.
As illustrated in Figure~\ref{fig:action-injection}, action features are injected directly into the block’s cross-attention stage.

In contrast to \texttt{residual} conditioning, \texttt{internal} conditioning integrates action information through a block-internal attention mechanism rather than an external residual branch.
This design mirrors the multi-stream injection strategy used in Matrix-Game-2~\citep{he2025matrix} and yields the best SSIM/FVD trade-off for fine-grained GUI interaction in our ablations.
In this setting, precise temporal alignment and spatial locality are critical.
Each conditioning mode (\texttt{external}, \texttt{contextual}, \texttt{residual}, and \texttt{internal}) is trained as a separate ablation, and no combinations are used.

\subsubsection{Implementation Details}
We train one model per injection mode (\texttt{external}, \texttt{contextual}, \texttt{residual}, \texttt{internal}), keeping the backbone and all non-action components fixed.
Each run lasts about 64k steps.
We tune only the action encoder and learning-rate schedule.
Training optimizes the diffusion loss together with a small temporal contrastive loss that aligns frame features with action and mouse embeddings (Appendix~\ref{appendix:gui-actions}).
Runs use 64 GPUs for about 15 days, totaling about 23k GPU-hours per full pass.

Preprocessing is implemented in the data loader in two stages.
First, we normalize each recording to a fixed resolution and frame rate.
This produces tensors for RGB video, per-frame cursor coordinates, and mouse/keyboard event traces (in both raw-action and meta-action views).
Second, we render an SVG cursor at each logged position to produce per-frame masks and cursor-only reference frames.
The first reference frame contains the full desktop with a unit mask.
Later references paste only the cursor over a neutral background, with a mask restricted to arrow pixels.
After VAE encoding, these references become latent slots that pin down the static GUI layout at $t{=}0$.
For $t{>}0$, they supervise only a small patch around the cursor and leave the rest of the frame unconstrained.
We drop clips without valid cursor or action traces to keep~supervision~consistent.

\subsubsection{Evaluation setup}
Our ablations target three capabilities: global fidelity, post-action responsiveness, and cursor-control precision.
We use the \FVD/\LPIPS/\SSIM~suite as the core metrics.
We also add action-driven metrics that focus on post-interaction frames after clicks, scrolls, and key/type events.
For example, we compute \SSIM/\LPIPS~averaged over the $k$ frames after each logged action, and action-driven \FVD~on post-action clips.
Ablations vary conditioning design and action encoding to measure how these choices affect perceptual quality and responsiveness when rolled out against ground-truth interfaces\footnote{Full metric definitions and implementation details are provided in Appendix~\ref{appendix:gui-metrics}.}.

\begin{tcolorbox}[
  colback=apricotglaze!40!white,
  colframe=metablue!40,
  interior style={shade,shading angle=315,left color=white,right color=apricotglaze!55!white},
  boxrule=0pt,
  borderline west={1pt}{0pt}{gray!60},
  left=6pt,right=4pt,top=3pt,bottom=3pt]
\small
\begin{badgeitems}
  \item In GUIWorld, a small amount of goal-directed data outperforms much larger random exploration, showing that alignment quality matters more than nominal scale for action--response learning.
  \item Precise cursor control requires explicit visual supervision: SVG mask/reference conditioning raises cursor accuracy to 98.7\%, indicating that local GUI control primitives are learnable in controlled settings.
  \item Action injection depth matters: relative to shallow \texttt{external} conditioning, \texttt{contextual}, \texttt{residual}, and especially \texttt{internal} fusion improve post-action responsiveness and visual consistency.
  \item Action representation also matters: under the same injection mode, API-like meta-actions consistently outperform raw event-stream encoding.
\end{badgeitems}
\end{tcolorbox}

\exptitle[\guiworldLogo]{7}{Data quality dominates performance}

Interactive GUI modeling shows that data quality matters more than dataset size for action-driven performance.
We compare slow exploration, fast interaction, and supervised trajectories under \texttt{contextual} conditioning.
This isolates which behaviors best support neural computer training.

\begin{wraptable}{r}{0.48\linewidth}
  \vspace{-0.1em}
  \captionsetup{type=table,width=\linewidth}
  \centering
  \small
  \rowcolors{2}{tablewarmrow}{white}
  \caption{Overall performance across data sources.}
  \label{tab:gui-data-quality}
  \setlength{\tabcolsep}{7pt}
  \begin{tabularx}{\linewidth}{l >{\centering\arraybackslash}X >{\centering\arraybackslash}X >{\centering\arraybackslash}X}
    \toprule
    \rowcolor{tablewarmhead}\textbf{Split} & \textbf{FVD$_\text{all}$} & \textbf{SSIM$_\text{all}$} & \textbf{LPIPS$_\text{all}$} \\
    Untrained baseline & 149.61 & 0.496 & 0.605 \\
    Random Fast (train) & 48.17 & 0.695 & 0.483 \\
    Random Slow (train) & 20.37 & 0.830 & 0.237 \\
    Claude CUA (train) & 14.72 & 0.885 & 0.144 \\
    \bottomrule
  \end{tabularx}
  \vspace{-1.8em}
\end{wraptable}

Despite approximately 1{,}400 hours of random exploration across the slow and fast settings, these datasets are noisy.
They are comparatively sample-inefficient for learning stable action--response mappings.
They substantially improve global perceptual metrics over a baseline (Table~\ref{tab:gui-data-quality}).
However, high-frequency cursor jitter and irregular, non-goal-directed action bursts make consistent control difficult under dense, stochastic input streams.

In contrast, the substantially smaller high-quality dataset (110 hours from Claude CUA) yields markedly stronger performance across all metrics.
Goal-directed trajectories provide clearer action semantics and more predictable state transitions.
This enables robust action conditioning even with limited data volume.
These results indicate that neural computer development should prioritize curated, purposeful interactions over large-scale passive data collection.
At the current stage, this result primarily indicates that alignment quality matters more than nominal scale for learning action--response structure in NC prototypes.

\exptitle[\guiworldLogo]{8}{Precise cursor control requires explicit visual supervision}

\begin{figure}[h]
  \centering
  \includegraphics[width=\linewidth]{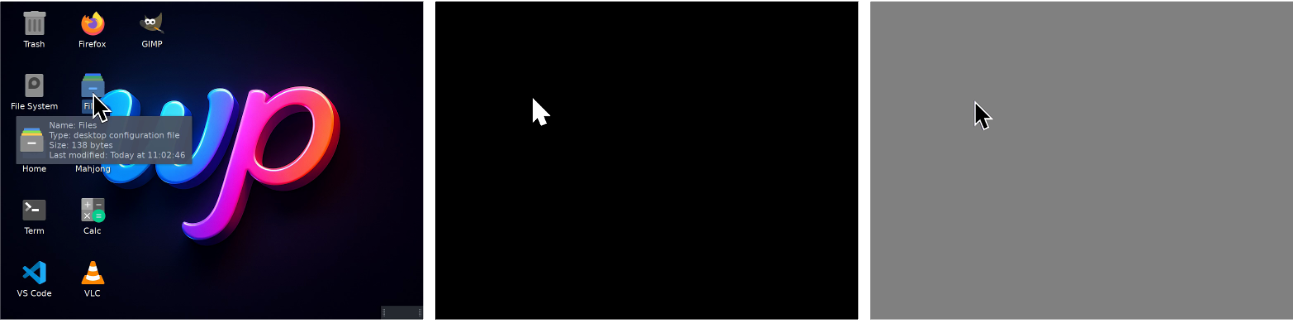}
\caption{\textbf{Cursor references in GUIWorld.}
\textbf{Left:} original desktop frames.
\textbf{Middle:} binary cursor masks.
\textbf{Right:} cursor-only reference frames rendered over a neutral background.}
  \label{fig:cursor-ref}
\end{figure}

\begin{wraptable}{r}{0.48\linewidth}
  \vspace{-1.0em}
  \captionsetup{type=table,width=\linewidth}
  \centering
  \small
  \rowcolors{2}{tablewarmrow}{white}
  \caption{Cursor conditioning losses versus accuracy.}
  \label{tab:cursor-loss}
  \setlength{\tabcolsep}{7pt}
  \begin{tabularx}{\linewidth}{l >{\centering\arraybackslash}X}
    \toprule
    \rowcolor{tablewarmhead}\textbf{Loss variant} & \textbf{Cursor accuracy} \\
    Position $(x,y)$ only & 8.7\% \\
    Position $(x,y)$ + Fourier & 13.5\% \\
    Position $(x,y)$ + SVG mask/ref & \textbf{98.7\%} \\
    \bottomrule
  \end{tabularx}
  \vspace{-0.8em}
\end{wraptable}

We examine whether the NC internalizes cursor dynamics.
A natural baseline is to condition on normalized cursor-coordinate sequences $\texttt{mouse\_trajectories}\subset[0,1]^{T\times 2}$ (details in Appendix~\ref{appendix:mouse-traj-normalization}). 
To strengthen this signal, we further encode the normalized trajectories using a Fourier mouse encoder.
We map coordinates to $[-1,1]^2$ and project them through a fixed Gaussian matrix to obtain random Fourier features.
A small MLP produces per-frame embeddings, which we aggregate into lag-aware windows aligned with the VAE stride.
The resulting latent action sequence conditions the action modules and participates in the temporal contrastive loss.

However, Table~\ref{tab:cursor-loss} shows that coordinate-based supervision remains insufficient for precise interaction.
Position-only supervision achieves 8.7\% accuracy, and even enhanced position features reach only 13.5\%.
This suggests that richer coordinate encodings alone do not resolve cursor drift and jitter.

Motivated by the importance of precise cursor placement, we introduce explicit visual cursor supervision.
We render an SVG cursor at each $(x_t,y_t)$ to produce per-frame cursor masks $m_t$ and cursor-only foregrounds $f_t$ (right panel of Figure~\ref{fig:cursor-ref}).
Following Figure~\ref{fig:cursor-ref}, we construct a reference stream.
The first frame contains the full desktop image, while subsequent frames contain only the cursor foreground over a neutral background, masked to the cursor region.
We encode both the video and reference streams with the shared VAE, yielding video latents $z_{1:T}$, reference latents $z^{\text{ref}}_{1:T}$, and mask tags $\tau_{1:T}$.
The diffusion transformer receives the concatenated tensor
\[
  \mathrm{concat}\bigl(z_{1:T},\,\tau_{1:T},\,z^{\text{ref}}_{1:T}\bigr),
\]
which anchors the static GUI layout at $t{=}0$ and provides localized supervision around the cursor for $t{>}0$.

Under this explicit visual conditioning, cursor accuracy improves to 98.7\%.
This suggests that neural computers benefit from learning the cursor state as a visual object rather than relying solely on abstract coordinates.
Explicit pixel-level supervision helps model cursor acceleration, hover states, and click feedback, which are essential for reliable GUI interaction.
At the same time, this result is best viewed as evidence that local GUI control primitives are learnable under explicit supervision in controlled settings.

\exptitle[\guiworldLogo]{9}{Action injection under different schemes}

\begin{table}[h]
  \centering
  \footnotesize
  \rowcolors{2}{tablewarmrow}{white}
  \caption{Action-driven metrics across injection schemes (15 frames after action).}
  \label{tab:gui-action-ssim}
  \setlength{\tabcolsep}{11pt}
  \begin{tabularx}{1\linewidth}{l >{\centering\arraybackslash}X >{\centering\arraybackslash}X >{\centering\arraybackslash}X}
    \toprule
    \rowcolor{tablewarmhead}
    \textbf{Mode} &
    \textbf{SSIM$_{+15}$} $\uparrow$ &
    \textbf{\LPIPS$_{+15}$} $\downarrow$ &
    \textbf{FVD$_{+15}$} $\downarrow$ \\
    \texttt{baseline$_1$} (untrained)                          & 0.326 & 0.649 & 184.3 \\
    \texttt{baseline$_2$}\textsuperscript{\ensuremath{\dagger}} (\texttt{external}) & 0.746 & 0.251 & 33.4  \\
    \texttt{contextual}                      & 0.813 & 0.190 & 24.8  \\
    \texttt{residual}                        & 0.857 & \textbf{0.138} & 18.8 \\
    \texttt{internal}                        & \textbf{0.863} & 0.141 & \textbf{14.5} \\
    \bottomrule
  \end{tabularx}
  \vspace{-0.35em}
  {\scriptsize\raggedright
  \textsuperscript{\ensuremath{\dagger}}\,\texttt{baseline$_2$} (\texttt{external}) was early-stopped at $\sim$50\% of the planned training budget after preliminary rollouts did not warrant further compute.
  Included only as a rough reference.\par}
\end{table}

Holding data and the action encoder fixed, we compare injection schemes on clean runs (Table~\ref{tab:gui-action-ssim}).
We compute action-driven metrics over the 15 frames following each click, scroll, or key event.
Relative to both baselines (untrained and \texttt{external}), mid- and deep-level fusion yields consistent improvements in post-action quality.
This includes \texttt{contextual}, \texttt{residual}, and \texttt{internal} injection.

Specifically, moving from input-level conditioning (\texttt{external}) to token-level fusion (\texttt{contextual}) improves SSIM from 0.746 to 0.813 and reduces FVD from 33.4 to 24.8.
Deeper injection sharpens these gains.
\texttt{internal} achieves the highest structural consistency (SSIM 0.863) and the lowest temporal distortion (FVD 14.5), while \texttt{residual} attains the lowest perceptual distance (LPIPS 0.138).
Together, these trends associate deeper action injection with improved tracking of fine-grained cursor motion and layout changes.
\footnote{Appendix~\ref{appendix:gui-actions} summarizes the corresponding injection schemes and alignment details.}

\exptitle[\guiworldLogo]{10}{Do action encodings matter?}

\begin{table}[h]
  \centering
  \footnotesize
  \rowcolors{2}{tablewarmrow}{white}
  \caption{Raw-action vs.\ API-like action encoding under the same injection mode (15 frames after action).}
  \label{tab:gui-encoding-ablation}
  \setlength{\tabcolsep}{11pt}
  \begin{tabularx}{1\linewidth}{l l >{\centering\arraybackslash}X >{\centering\arraybackslash}X >{\centering\arraybackslash}X}
    \toprule
    \rowcolor{tablewarmhead}
    \textbf{Mode} & \textbf{Encoding} &
    \textbf{SSIM$_{+15}$} $\uparrow$ &
    \textbf{\LPIPS$_{+15}$} $\downarrow$ &
    \textbf{FVD$_{+15}$} $\downarrow$ \\
    \texttt{internal} & raw-action (event-stream) & 0.847 & 0.144 & 16.6 \\
    \texttt{internal} & meta-action (API-like)   & \textbf{0.863} & \textbf{0.141} & \textbf{14.5} \\
    \bottomrule
  \end{tabularx}
\end{table}

We compare two action encodings under the same injection mode to isolate the effect of representation choice (Table~\ref{tab:gui-encoding-ablation}).
Under \texttt{internal} conditioning, the meta-action (API-like) encoding yields small but consistent improvements over the raw-action representation.
SSIM increases from 0.847 to 0.863, LPIPS drops from 0.144 to 0.141, and FVD drops from 16.6 to 14.5.
However, these gains are modest compared to the substantially larger improvements observed when varying the action injection scheme itself (Table~\ref{tab:gui-action-ssim}).
This suggests that encoding granularity is not the dominant factor governing GUI interaction fidelity.

\begin{table}[t]
  \centering
  \caption{Encoding examples for raw-action and meta-action encoders.}
  \label{tab:keyboard-encoding}
  \setlength{\tabcolsep}{6pt}
  \renewcommand{\arraystretch}{1.08}
  \rowcolors{2}{tablewarmrow}{white}

  \newcommand{\tb}{\raisebox{0.2ex}{\scriptsize$\bullet$}\hspace{0.35em}}

  \begin{minipage}[c]{0.26\linewidth}
    \centering
    \includegraphics[width=0.85\linewidth]{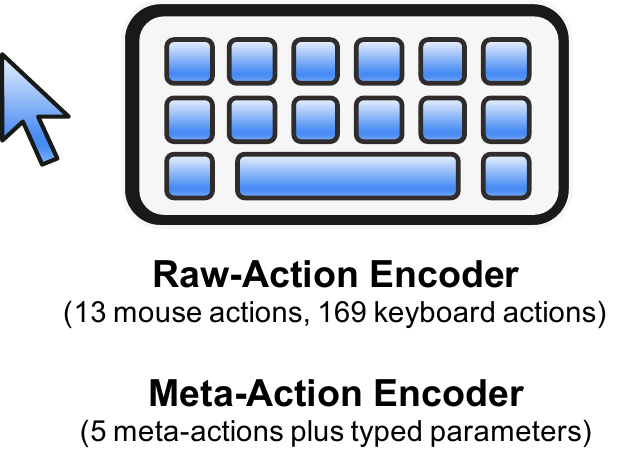}
  \end{minipage}
  \hfill
  \begin{minipage}[t]{0.71\linewidth}
    \small
    \begin{tabularx}{\linewidth}{@{}>{\raggedright\arraybackslash}p{2.2cm}
                                >{\raggedright\arraybackslash}X
                                >{\raggedright\arraybackslash}X@{}}
      \toprule
      \rowcolor{tablewarmhead}
      \textbf{User intent} &
      \textbf{Raw-action encoder (event stream)} &
      \textbf{Meta-action encoder (API-like slot)} \\
      \midrule

      \texttt{ls -l} &
      \parbox[t]{\linewidth}{%
        \tb Key events per character (e.g., \texttt{l}, \texttt{s}, \texttt{<space>}, \texttt{-}, \texttt{l})\\
        \tb Activates entries in a 169-d keyboard multi-hot\\
        \tb No explicit command semantics; inferred from sequence%
      } &
      \parbox[t]{\linewidth}{%
        \tb \texttt{type: KeyboardType}\\
        \tb \texttt{text: "ls -l"}\\
        \tb Encoded by shared text encoder%
      } \\
      \midrule[0.8pt]

      \texttt{ctrl+v} &
      \parbox[t]{\linewidth}{%
        \tb Separate keydown/keyup events for \texttt{ctrl} and \texttt{v}\\
        \tb Activated in multi-hot (or shortcut entry)%
      } &
      \parbox[t]{\linewidth}{%
        \tb \texttt{type: Shortcut}\\
        \tb \texttt{id: ctrl+v}\\
        \tb Embedded via shortcut table%
      } \\
      \bottomrule
    \end{tabularx}
  \end{minipage}
\end{table}

Table~\ref{tab:keyboard-encoding} contrasts how short commands and shortcuts (e.g., \texttt{ls -l}, \texttt{ctrl+v}) are represented under the two encodings.
The raw-action encoder treats typing as a stream of individual key events, leaving command or shortcut semantics to be inferred from the sequence.
In contrast, the meta-action encoder collapses each interaction into a single typed action with associated text or a shortcut identifier.
This design aims to model user actions as structured, tool-like operations rather than fragmented event streams.

In practice, this more structured abstraction does not translate into clear qualitative gains.
Rendered text remains similarly smeared under both encodings, and robustness under theme changes and timing noise is largely unchanged.
Task-level failures such as re-centering, re-acquisition, and multi-step interactions persist across both representations.
We adopt the meta-action encoder as the default for its simplicity and semantic alignment with system-level conditioning.
These results suggest that encoding granularity is secondary to alignment quality and injection strategy.

\subsubsection{Visualizations}

Across GUIWorld interactive rollouts, failure modes are dominated by data quality and by \emph{where} action information enters the backbone.
Goal-directed supervision produces smooth, target-aligned cursor paths and consistent post-click UI transitions, whereas random exploration yields bursty jitter and spurious actions that degrade visual coherence (Table~\ref{tab:gui-data-quality}; Figures~\ref{fig:guiworld-sample-1}--\ref{fig:guiworld-sample-5}).
Consistent with the action-driven metrics in Table~\ref{tab:gui-action-ssim}, deeper token-level injection (\texttt{contextual}/\texttt{internal}) yields more reliable post-action updates in interactive elements (hover states, dropdowns, modals) and maintains cursor alignment under rapid motion.

Figures~\ref{fig:guiworld-sample-6}--\ref{fig:guiworld-sample-8} emphasize how small low-level deviations compound.
Figures~\ref{fig:guiworld-sample-9}--\ref{fig:guiworld-sample-11} focus on numeric/UI fidelity and interaction semantics.
Figures~\ref{fig:guiworld-sample-12}--\ref{fig:guiworld-sample-14} add stress cases where correctness hinges on precise field edits and page state.
All full-resolution pages are in Appendix~\ref{appendix:vis}; below we keep clickable thumbnails at the original location for quick navigation.

\newcommand{\GuiThumbPageHeader}[1]{%
  {\par\noindent\centering\Large\bfseries #1\par}
  {\par\noindent\centering%
    \begin{tcolorbox}[
      enhanced,
      boxrule=0.85pt,
      colframe=orange!80!black,
      colback=yellow!14,
      arc=5.5pt,
      left=6pt,right=6pt,top=2.5pt,bottom=2.5pt,
      width=0.78\linewidth
    ]
      \centering\small\bfseries\sffamily\color{orange!85!black}
      Click any thumbnail to jump to its full-resolution page in Appendix
    \end{tcolorbox}\par
  }
  \vspace{0.26em}%
}
\newcommand{\GuiThumbPageFrame}{%
  \begin{tikzpicture}[remember picture,overlay]
    \draw[
      draw=black,
      line width=1.9pt,
      rounded corners=14pt
    ]
      ([xshift=0.48cm,yshift=-0.48cm]current page.north west)
      rectangle
      ([xshift=-0.48cm,yshift=0.48cm]current page.south east);
  \end{tikzpicture}%
}
\newcommand{\GuiThumbCard}[4]{%
  \par\noindent\makebox[\linewidth][c]{%
    \hyperref[#1]{%
      \begin{tikzpicture}
        \node[inner sep=0pt] (img) {\includegraphics[width=#4\linewidth,keepaspectratio]{#2}};
        \node[
          anchor=south east,
          xshift=-0.38em,
          yshift=0.26em,
          fill=white,
          fill opacity=0.93,
          text opacity=1,
          draw=metablue!50,
          rounded corners=1.5pt,
          inner xsep=0.45em,
          inner ysep=0.2em,
          minimum width=8.4em,
          align=center,
          font=\scriptsize\bfseries,
          text=metablue!90!black
        ] at (img.south east) {#3};
      \end{tikzpicture}%
    }%
  }\par\vspace{0.28em}%
}
\newcommand{\GuiThumb}[4]{\GuiThumbCard{#1}{#2}{#3}{#4}}

\clearpage
\newgeometry{top=0.9cm,bottom=1.35cm,left=0.9cm,right=0.9cm,includefoot,footskip=18pt}
\thispagestyle{plain}
\hypertarget{main-gui-vis-thumbs}{}
\hypertarget{main-gui-vis-thumbs-p1}{}
\hypertarget{main-gui-vis-thumbs-p2}{}
\GuiThumbPageFrame
\GuiThumbPageHeader{GUIWorld Visualization Thumbnails}
\noindent\begin{tabular}{@{}p{0.495\linewidth}!{\color{metablue!55}\vrule width 0.8pt}p{0.495\linewidth}@{}}
\begin{minipage}[t]{\linewidth}
  \vspace*{0pt}
  \centering
  {\small\bfseries \guiworldLogo{}\ GUIWorld Samples 1--5}\par\vspace{0.14em}
  \GuiThumb{fig:guiworld-sample-1}{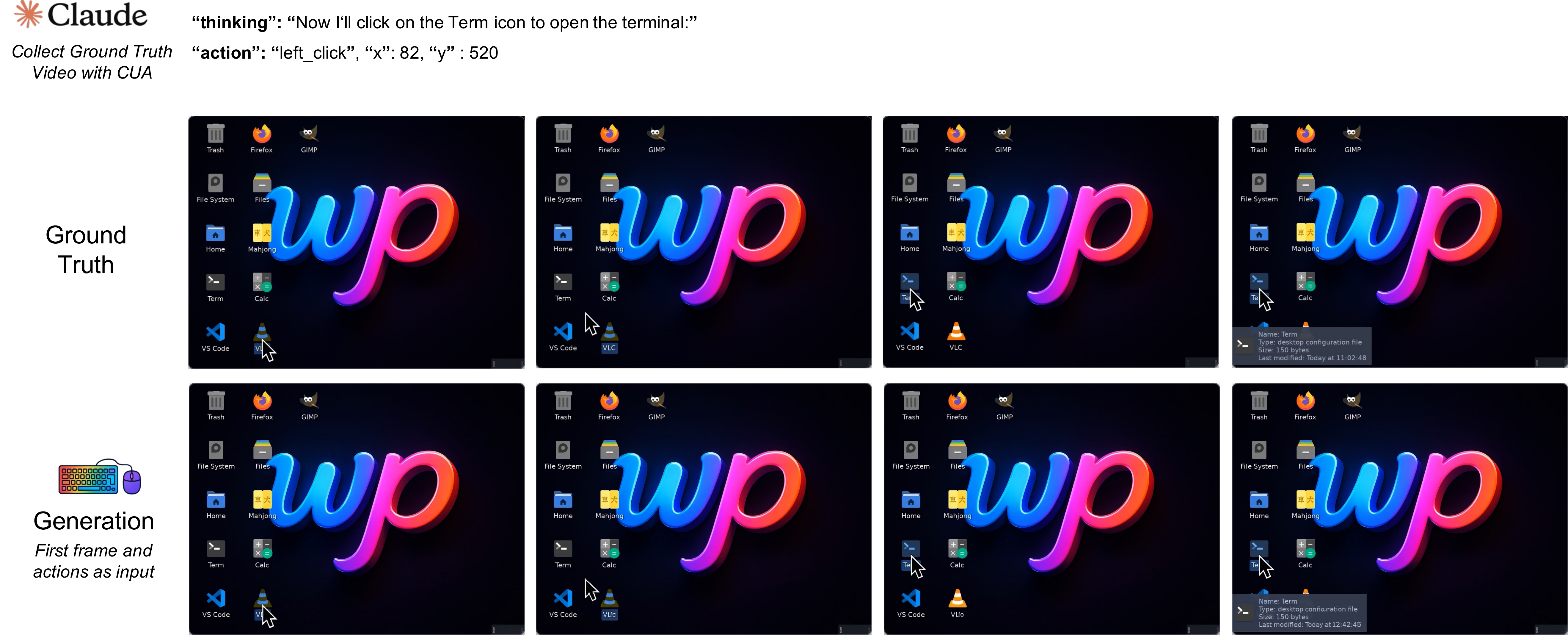}{\guiworldLogo{}\ Samples 1}{0.94}
  \GuiThumb{fig:guiworld-sample-2}{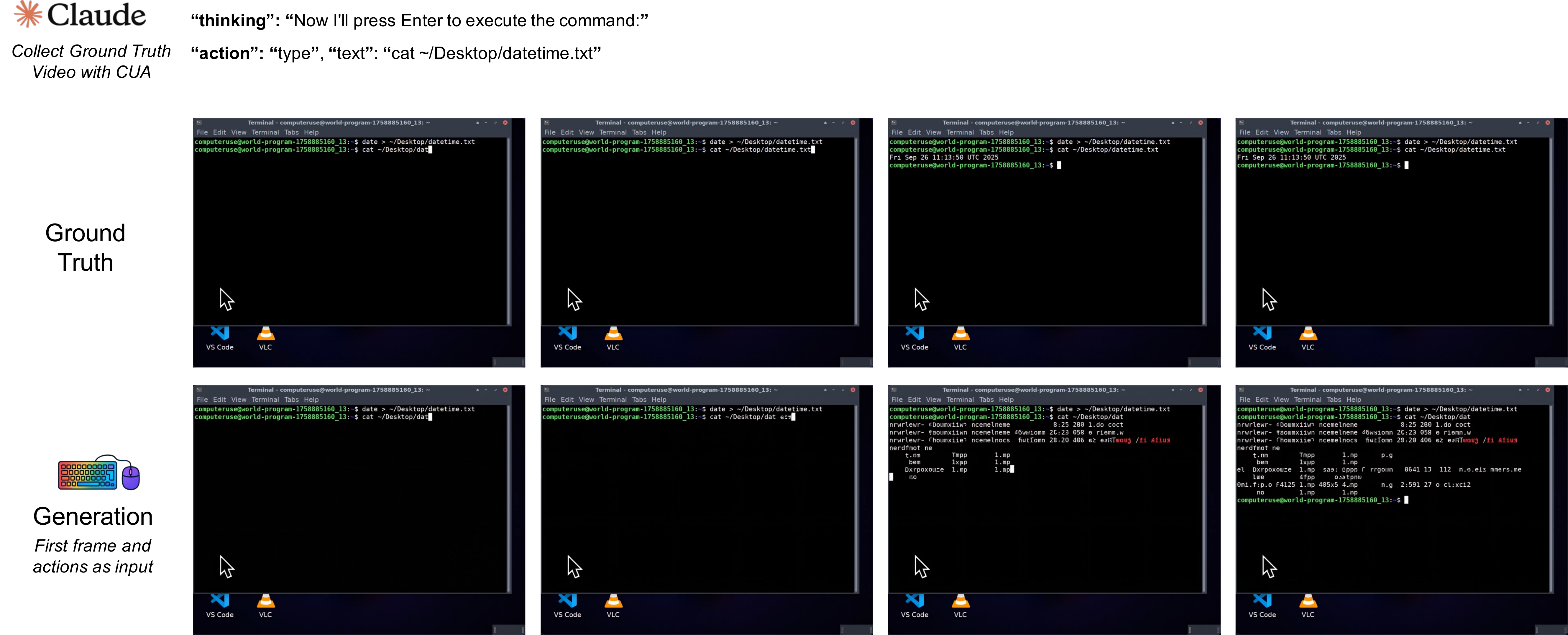}{\guiworldLogo{}\ Samples 2}{0.94}
  \GuiThumb{fig:guiworld-sample-3}{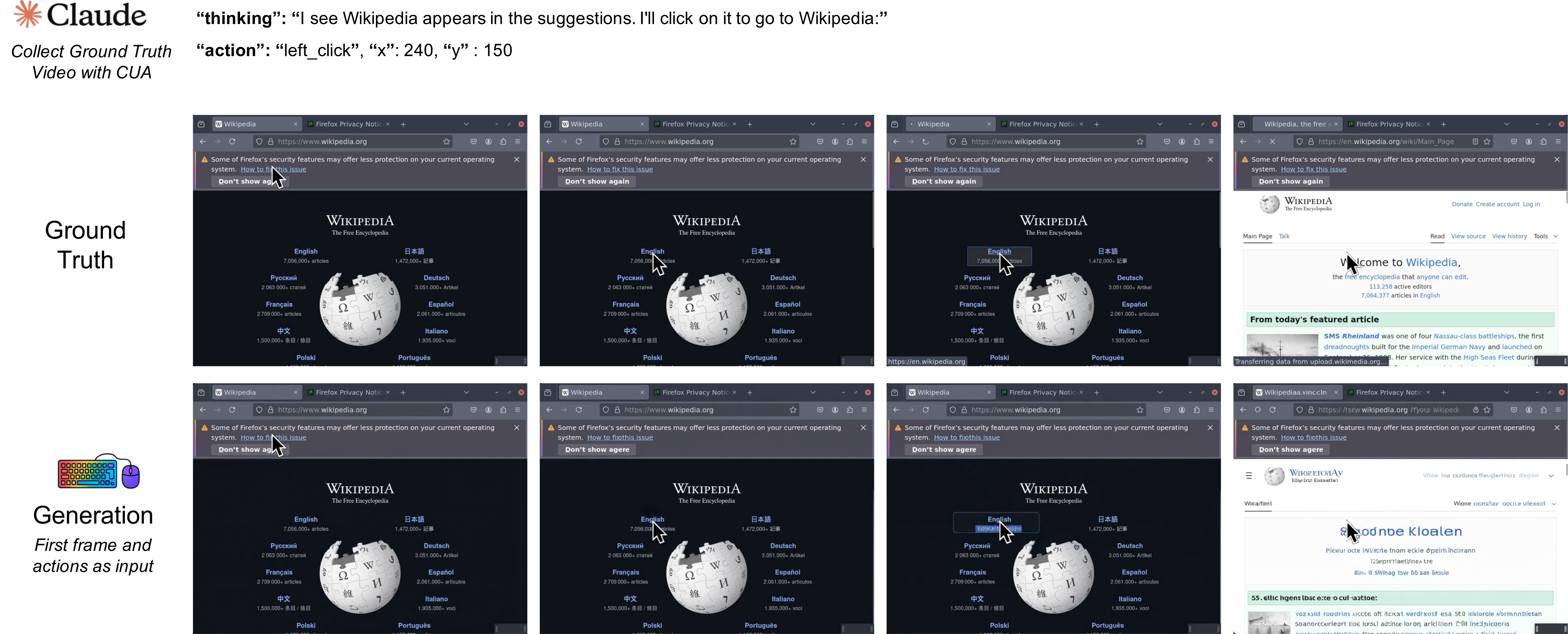}{\guiworldLogo{}\ Samples 3}{0.94}
  \GuiThumb{fig:guiworld-sample-4}{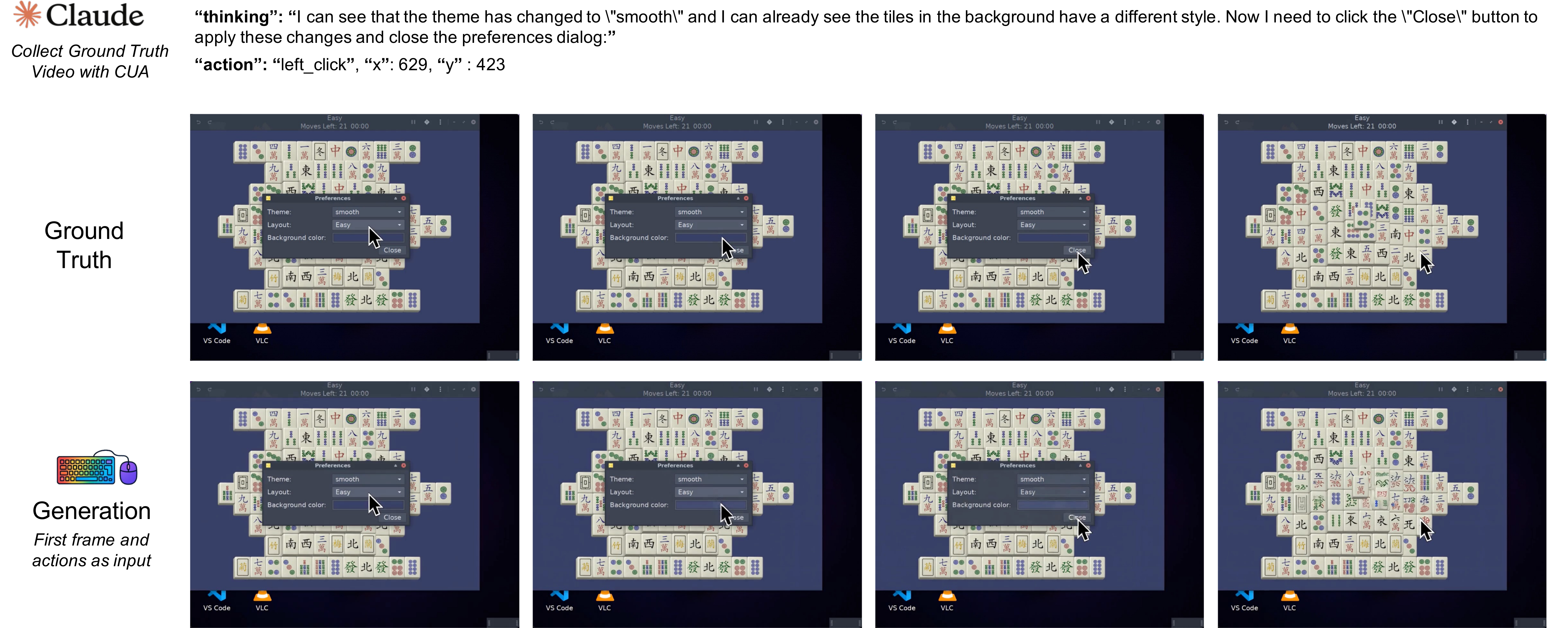}{\guiworldLogo{}\ Samples 4}{0.94}
  \GuiThumb{fig:guiworld-sample-5}{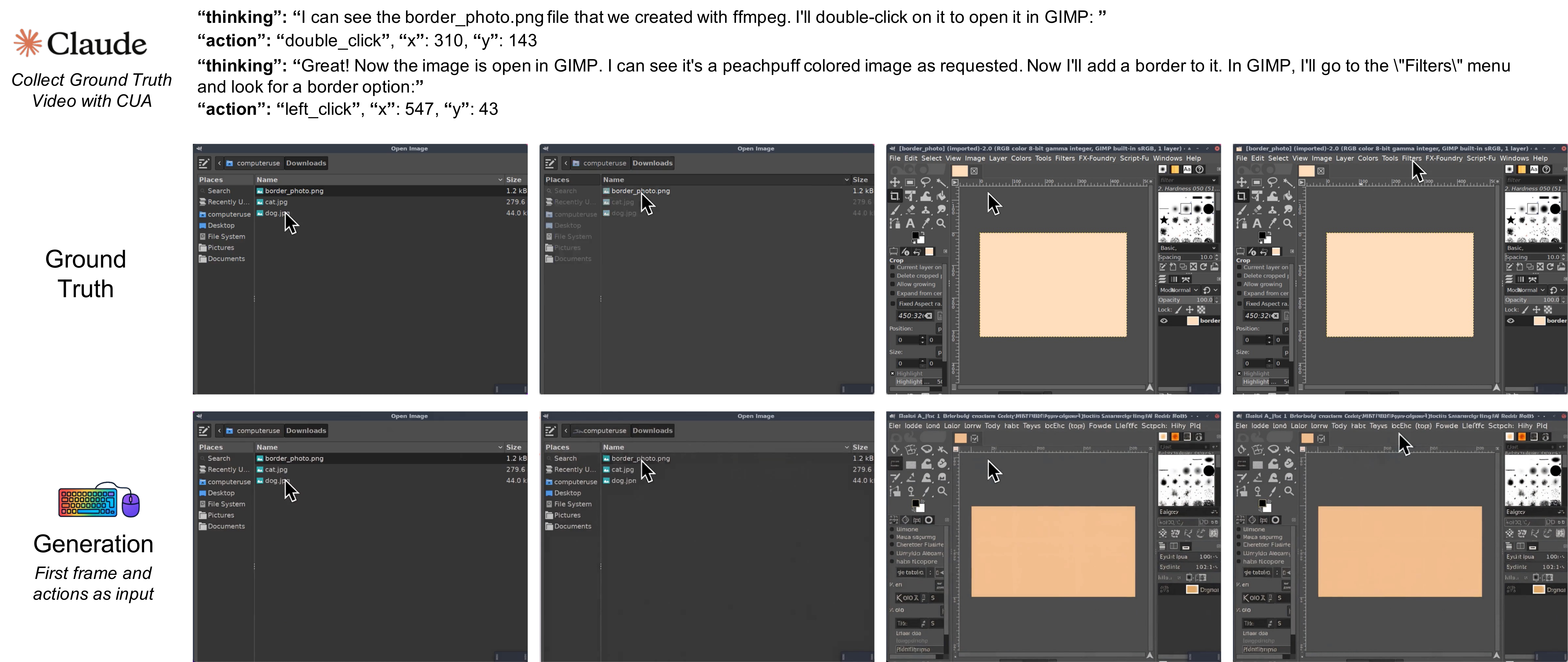}{\guiworldLogo{}\ Samples 5}{0.94}
\end{minipage}
&
\begin{minipage}[t]{\linewidth}
  \vspace*{0pt}
  \centering
  {\small\bfseries \guiworldLogo{}\ GUIWorld Samples 6--10}\par\vspace{0.14em}
  \GuiThumb{fig:guiworld-sample-6}{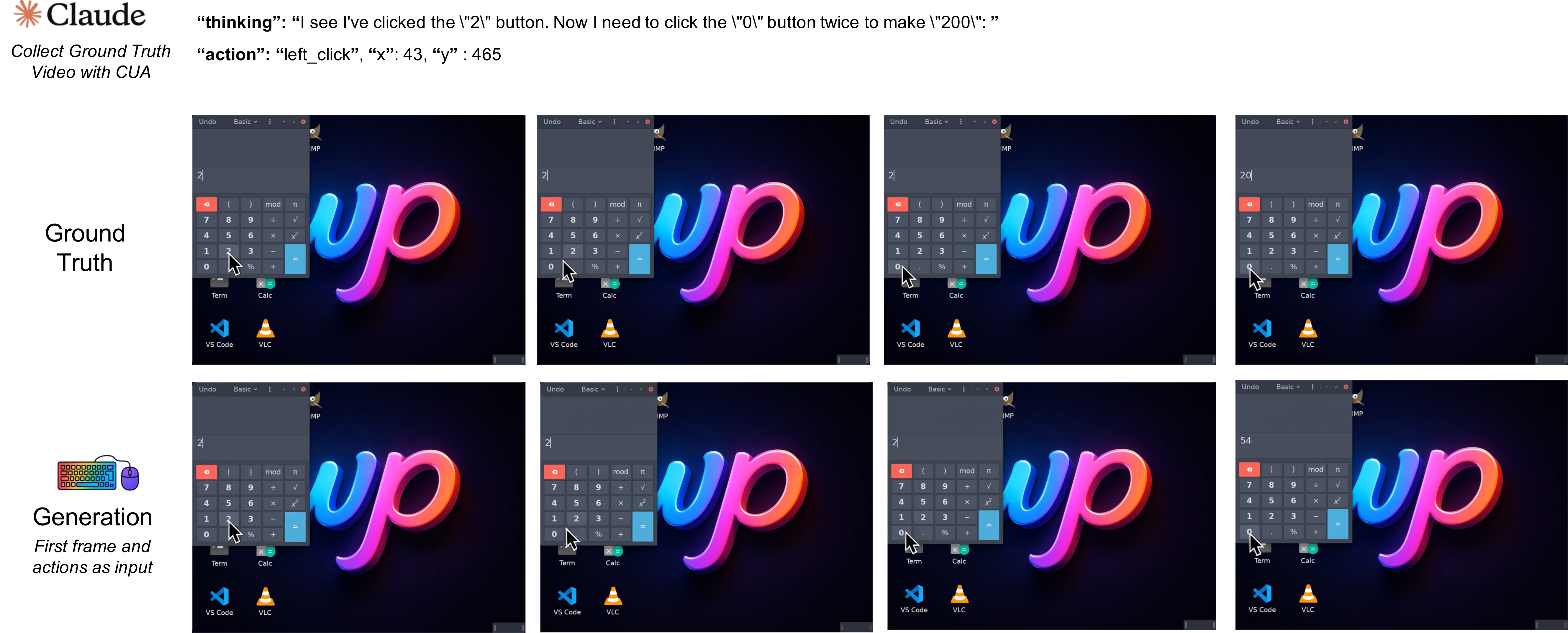}{\guiworldLogo{}\ Samples 6}{0.94}
  \GuiThumb{fig:guiworld-sample-7}{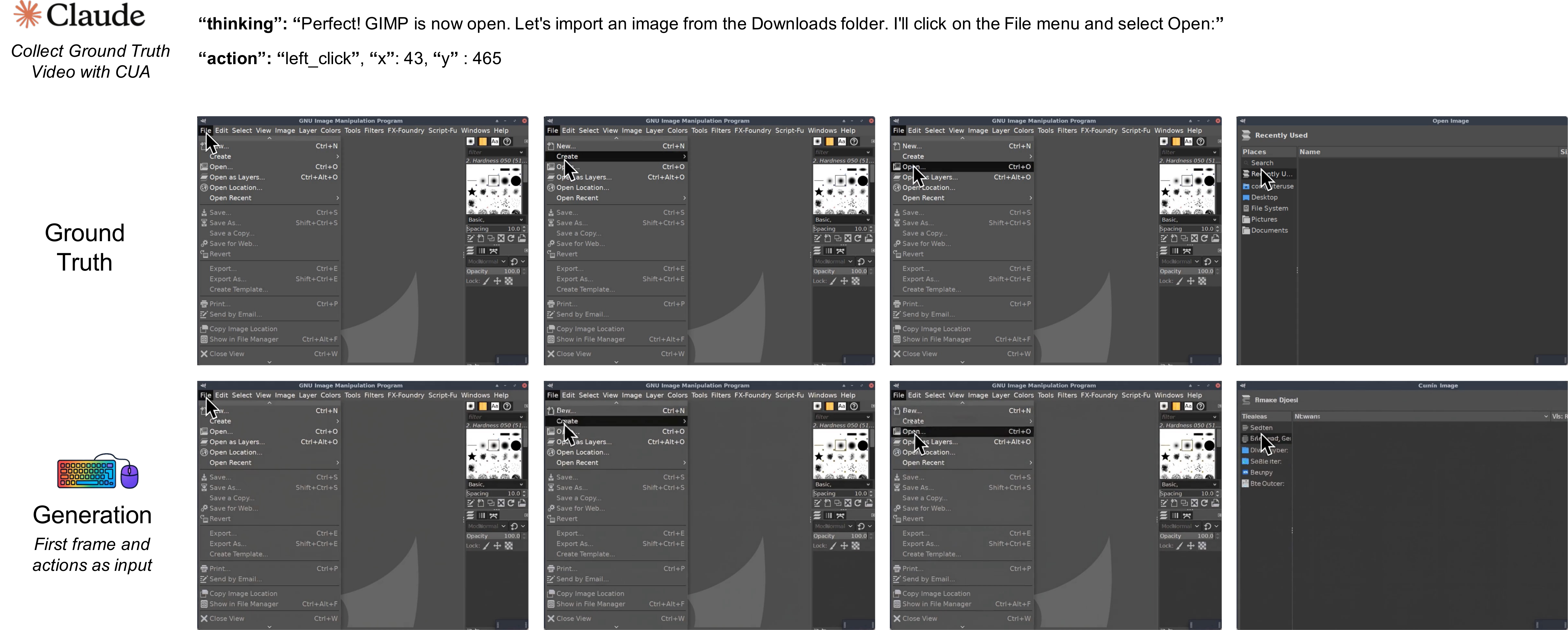}{\guiworldLogo{}\ Samples 7}{0.94}
  \GuiThumb{fig:guiworld-sample-8}{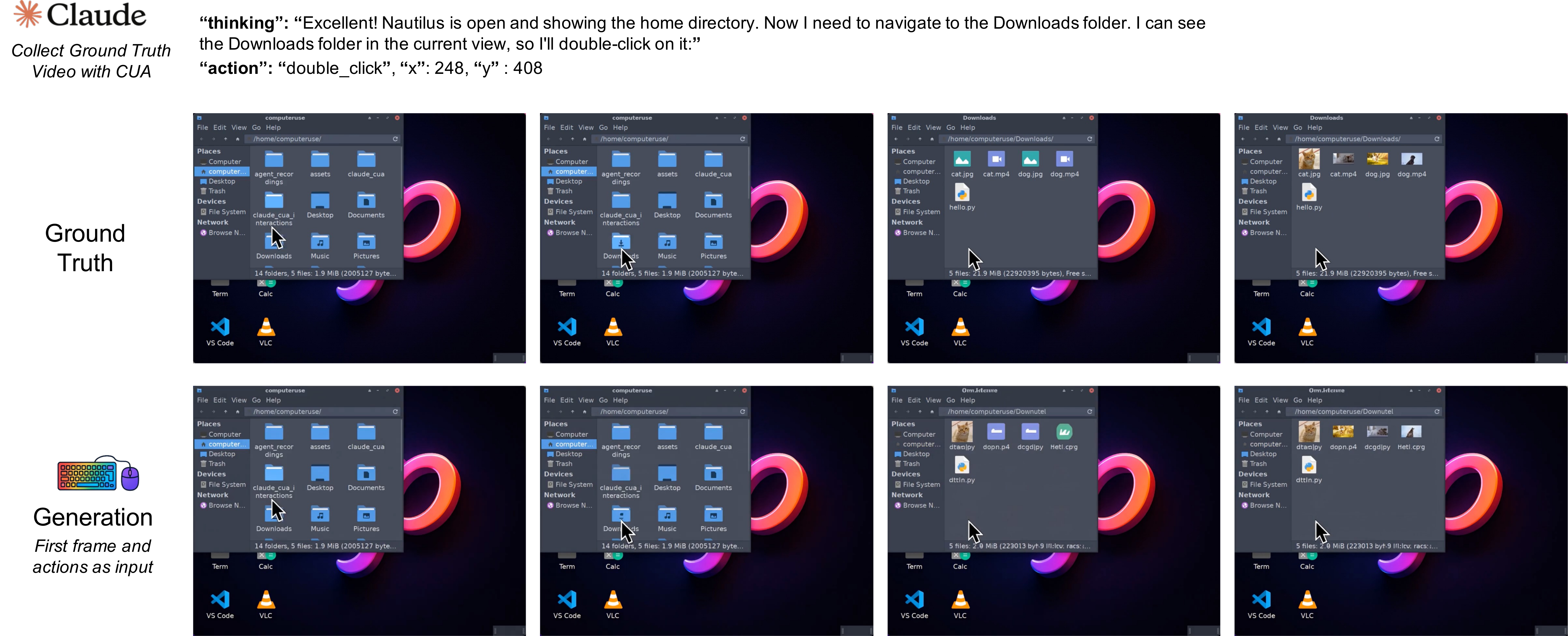}{\guiworldLogo{}\ Samples 8}{0.94}
  \GuiThumb{fig:guiworld-sample-9}{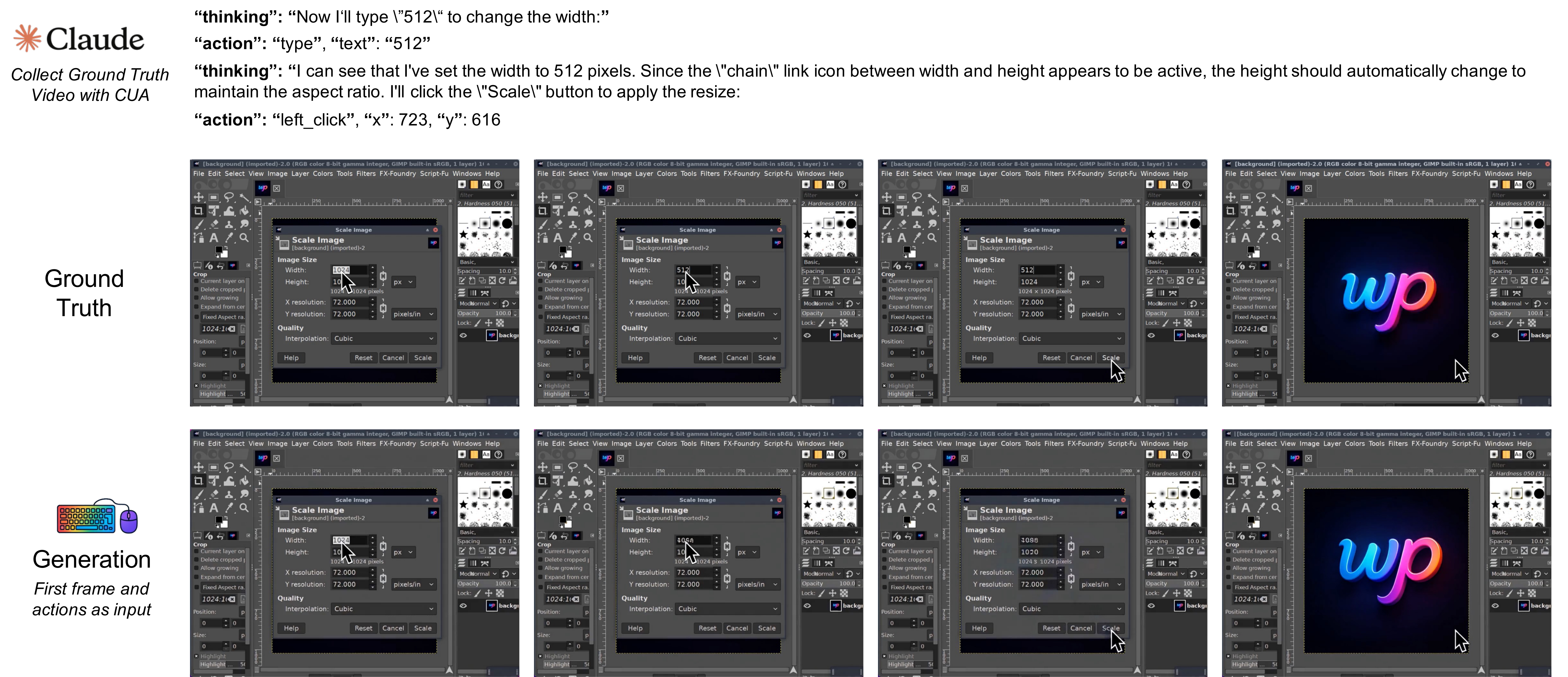}{\guiworldLogo{}\ Samples 9}{0.94}
  \GuiThumb{fig:guiworld-sample-10}{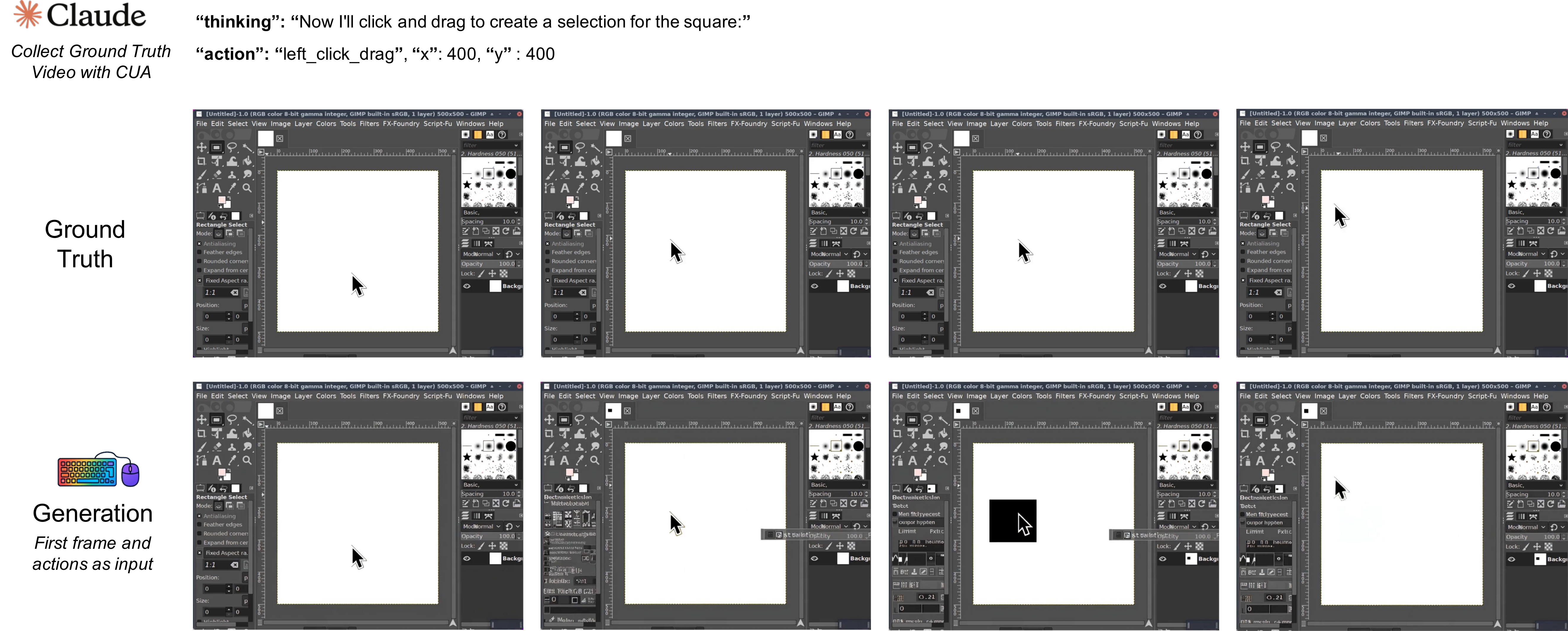}{\guiworldLogo{}\ Samples 10}{0.94}
\end{minipage}
\end{tabular}
\restoregeometry

\clearpage
\newgeometry{top=0.9cm,bottom=1.35cm,left=0.9cm,right=0.9cm,includefoot,footskip=18pt}
\thispagestyle{plain}
\hypertarget{main-gui-vis-thumbs-p3}{}
\hypertarget{main-gui-vis-thumbs-p4}{}
\hypertarget{main-gui-vis-thumbs-p5}{}
\GuiThumbPageFrame
\GuiThumbPageHeader{GUIWorld Visualization Thumbnails}
\noindent\begin{tabular}{@{}p{0.495\linewidth}!{\color{metablue!55}\vrule width 0.8pt}p{0.495\linewidth}@{}}
\begin{minipage}[t]{\linewidth}
  \vspace*{0pt}
  \centering
  {\small\bfseries \guiworldLogo{}\ GUIWorld Samples 11--12}\par\vspace{0.14em}
  \GuiThumb{fig:guiworld-sample-11}{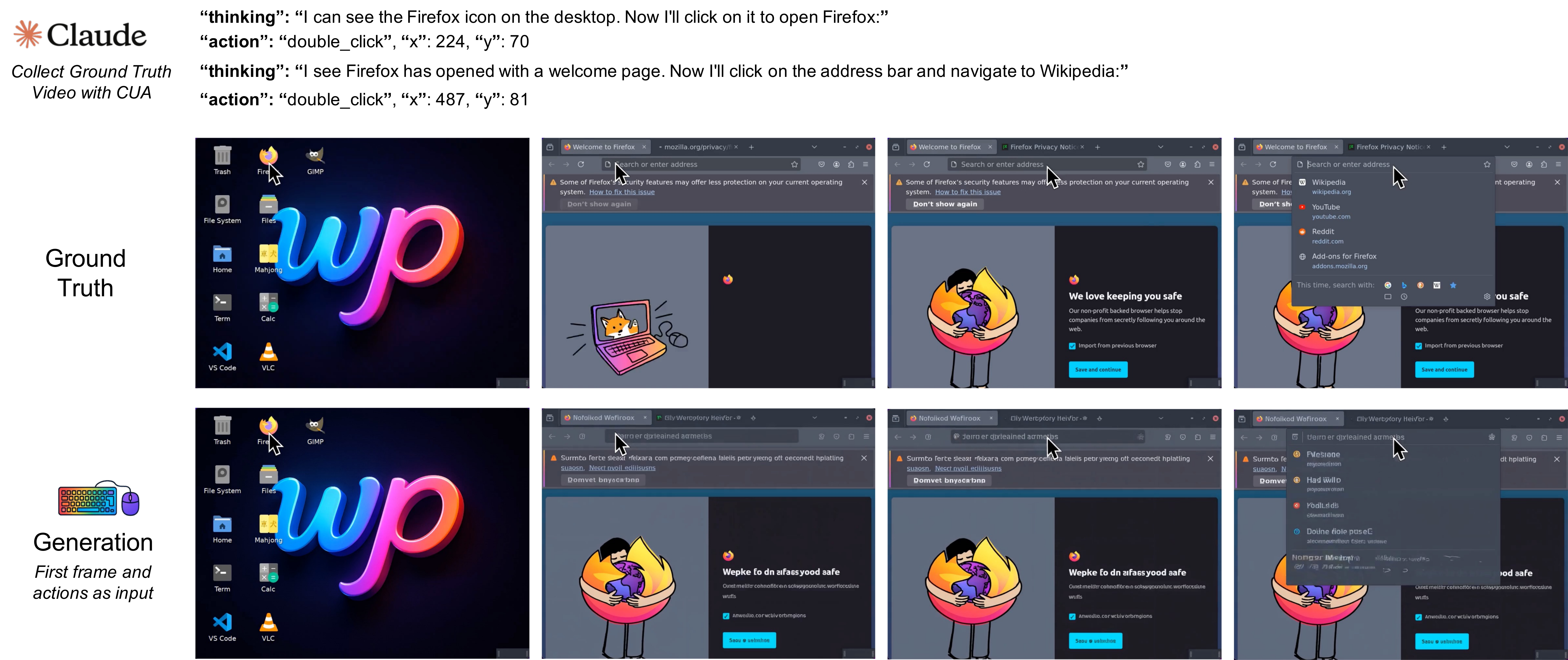}{\guiworldLogo{}\ Samples 11}{0.96}
  \GuiThumb{fig:guiworld-sample-12}{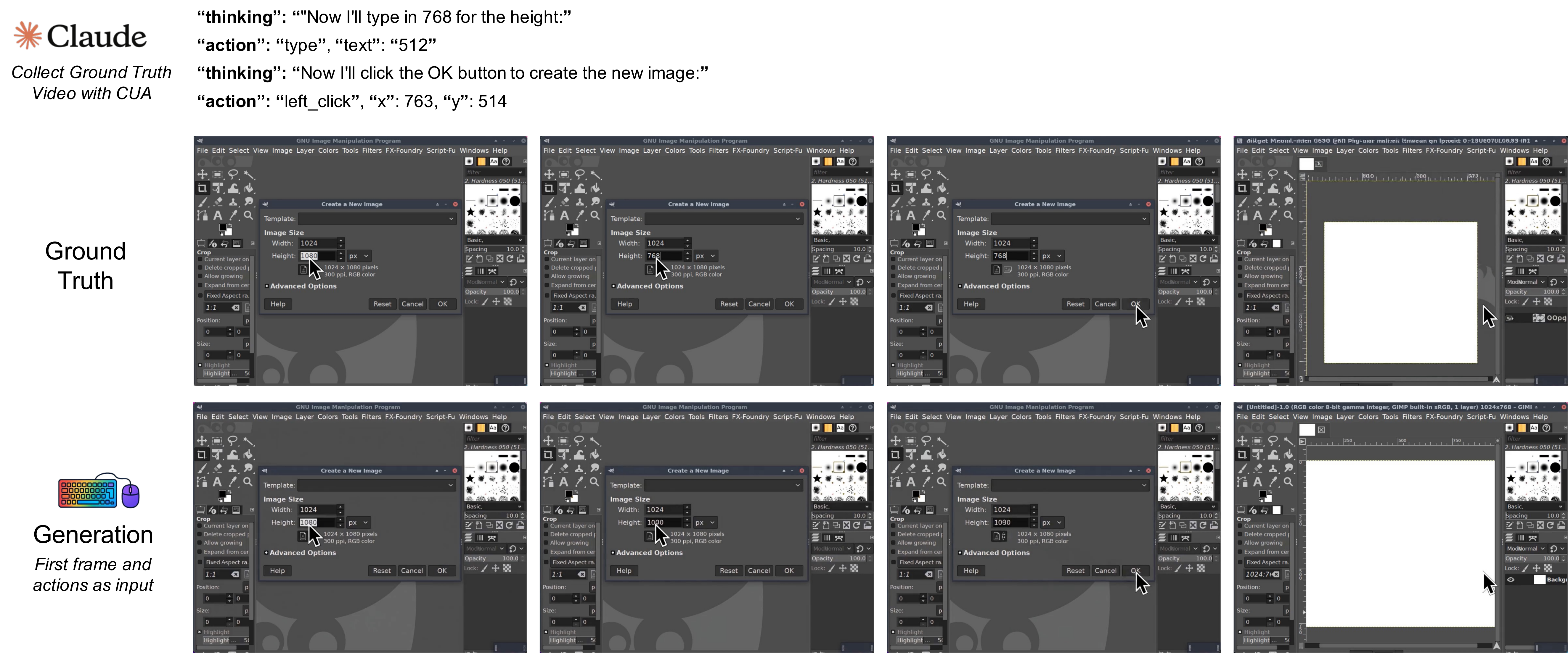}{\guiworldLogo{}\ Samples 12}{0.96}
\end{minipage}
&
\begin{minipage}[t]{\linewidth}
  \vspace*{0pt}
  \centering
  {\small\bfseries \guiworldLogo{}\ GUIWorld Samples 13--14}\par\vspace{0.14em}
  \GuiThumb{fig:guiworld-sample-13}{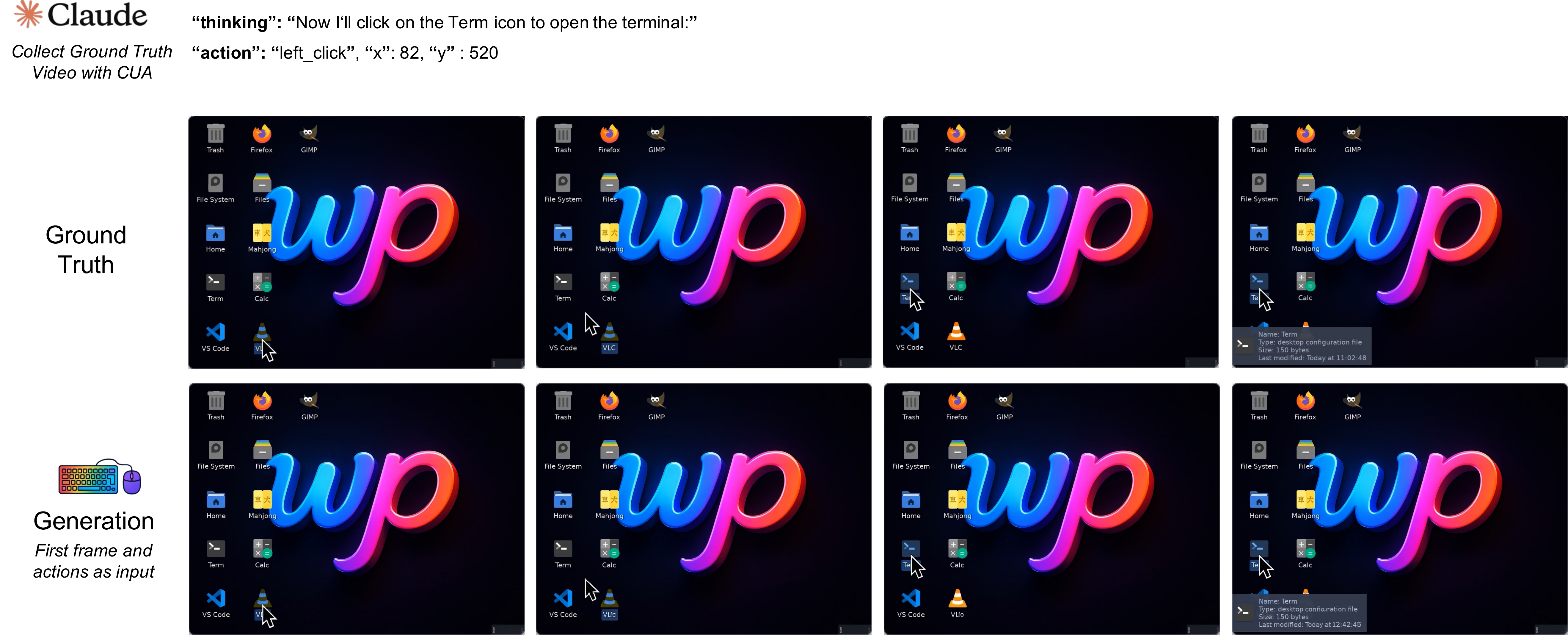}{\guiworldLogo{}\ Samples 13}{0.96}
  \GuiThumb{fig:guiworld-sample-14}{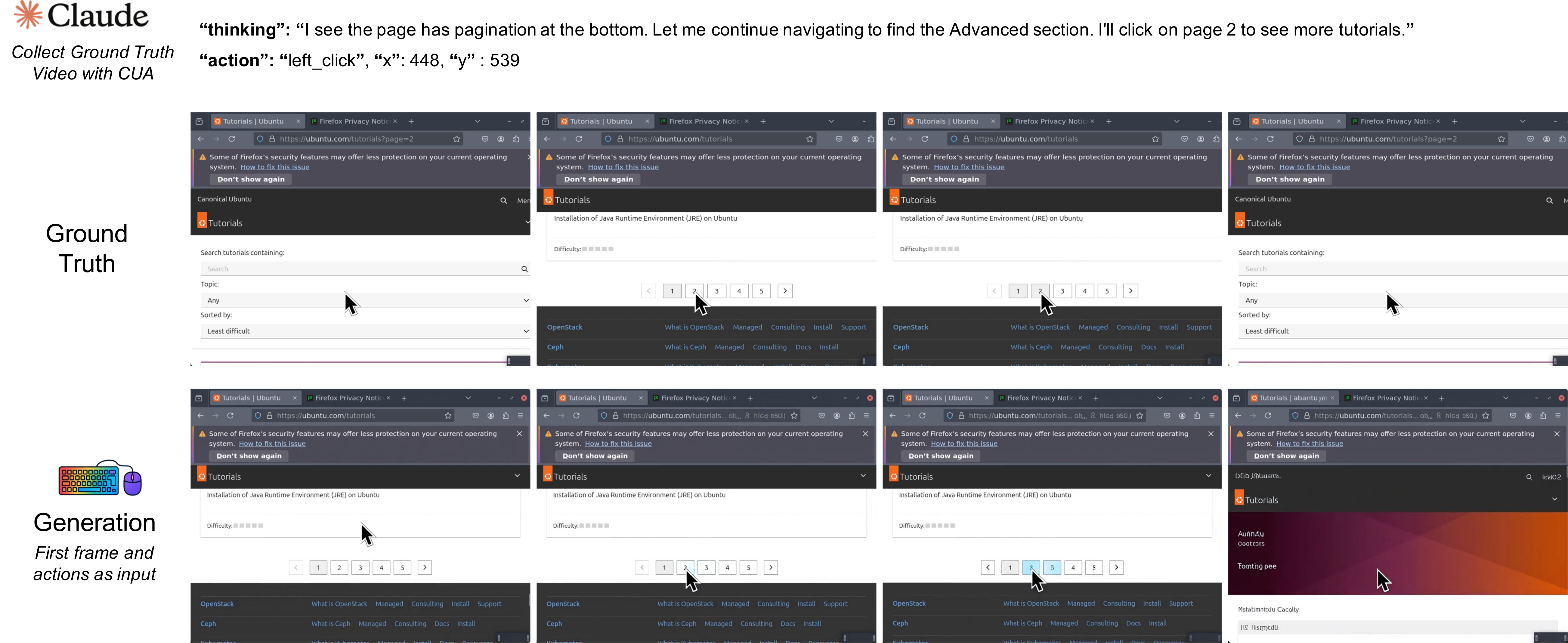}{\guiworldLogo{}\ Samples 14}{0.96}
\end{minipage}
\end{tabular}
\restoregeometry

%% file: section/toward_cnc.tex
\paragraph{Section Overview}
In this section, we ask what current Neural Computer (NC) prototypes have already shown, what still prevents them from becoming usable or general-purpose runtimes, and why neither current world models~\citep{ha2018world,openai2024sora,polyak2024movie,deepmind_veo_2025} nor AI agents~\citep{hong2023metagpt,claudeComputerTool,zhuge2026ai} yet amount to this emerging machine form.
We then contrast NCs with conventional computers, clarifying that they are not a smarter layer on top of the existing stack, and define their mature general-purpose form, namely Completely Neural Computers (CNCs).
Finally, we outline a roadmap toward CNCs, relate NCs to other system objects, and close with several remarks on NCs.

\subsection{From Neural Computers to Completely Neural Computers}

\paragraph{Current Status of NCs}
Our CLI and GUI-based neural computers already show that early runtime primitives can be learned with measurable interface fidelity. In terminal environments, OCR-based text fidelity is already measurable (\Cref{tab:cligen-ocr}); in GUI settings, explicit visual supervision yields strong local cursor control (\Cref{tab:cursor-loss}); and in GUIWorld, aligned goal-directed data clearly outperforms much larger random exploration (\Cref{tab:gui-data-quality}). Taken together, these results suggest that current NCs already support early runtime primitives, especially I/O alignment and short-horizon control, while stable reuse and general-purpose execution remain out of reach.
This does not mean that current prototypes are already close to CNCs; it means that the outline of a distinct machine form has begun to emerge at prototype scale.

However, the current video-based prototypes are only early NC instantiations: if NCs are to mature into general-purpose runtimes, they must go well beyond basic I/O and short-term execution.
At the formal level, this ultimately requires Gödel universality or Turing completeness~\citep{Goedel:31,Church:35,turing1936computable,siegelmann1992computational}, universal programmability~\citep{von1993first,wilkes1981best}, and behavior consistency unless explicitly reprogrammed~\citep{queloz2025explainability}. 

Before those conditions are met in full, progress is better read through practical acceptance lenses: routine reuse, execution consistency, and explicit update governance.
These lenses matter because the immediate question is not whether CNCs have already been achieved, but whether NCs are beginning to behave more like usable runtimes than isolated demonstrations.
For example, once an incident-response routine has been installed, the system should reuse it on later alerts rather than rediscovering the procedure from scratch each time; and if its behavior changes, that change should be attributable to an explicit update rather than ordinary execution.
In practice, this reduces to three acceptance lenses: install--reuse, execution consistency, and update governance, which together offer a more useful view of current NC progress than the full CNC definition alone.

While certain sequential neural architectures are Turing complete~\citep{siegelmann1992computational,perez2021attention} in principle, turning a trained instance into a reliably programmable runtime behavior remains challenging.
Preliminary attempts, including Neural Virtual Machine~\citep{katz2019programmable} and NeuroLISP~\citep{davis2022neurolisp}, have been explored.
Furthermore, ensuring stable behavior over long temporal horizons remains an open problem in neural systems~\citep{kirkpatrick2017overcoming,calanzonelogically}. 
Section~\ref{sec:roadmap} provides a more detailed discussion of these requirements. 
See Section~\ref{sec:relations} for more discussion between NCs and other system objects, including world models and AI agents.

\begin{table}[h]
  \centering
  \small
  \rowcolors{2}{tablewarmrow}{white}
  \caption{Four system objects compared at a common systems level.}
  \label{tab:system-objects}
  \setlength{\tabcolsep}{5pt}
  \renewcommand{\arraystretch}{1.08}
  \begin{tabularx}{\linewidth}{l X X X}
    \toprule
    \rowcolor{tablewarmhead}
    \textbf{System object} & \textbf{Organized around} & \textbf{Source of truth} & \textbf{Primary role} \\
    \textbf{Conventional computer} & Explicit programs & Explicit programs and explicit machine state & Reliably execute explicit programs \\
    \textbf{AI agent} & Tasks & External environments, tools, and workflow state & Accomplish tasks through an existing software stack \\
    \textbf{World model} & Environment dynamics & A learned model of state evolution & Predict and roll out how an environment may evolve \\
    \textbf{Neural computer} & Runtime & Installed capabilities and runtime state inside the learned system & Sustain execution, accumulate capability, and govern updates within one learned machine \\
    \bottomrule
  \end{tabularx}
\end{table}

\paragraph{Fundamental differences between NCs and conventional computers}
We compare NCs and conventional computers.
Here, \emph{conventional computers} denote random-access machines with instruction set architecture~\citep{hartmanis1974power,anagnostopoulos1973computer} and layered OS/application stacks programmed via human-designed high-level languages~\citep{backus1957fortran,backus1963revised}.
NCs differ fundamentally from conventional computers in their architectural and programming-language semantics.

At the architectural level, random-access machines instantiate local, compositional symbolic semantics~\citep{Newell1976}, yielding exactness and interpretability, but brittleness under noise and model mismatch~\citep{BROOKS1991139}. 
Neural computers, by contrast, realize holistic, distributed numerical semantics, trading precise local semantics for robustness and generalization~\citep{ivakhnenko1965,ivakhnenko1967,ivakhnenko1968,ivakhnenko1971,Hinton1986,Bishop2006}. 
Empirical evidence indicates that such holistic numerical representations are particularly well suited to domains characterized by high-dimensional representations~\citep{bengio2013representation}, soft or statistical constraints~\citep{Smolensky1988-SMOOTP-2}, and globally coupled structures~\citep{silver2017mastering,vaswani2017attention}, including perception, natural language, planning under uncertainty, and approximate reasoning.
Although conventional computers can, in principle, emulate NCs, doing so often introduces unnecessary conceptual and engineering complexity when the target tasks are already well matched to neural architectures.

At the programming-language level, NCs differ from conventional computers because their ``language semantics'' are the meanings of user input sequences learned from data rather than explicitly designed by humans. For example, LLMs can be viewed as programmable computers in which prompts act as programs~\citep{Reynolds2021}. In this case, the programming language is a natural language, which no non-neural system has historically been able to interpret robustly at scale~\citep{jm3}. 
More broadly, learned programming-language semantics are not constrained by a human-specified syntax/semantics boundary and can, therefore, encode task-relevant conventions implicitly~\citep{wei2023chainofthoughtpromptingelicitsreasoning}.

\paragraph{Definition of Completely Neural Computers} We use CNC to denote the mature form of an NC. Formally, a Neural Computer instance is complete if it is (1) Turing complete, (2) universally programmable, (3) behavior-consistent unless explicitly reprogrammed, and (4) realizes the architectural and programming-language advantages of NCs relative to conventional computers. The following section unpacks these conditions in operational terms.

\begin{table}[h]
  \centering
  \small
  \rowcolors{2}{tablewarmrow}{white}
  \caption{Operational reading of the four CNC requirements.}
  \label{tab:cnc-operational}
  \setlength{\tabcolsep}{5pt}
  \renewcommand{\arraystretch}{1.08}
  \begin{tabularx}{\linewidth}{l X X}
    \toprule
    \rowcolor{tablewarmhead}
    \textbf{CNC requirement} & \textbf{Plain reading} & \textbf{What engineering evidence should look like} \\
    \textbf{Turing complete} & The system is not restricted to a narrow family of fixed tasks, but can in principle express general computation. & As effective memory and context grow, the same NC should remain able to carry longer and more structured procedures rather than failing by a different shortcut each time. \\
    \textbf{Universally programmable} & Inputs should not only trigger one-off behavior, but install routines or internal executors that remain callable later. & Capabilities can be installed, invoked, composed, and retained across tasks rather than being relearned or outsourced each time. \\
    \textbf{Behavior-consistent} & Ordinary use should not silently change the machine; behavioral change should come from explicit updates. & Same-version behavior is reproducible; execution and update traces can be inspected, replayed, and rolled back; long-horizon drift is measurable and governable. \\
    \textbf{Machine-native semantics} & The system should not merely imitate conventional computers with neural components, but develop its own machine semantics and programming interfaces. & Composition, routing, continuous state, and internal executors yield usable system-level advantages; prompts, demonstrations, traces, and constraints begin to function as programming interfaces rather than mere logs. \\
    \bottomrule
  \end{tabularx}
\end{table}

\subsection{A Roadmap Towards CNC}
\label{sec:roadmap}

We frame the path toward CNCs through a set of formal requirements together with the practical challenges that must be resolved before those requirements become engineerable.

\paragraph{Turing completeness}
A Neural Computer (NC) instance (a specific architecture with fixed learned weights) defines a class of computational models in which each model corresponds to at least one memory state instance.
In the formal computability discussion below, ``memory state'' is used in the classical state-machine sense; operationally, it corresponds to the NC runtime state introduced earlier.
An NC instance is Turing complete if, for any given Turing machine, there exists an initial memory state that allows the NC to emulate that machine exactly.
Notice that although Recurrent Neural Networks (RNNs), Neural Turing Machines (NTM)~\citep{graves2014neural}, and Differentiable Neural Computers (DNC)~\citep{graves2016hybrid} are Turing complete in the asymptotic sense, a particular RNN, NTM, or DNC instance with finite precision cannot be Turing complete due to their fixed finite memory size.
For an NC instance to achieve universality, unbounded effective memory is necessary. 
An NC instance has unbounded effective memory if there are infinitely many possible memory state instances. 
Existing works approach such unboundedness by progressively growing model parameters~\citep{fritzke1994growing,rusu2016progressive} or context~\citep{vaswani2017attention}.

\paragraph{Universal programmability}
An NC is universally programmable if, for each given Turing machine, there exists an input sequence such that the NC taking this input realizes a new memory state representing the given machine.
Most existing universal programmability results for neural networks are established by constructing computational primitives and proving that their composition can simulate a universal computational model~\citep{reed2015neural}.
Likewise, we believe that universal programmability in NCs can be achieved through compositional neural programs~\citep{pierrot2019learning}.

\paragraph{Behavior consistency}
A CNC must preserve its function unless explicitly reprogrammed. For each memory state, there must be a non-empty set of inputs that executes the CNC without changing its pure function.
Operationally, this requires a separation between run and update: ordinary inputs should execute installed capability without silently modifying it, while behavior-changing updates should occur explicitly through a programming interface.
This in turn motivates training and architectural mechanisms that disentangle function use from function update, so that routines can be installed, executed, and composed without accidental functional drift.
We hypothesize that gating mechanisms, such as those in LSTM \citep{hochreiter1997long}, are effective in achieving this conditional invariance.
In practice, making this separation reliable requires clear boundaries around what state persists across tasks, what counts as an explicit update, and what execution evidence can be replayed, compared, or rolled back.

\begin{tcolorbox}[
  colback=apricotglaze!40!white,
  colframe=metablue!40,
  interior style={shade,shading angle=315,left color=white,right color=apricotglaze!55!white},
  boxrule=0pt,
  borderline west={1pt}{0pt}{gray!60},
  left=6pt,right=4pt,top=3pt,bottom=3pt]
\small
\textbf{Run / update contract.}
\begin{itemize}
  \setlength{\itemsep}{2pt}
  \setlength{\topsep}{2pt}
  \item \textbf{Run:} invoke installed capability without silently changing persistent behavior.
  \item \textbf{Update:} any behavior-changing modification should occur explicitly through a programming interface.
  \item \textbf{Required boundaries:} state (what persists), update (what counts as reprogramming), and evidence (what can be replayed, compared, or rolled back).
\end{itemize}
\end{tcolorbox}

\paragraph{Architectural semantics}
Since NC behavior is governed by real-valued parameters, learning can produce input–output mappings that generalize across variations within the training distribution~\citep{poggio2019theoretical}. For example, after observing many instances of how the visual state of a spreadsheet interface changes when values are typed into cells, a model may learn the underlying transformation and correctly predict the screen updates for previously unseen spreadsheets that follow the same interaction rules. Such in-distribution generalization arises from the smooth function approximation properties of neural networks and their ability to interpolate across previously observed patterns. 
Furthermore, learning can also produce novel input–output mappings that are not explicitly represented in the training data, potentially introducing new computational primitives~\citep{ha2016hypernetworks}. The combination of such newly formed primitives could enable qualitatively new functions, yielding out-of-distribution functional generalization~\citep{lake2018generalization}.

Beyond emulating conventional computers, NCs can natively support functions whose semantics are ill-suited to symbolic APIs~\citep{marcus2018deep}, including probabilistic inference over high-dimensional latent states~\citep{kingma2013auto}, representation learning~\citep{bengio2013representation}, retrieval over dense memories~\citep{graves2016hybrid}, and end-to-end differentiable pipelines that couple perception and control~\citep{silver2017mastering}. These functions are first-class at the architectural level and operate directly on distributed states. This enables capabilities such as learned heuristics~\citep{silver2017mastering}, uncertainty-aware decision-making~\citep{clements2019estimating}, and continual adaptation~\citep{parisi2019continual}.

Because the memory state of an NC/CNC is numerical, computer configuration and design emerge as alternatives to application-level programming: the computer itself is configured by optimizing its internal state to achieve desired computational behaviors under task-defined objectives. 
Depending on the differentiability of the loss, methods such as Adam \citep{adam2014method} and natural evolution strategies\citep{wierstra2014natural} apply. In a CNC, the memory constitutes a continuous manifold, so realizing a target capability amounts to synthesizing a machine configuration (a memory state) that minimizes a user-specified loss (e.g., “minimize proof error”) via direct numerical updates to the computer’s state. This reframes system construction from discrete code authoring to differentiable configuration of the computer itself \citep{innes2019differentiable}, with progress evaluated by solver convergence, stability, and reliability relative to combinatorial program search (e.g., LLM-based code generation \citep{hong2023metagpt}).

\paragraph{Programming-language semantics}
The learned programming-language semantics of NCs enable a shift from rigid coding to learned specifications, in which user inputs themselves function as programs~\citep{brown2020language,wei2022chain}. 
Rather than centering development on explicitly authored code, NCs expose a learned language whose syntax and semantics are acquired from data~\citep{radford2019language,bommasani2021opportunities}, so natural-language instructions, examples, and constraints serve as executable specifications~\citep{ouyang2022training}. 
Consequently, brief user inputs can replace long sequences of low-level actions. 
Development, therefore, moves from code authoring to curating, specifying, and verifying inputs under a learned programming-language semantics, aligning system behavior with human intent via in-context specification rather than forcing users to conform to rigid, brittle interfaces~\citep{austin2021program}.
This does not imply that code disappears, but rather that code becomes one installation medium among several, alongside prompts, demonstrations, trajectories, and constraints.

Since NCs are programmed via users' input sequences under learned programming-language semantics, the training data for programming NCs, i.e., paired user I/O traces~\citep{flener2008introduction}, is far more abundant and continuously generated than high-quality, human-written code.
Every interaction with digital systems produces structured streams of inputs, interface states, and effects that can be logged at scale (e.g., keystrokes, cursor trajectories, screen transitions), yielding orders-of-magnitude more supervision than curated program corpora~\citep{yao2022webshop}. 
These I/O traces constitute executable specifications, revealing user intentions and computer behavior~\citep{cypher1993watch}.
This enables end-to-end learning of interface conventions, control policies, and task semantics without requiring explicit program text~\citep{yao2022react}. This asymmetry in data availability favors NC training regimes that leverage ubiquitous interaction logs and, by supporting broader task coverage, reduces dependence on brittle, sparsely available code datasets.

\subsection{Relations to Other System Objects}
\label{sec:relations}
\begin{figure}[t]
  \centering
  \includegraphics[width=\linewidth]{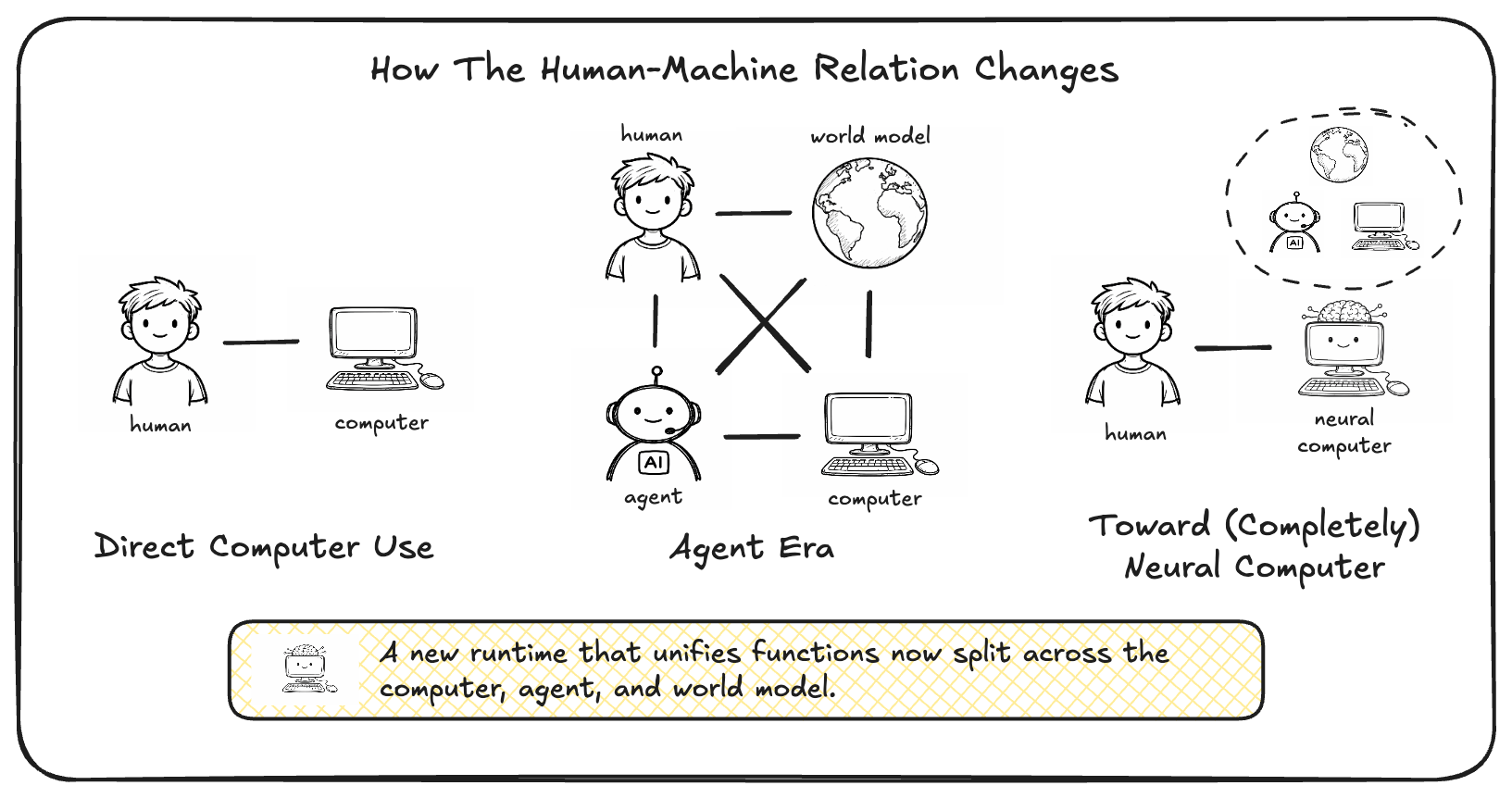}
  \caption{Changing human-machine relations across conventional computers, the agent era, and neural computers. Conventional computers are used directly; in today's agent stack, agents mediate existing computers while world models serve as a parallel predictive layer; NCs aim to unify these split functions within one learned runtime. In this sense, NCs are motivated not by replacing the current stack from the outside, but by internalizing its split functions within one learning machine. A Completely Neural Computer (CNC) is the mature, general-purpose realization of this machine form.}
  \label{fig:relations}
\end{figure}

Figure~\ref{fig:relations} summarizes the systems-level shift: conventional computers are used directly, AI agents mediate existing computers, world models act as a parallel predictive layer, and NCs aim to make the learned runtime itself the machine.

The comparisons below unpack this shift relative to conventional computers, world models, and AI agents.
\paragraph{Conventional Computers}
Conventional computers remain the reference system object for reliable execution, explicit programmability, and mature governance.
NCs differ not by adding a smarter application layer on top of this substrate, but by shifting computation, memory, and I/O into a learned runtime state.
In this sense, NCs are best viewed as a different candidate machine form and computing substrate rather than as a direct extension of the conventional software stack.
This framing does not imply that conventional computers will disappear soon, but rather that future systems may be built from a different underlying runtime substrate.

\paragraph{World Models}
World models learn environment dynamics by predicting action-conditioned transitions~\citep{ha2018world}.
Such target environments range from the most ambitious, where all sensory inputs and control outputs (including agent actions, pain signals, and reinforcement signals) represent the interface to the real world~\citep{schmidhuber1990making}, to much narrower scopes, where just a few control parameters of a robot arm with restricted perceptions are considered~\citep{feng2023finetuning}.
They provide one technical perspective on current NC prototypes, since interactive computers are an important class of action-conditioned environments, but they do not by themselves define the NC abstraction.
Many current approaches to modeling computational environments, such as the physical world, also rely heavily on computer-generated data~\citep{richter2016playing,tobin2017domain,dosovitskiy2017carla,garcia2023robust}, potentially leading to models that share characteristics with neural computers. 

\paragraph{AI Agents}
Another important comparison point is AI agents built on top of modern AI models and external software substrates, including computer-use agents, coding and multi-agent systems~\citep{openaiComputerUsingAgent,claudeComputerTool,hong2023metagpt,zhuge2024gptswarm_icml,sager2025comprehensivesurveyagentscomputer}, and recursive self-improvement loops~\citep{zhuge2026ai_recursive}. 
These systems place a learned agent between the user and an external execution substrate, whether that substrate is a GUI, a codebase, or a broader software toolchain.
This provides strong leverage from existing computers and software stacks, but it also preserves a strict separation between the learned model and the runtime that actually stores executable state, applies updates, and enforces system contracts.
Computer-use agents operate through low-bandwidth I/O; coding agents typically emit symbolic artifacts that must be executed elsewhere; and RSI-style loops improve the agent by iterating over external tools, prompts, or code rather than by turning the runtime itself into the computer.
Such systems also increasingly rely on automated evaluators, including agent-as-a-judge schemes, to rank outputs, validate task completion, and close iterative improvement loops~\citep{zhuge2024agent_judge}.
We hypothesize that a sufficiently capable NC can internalize many of these agentic functions within one persistent neural runtime.

\subsection{Additional Thoughts} 
The remarks in this section are intended as hypotheses and design directions motivated by the present results, rather than as empirical conclusions established by the current prototypes.

\paragraph{ONE}
ONE~\citep{schmidhuber2018one} proposed a single neural substrate that incrementally absorbs and reuses diverse learned skills. A mature CNC can be viewed as a plausible systems-level realization of this idea. In this sense, many specialized world-model-like components may ultimately appear not as separate external systems, but as installable capabilities within one persistent neural runtime.

\paragraph{Video models as a pragmatic prototype substrate}
We build our prototypes on state-of-the-art video models because they currently provide the simplest path to an end-to-end learned latent runtime state that jointly models pixels, dynamics, and action-conditioned control. This choice is pragmatic rather than fundamental. In our experiments, symbolic and algorithmic reasoning in terminal settings remains inconsistent for most strong video models, and even simple arithmetic can fail (\Cref{tab:cligen-arith}). Sora2 is a notable exception in our probe, achieving $71\%$ arithmetic accuracy, suggesting that some terminal symbolic reasoning is already possible in modern video generators. At the same time, we do not claim that video models cannot reason more broadly: recent work reports that video models can act as zero-shot learners and reasoners in naturalistic settings \citep{wiedemer2025video}. We expect reasoning capabilities to improve quickly with continued progress in video modeling, but our results suggest that CNC-level reliability will likely require additional architectural and training ingredients beyond scaling today's video generators.

\paragraph{A hypothesis: machine-native neural architectures}
We emphasize that the following is a conjecture rather than a conclusion drawn from our experiments. Closing the reasoning gap may not require designing neural networks that more closely mimic animal cognition or the human brain.
Many influential architectures, including convolutional networks \citep{fukushima1980neocognitron} and linear/quadratic Transformers \citep{schmidhuber92fastweights,vaswani2017attention}, are highly engineered systems, but their core inductive biases remain strongly influenced by biological perception and attention. These models primarily rely on continuous, distributed representations, in which reasoning behavior emerges implicitly from large-scale training.
We hypothesize that CNCs may instead benefit from designs that are explicitly machine-native. Developing discrete operations, compositional structures, and verifiable computation that are harmonious in neural systems may play an essential role in designing such systems. This approach follows more closely the construction of conventional computers from well-defined computational primitives and stands in contrast to relying on emergent reasoning in generic video generation models.

\paragraph{Neural networks generation via NC interaction}
Neural network generation can be viewed as a form of programming, i.e., the synthesis of a neural architecture and its corresponding weights. Because NCs' architectural semantics are already neural and numerical, neural components are first-class, and generation directly manipulates the memory state rather than translating it into symbolic code. Moreover, NCs can be programmed through I/O interaction: sequences of inputs, observations, and outcomes act as executable specifications that shape the internal state and routines of the system~\citep{cypher1993watch,myers2002demonstrational}. This suggests a path in which users generate and refine neural modules within NCs through interactive traces, treating interaction logs as programs that configure and compose neural computation.

\paragraph{Unified hardware requirements and data representation}
In NCs, tensors and tensor-to-tensor transformations act as primary computational primitives, replacing the heterogeneous mix of data structures and subsystem-specific abstractions common in conventional computers. Traditional systems span many distinct domains—scalars, pointers, linked structures, files, sockets, and processes—each with its own memory layout, invariants, APIs, and failure modes, coordinated by operating systems through largely disjoint subsystems (virtual memory, filesystems, networking, scheduling, and drivers)~\citep{tanenbaum2013structured,silberschatz2019operating}. Although this heterogeneity supports broad generality, it also fragments optimization and tooling because compilers, profilers, and debuggers must reason across incompatible abstractions~\citep{gregg2014systems}. By contrast, a tensor-uniform pipeline concentrates representation and execution into a compact set of composable primitives, such as linear algebra and elementwise operations, allowing tooling to target a shared intermediate representation~\citep{paszke2019pytorch}. As a result, optimizations such as operator fusion, memory planning, and computational-graph rewriting can be applied system-wide~\citep{vasilache2018tensor}; profiling can focus on throughput and memory bandwidth; and accelerators such as GPUs can be targeted through common tensor runtimes~\citep{sze2017efficient}. This shared numerical representation also naturally supports multimodal computation: vision (pixel tensors), language (sequence embeddings), audio (waveforms or spectrograms), control (state-action tensors), and planning (latent trajectory tensors) all reside in one representational space and can be jointly reasoned over and optimized in a single graph~\citep{ramachandram2017deep}, without repeated type bridging or subsystem translation—steps that are substantially harder in traditional heterogeneous stacks.

%% file: section/conclusion.tex
Neural computers point toward a machine form in which a single latent runtime state acts as the computer itself, driving pixels, text, and actions while subsuming what operating systems and interfaces handle today.
In this paper, the main result is that NCs have begun to exhibit early runtime primitives---most notably I/O alignment and short-horizon control---while stable reuse, symbolic reliability, and runtime governance remain unresolved.
Our CNC capability map remains useful as a longer-horizon view, spanning efficiency, computation \& reasoning, memory \& storage, I/O \& control, tool bridges, condition-driven generalization, programmability, and artifact generation.
The map is staged and dependency-informed, but the more immediate gap is still the gap from prototype behavior to usable runtime behavior.
Progress toward CNCs will therefore depend not only on stronger models, but also on whether reuse, consistency, and governance become sustained and testable.
If these gaps continue to close, neural computers will look less like isolated demonstrations and more like a plausible candidate machine form for next-generation computers.

\section*{Acknowledgements}

The authors thank Deyao Zhu and Firas Laakom for their feedback on the manuscript. 
Thanks also to those who provided useful online comments on version v1 of this paper. 
Mingchen Zhuge, Haozhe Liu, Shuming Liu, Wenyi Wang, Wenxuan Zhang, Junjie Fei, Yasheng, and Jürgen Schmidhuber were supported by funding from the King Abdullah University of Science and Technology (KAUST) Center of Excellence for Generative AI (award number 5940) and the SDAIA-KAUST Center of Excellence in Data Science and Artificial Intelligence.

%% file: section/appendix_explorations.tex
\noindent Beyond the data collection pipelines used in the main text and Appendix~\ref{appendix:pipeline}, we explored alternative data sources for neural-computer prototyping.
These routes were not incorporated into the final pipeline, but the trials yielded useful insights and suggest directions for future work as tooling matures and data-collection infrastructure scales.

\begin{figure}[h]
  \centering
  \includegraphics[width=\linewidth]{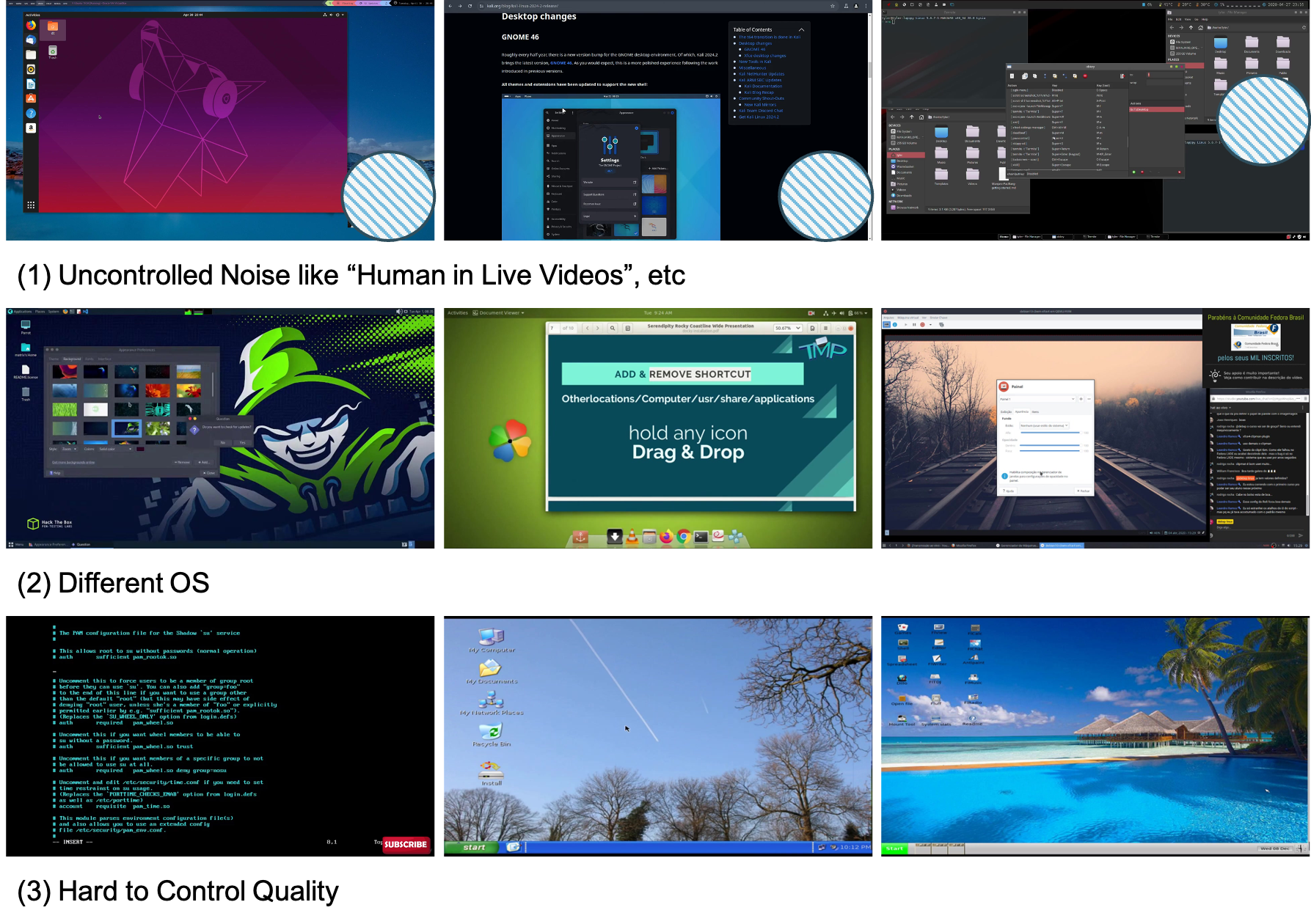}
  \vspace{-0.4em}
  \caption{\textbf{Common issues in web-scale screen video crawling.}
  Web videos mix uncontrolled content, heterogeneous OS/UI configurations, and inconsistent capture quality.
  This makes terminal localization, OCR, and time alignment unreliable without heavy filtering and sanitization.}
  \vspace{-10pt}
  \label{fig:web-video-issues}
\end{figure}

\subsection{Web video extraction}
We initially tested crawling computer-use videos from the web.
We used OCR and layout detectors to locate terminal regions, estimate text content and timestamps, and extract related clips.
We did not adopt this route for two main reasons:
\begin{tcolorbox}[
  colback=apricotglaze!40!white,
  colframe=metablue!40,
  interior style={shade,shading angle=315,left color=white,right color=apricotglaze!55!white},
  boxrule=0pt,
  borderline west={1pt}{0pt}{gray!60},
  left=6pt,right=4pt,top=3pt,bottom=3pt]
\small
\begin{itemize}
  \setlength{\itemsep}{2pt}
  \setlength{\topsep}{2pt}
  \item \textbf{Data governance constraints:} privacy and copyright risks are difficult to manage at web scale.
  \item \textbf{Data quality burden:} heavy filtering and sanitization are required before the data can support reliable OCR and temporal alignment.
\end{itemize}
\end{tcolorbox}
Screen recordings can contain personal identifiers (usernames, emails, file paths, chat content) and may come with licensing constraints that are difficult to verify at scale.
Cleaning the resulting data is substantially more complex than it appears (Figure~\ref{fig:web-video-issues}).
Typical failure modes include (i) uncontrolled content such as faces/hands, picture-in-picture overlays, and unrelated desktop activity;
(ii) domain shift across operating systems, themes, fonts, resolutions, and window managers; and
(iii) quality factors such as compression artifacts, variable frame rates, zoom/crop edits, and inconsistent capture pipelines.
These factors degrade OCR and temporal alignment.

Despite these challenges, web videos remain a potential long-term scaling axis for interface experience.
In this work, however, the cost--quality trade-off was unfavorable.
Our setting benefits disproportionately from clean, temporally aligned text and interaction signals.
Building a high-precision web filter also requires substantial upfront investment.
This includes rights-cleared sourcing or licensing, privacy review and redaction, and large-scale multimodal filtering/OCR pipelines that often rely on paid APIs.
Given these constraints and our emphasis on high-quality supervision, we prioritized curated, rights-cleared interface trajectories.
Future efforts that invest in rights-respecting acquisition and stronger automated filtering could unlock web-scale data as a complementary scaling axis.

\subsection{Online environment interaction}
\begin{figure}[h]
  \centering
  \includegraphics[width=\linewidth]{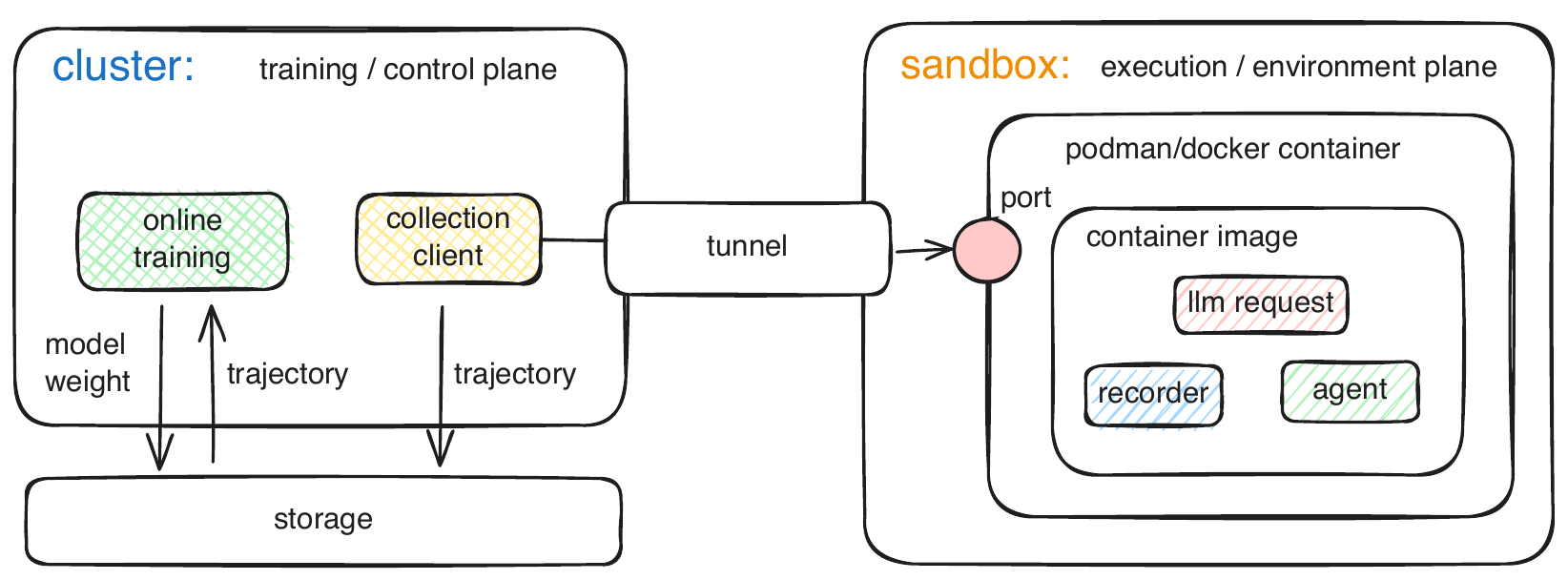}
  \vspace{-1.4em}
  \caption{\textbf{Agentic online interaction pipeline (early exploration).}
  A control plane ingests trajectories for storage and video-model updates (online where feasible, with offline replay support).
  A sandboxed environment plane executes an agent in an isolated container and records synchronized state/action traces.}
  \vspace{-10pt}
  \label{fig:online-interaction-pipeline}
\end{figure}

We prototyped an agentic interaction pipeline that separates a control plane from an execution environment plane (Figure~\ref{fig:online-interaction-pipeline}).
In the environment plane, a sandboxed container runs a live shell together with LLM agents (planner/controller) and a recorder exposed via a narrow port-based interface.
The agents issue commands and control actions.
The recorder captures synchronized terminal renders, structured terminal state when available (e.g., buffer/text), and action traces.
Structured terminal state is logged for diagnostics/alignment and is not fed to video models as privileged state input.
Trajectories are streamed to the control plane for storage and video-model updates.

Concretely, the environment plane exposes a minimal ``step/reset'' interface over a port.
It returns multimodal observations that can be logged deterministically (rendered screenshots plus structured state when available).
The agent emits structured actions (typed command text, key/mouse events when applicable, and timing).
This separation makes rollouts auditable and lets the control plane scale rollout collection and model updates with decoupled throughput.
From the perspective of Toward CNC, this design also makes the evidence boundary more explicit by preserving replayable execution traces with clear provenance across collection and training.

We explored this setup because closed-loop interaction can induce a natural curriculum by continually sampling the boundary of current video-model behavior.
It can also surface rare and safety-critical failure modes that do not appear in offline logs.
It supports targeted data collection (e.g., focusing on specific tools, error recovery, or long-horizon tasks).
In principle, it also offers a direct path to scaling \emph{experience} rather than only scaling static demonstrations.

Early trials showed promise, but the dominant bottlenecks in the end-to-end system were systems-level: cross-cluster communication latency between rollout workers and training nodes, and high debugging complexity in asynchronous distributed execution.
Safety controls (e.g., isolation of untrusted code execution, monitoring and abuse prevention, and deterministic resets and environment control) remained necessary constraints, but they were not the primary bottleneck.
The setup also requires robust recording and serialization across heterogeneous environments.
Under time and cost constraints, we therefore prioritized controlled video data for the main experiments. 
Despite not being used in the final pipeline, we consider this design a useful systems template for future work.
It supports scalable multi-environment rollout collection and consistent provenance and storage of trajectories.
It can support multiple downstream learning algorithms (e.g., behavior cloning or preference-based learning).
Execution remains sandboxed and auditable.

%% file: section/appendix_pipeline.tex
\noindent This appendix summarizes collection, preprocessing, and evaluation details for the datasets used in the paper.
For concrete examples of raw trajectory formats (asciinema \texttt{.cast} and \texttt{vhs} scripts), see Appendix~\ref{appendix:cligen-samples}. 
Data collection uses a three-stage pipeline to maintain synchronized timing, privacy controls, and consistent, well-documented artifacts across CLIGen and GUIWorld.

\noindent\textbf{Sourcing.}
We construct our datasets from three complementary sources---public terminal recordings, scripted terminal replays, and a controlled desktop-capture rig.
Across all sources, we emphasize rights-respecting acquisition, privacy filtering, and temporally aligned interface signals (frames, actions, and text when available):
\begin{tcolorbox}[
  colback=apricotglaze!40!white,
  colframe=metablue!40,
  interior style={shade,shading angle=315,left color=white,right color=apricotglaze!55!white},
  boxrule=0pt,
  borderline west={1pt}{0pt}{gray!60},
  left=6pt,right=4pt,top=3pt,bottom=3pt]
\small
\begin{itemize}
  \setlength{\itemsep}{2pt}
  \setlength{\topsep}{2pt}
  \item \cligenGeneralLogo{}\,\textbf{CLIGen (General)}: public \texttt{asciinema} \texttt{.cast} archives; traces are replayed with official tools to preserve recorded terminal appearance (color schemes, cursor visibility, window geometry).
  \item \cligenCleanLogo{}\,\textbf{CLIGen (Clean)}: deterministic \texttt{vhs} scripts (e.g., package installs, REPLs, log filters) executed in isolated environments.
  \item \guiworldLogo{}\,\textbf{GUIWorld}: footage captured with the rig in \Cref{section:impl-guiworld}, pairing RGB video with low-latency pointer/key logs and optional accessibility cues (logged for analysis; not used as model inputs).
\end{itemize}
\end{tcolorbox}

\noindent\textbf{Alignment and sanitization.} 
All modalities share a common clock.
We align pointer/key events to the nearest frame, apply drift correction when needed, and drop clips with residual misalignment.
Privacy filters remove terminal sessions with sensitive strings and redact GUI regions likely to contain private content.
Frozen or repeated-frame recordings (capture artifacts) are discarded.

\noindent\textbf{Episode packaging.}
Runs are windowed into fixed-length, overlapping episodes (window sizes and strides are specified in the released configs).
Each shard stores RGB frames, terminal buffers or GUI metadata, serialized actions, the source tool (\texttt{asciinema}/\texttt{vhs}/GUI capture), and environment metadata.
Structured fields (buffers/metadata) are used for alignment and evaluation, but are not provided to the video models as state inputs.
Downstream dataloaders reconstruct batches directly.
Released configs specify preprocessing and windowing so external users can rebuild the corpus.
This packaging turns each episode into a replayable artifact with provenance, which is useful not only for evaluation reproducibility but also for the broader evidence and governance requirements discussed in Toward CNC.

\subsection{Caption fields and metadata (CLIGen General)}
For CLIGen (General), each replayed \texttt{.cast} fragment (example shown in \Cref{appendix:cast-format}) is paired with three aligned descriptions and a compact metadata record.
Table~\ref{tab:cast-captions} summarizes the fields for clip \texttt{7****\_0001}.

\begin{smallschemetable}{Caption tiers and metadata for CLIGen (General) clip \texttt{7****\_0001}.}{tab:cast-captions}
  \rowcolors{2}{tablewarmrow}{white}
  \begin{tabularx}{\linewidth}{@{}l X@{}}
    \toprule
    \rowcolor{tablewarmhead}\textbf{Field} & \textbf{Content (abridged)} \\
    \texttt{caption} &
      The user (username field shown verbatim as \texttt{annonomous} [sic]) logged into their account on the host \texttt{NeuralComputer}, and after some initial terminal setup and cursor movements,
      they started typing the command \texttt{nvim} to likely open the Neovim text editor, but the session recording ends abruptly without showing the actual command execution or any further interaction. \\
    \texttt{caption\_detailed} &
      At the \texttt{annonomous@NeuralComputer:~\$} prompt, the user types \texttt{nvim}, but before the command executes, the screen rapidly displays a colorful,
      pixelated animation with various RGB colors, including shades of blue, green, and purple, filling the 80x24 terminal window.
      The animation is briefly interrupted by a cursor blink, then the prompt returns, awaiting further input. \\
    \texttt{caption\_semantic} &
      The terminal displays a rapid sequence of colored text and pixel art animations, with syntax-highlighted code and prompts appearing in a mix of blue, green,
      and white hues, as the user types commands and the system responds with scrolling output, error messages, and success indicators. \\
    \midrule
    \texttt{data\_info.version} & \texttt{2} (\texttt{asciinema} v2 header) \\
    \texttt{data\_info.size} & \texttt{width=80}, \texttt{height=24}; original/target/scaled sizes all match, padding \texttt{[0,0]}. \\
    \texttt{data\_info.env} & \texttt{SHELL=/bin/bash}, \texttt{TERM=xterm-256color}. \\
    \texttt{meta.videogen} &
      Automatically derived stats: visual complexity \texttt{2897.7}, interaction density \texttt{1.0}, color usage \texttt{518}, screen clears \texttt{6},
      event rate \texttt{7.8} events/s, typing rhythm (avg interval \texttt{0.13}s, variance \texttt{0.34}), and \texttt{38} visual-change events. \\
    \texttt{metadata} &
      Source recording on Asciinema: id \texttt{7****}, title \texttt{``aneo.nvim demo''}, author \texttt{annonomous} [sic], creation time \texttt{2025-05-13T23:37:53Z}, URLs for the page and raw \texttt{.cast}. \\
    \bottomrule
  \end{tabularx}
\end{smallschemetable}

These fields make each CLIGen (General) clip self-contained.
The three caption tiers provide prompts at different levels of detail.
\texttt{data\_info} and \texttt{metadata} preserve the structure needed to rebuild terminal geometry, environment, and source from the raw \texttt{.cast}. 

\subsection{OCR evaluation protocol (CLIGen Clean)}
For CLIGen (Clean), OCR-based metrics evaluate how closely generated terminal videos match reference renderings derived from the ground-truth buffers in text space rather than pixels.
Each sample consists of a generated video and its paired reference video (matched by clip ID).
We keep only IDs where both videos are present. 
From each paired video we use at most $K{=}5$ frames.
Let $T_\text{gen}$ and $T_\text{gt}$ be the frame counts of the generated and reference videos.
We set $T=\min(T_\text{gen},T_\text{gt})$.
If $T \le K$, we use all indices in $[0, T{-}1]$; otherwise, we select $K$ evenly spaced indices by deterministic rounding, then deduplicate and sort them so evaluation frames are spread across the trajectory.
For every selected index we read the corresponding frame from both videos.

Each frame is converted to RGB and passed to Tesseract OCR.
The resulting string is split into lines, leading and trailing whitespace is stripped, and internal whitespace is normalized by collapsing runs of spaces.
Empty lines are dropped.
We keep case and punctuation intact so that commands, paths, and symbols remain visible.
This gives an ordered list of normalized lines for the ground-truth frame ($g_1,\dots,g_{N_g}$) and the generated frame ($p_1,\dots,p_{N_p}$).

We summarize the OCR text-space metrics used per sampled frame as follows:
\begin{tcolorbox}[
  colback=apricotglaze!40!white,
  colframe=metablue!40,
  interior style={shade,shading angle=315,left color=white,right color=apricotglaze!55!white},
  boxrule=0pt,
  borderline west={1pt}{0pt}{gray!60},
  left=6pt,right=4pt,top=3pt,bottom=3pt]
\small
\textbf{Character accuracy.} This metric pools all lines into a single multi-line string for each side and measures normalized edit distance.
Let $s$ and $t$ be the concatenated ground-truth and generated texts and $d(s,t)$ their Levenshtein distance (insert/delete/replace cost $1$).
If both $s$ and $t$ are empty we set $\text{char\_acc}=1$; if only $s$ is empty we set $\text{char\_acc}=0$.
Otherwise,
\[
  \text{char\_acc} = \max\Bigl(0,\ 1 - \frac{d(s,t)}{\max(|s|,1)}\Bigr),
\]
Extra or missing characters are normalized by the reference length through the denominator $\max(|s|,1)$.
Frame-level scores are averaged over the selected frame pairs (up to $K{=}5$) to yield a per-video character accuracy, and group-level scores report the mean over videos.

\textbf{Exact-line accuracy.} This metric treats lines as position-sensitive units and reports a recall over ground-truth lines.
For a given frame, we compare line $g_i$ to $p_i$ at the same index.
A line is counted as correct only if $i\le N_p$ and $p_i=g_i$; lines that appear in the wrong position do not count.
If both lists are empty we set $\text{exact\_line\_acc}=1$; if the ground-truth list is empty but the generated list is not, we set $\text{exact\_line\_acc}=0$.
Otherwise,
\[
  \text{exact\_line\_acc} = \frac{1}{N_g}\sum_{i=1}^{N_g} \mathbf{1}[i \le N_p \land p_i = g_i].
\]
As with character accuracy, frame scores are averaged over the $K$ sampled frames to obtain a per-video score and then averaged over videos for the reported aggregate.
\end{tcolorbox} 
Together, these two metrics stress both fine-grained text fidelity and line-ordered terminal state reconstruction.

\subsection{Evaluation metrics and protocol (GUIWorld)}
\label{appendix:gui-metrics}
We report both global video metrics and action-driven metrics that focus on post-interaction frames. 
We compute these metrics using our GUIWorld evaluation suite.
We summarize the GUIWorld protocol at a glance as follows:
\begin{tcolorbox}[
  colback=apricotglaze!40!white,
  colframe=metablue!40,
  interior style={shade,shading angle=315,left color=white,right color=apricotglaze!55!white},
  boxrule=0pt,
  borderline west={1pt}{0pt}{gray!60},
  left=6pt,right=4pt,top=3pt,bottom=3pt]
\small
\begin{itemize}
  \setlength{\itemsep}{2pt}
  \setlength{\topsep}{2pt}
  \item \textbf{Global metrics} (\FVD$_\text{all}$, \SSIM$_\text{all}$, \LPIPS$_\text{all}$): computed over paired generated/ground-truth videos after standardized decoding, subsampling, and resizing.
  \item \textbf{Action-driven metrics} (\SSIM$_{+15}$, \LPIPS$_{+15}$, \FVD$_{+15}$): computed on post-action windows to measure interface fidelity after interaction events.
\end{itemize}
\end{tcolorbox}

\noindent\textbf{Global \FVD$_\text{all}$/\SSIM$_\text{all}$/\LPIPS$_\text{all}$}
We decode paired generated/ground-truth videos into RGB frames with temporal subsampling and resizing (\texttt{fps}=3, \texttt{size}=256, and \texttt{max\_seconds}=5 by default).
\SSIM$_\text{all}$~is computed using \texttt{torchmetrics} on frame tensors normalized to $[0,1]$ and averaged over frames.
\LPIPS$_\text{all}$~uses the AlexNet backbone on frames normalized to $[-1,1]$ and is averaged over frames.
\FVD$_\text{all}$~is computed in an \texttt{r3d18} embedding space (\texttt{prelogits} by default).
We extract features from fixed-length clips (16 frames at 112$\times$112 after uniform subsampling/padding).
We compute the Fr\'echet distance between the generated and reference feature distributions.

\noindent\textbf{Action-driven metrics (\SSIM$_{+15}$, \LPIPS$_{+15}$, \FVD$_{+15}$)}
For each paired rollout, we load recorded action timestamps (from JSON/CSV logs), map each timestamp $\tau$ to a frame index $f=\mathrm{round}(\tau \cdot \mathrm{fps})$, and clamp to the valid frame range.
We skip the action frame itself (\texttt{action\_start\_offset}=1) and evaluate the next $k{=}15$ frames after each action.
Concretely, for each action frame index $f$, we form the post-action set $\{f+1,\dots,f+k\}$ and clip it to valid frame indices.
We then take the deduplicated union over actions in the clip and keep frame indices in chronological order.
Clips with zero logged actions, or with empty valid post-action windows after clipping, are excluded from $+15$ metrics.
For \SSIM$_{+15}$~/\LPIPS$_{+15}$, we compute per-clip means over selected frame pairs and then average across clips.
For action-driven \FVD$_{+15}$, we build an \emph{after-action clip} per video by concatenating the same selected post-action frames, then uniformly subsample/pad to 16 frames and compute the Fr\'echet distance in the same \texttt{r3d18} feature space.

%% file: section/appendix_cligen.tex
\noindent This appendix provides concrete format examples for the two CLI sources referenced throughout the data pipeline (Appendix~\ref{appendix:pipeline}).
We show asciinema \texttt{.cast} trajectories for CLIGen (General) and \texttt{vhs} scripts for CLIGen (Clean).
\subsection{Asciinema (\texttt{.cast}) example}
\label{appendix:cast-format}
The header line stores the recording config (version, terminal size, timestamp, env).
In this excerpt, output rows follow \texttt{[time, "o", "payload"]}, where \texttt{"o"} indicates screen output.
The payload contains the terminal text with color codes at that timestamp.
\begin{verbatim}
{"version": 2, "width": 80, "height": 24,
 "timestamp": 1747177906,
 "env": {"SHELL": "/bin/bash", "TERM": "xterm-256color"}}
[0.082492, "o", "\u001b[H\u001b[2J\u001b[3J"]
[0.950038, "o", "\u001b[38;2;16;131;236m\u001b[39m\r\n..."]
[0.950733, "o", "\u001b[38;2;6;156;220m ... \u001b[38;2;1;195;187m█"]
\end{verbatim}

\subsection{VHS script example}
\label{appendix:vhs-format}
\begin{verbatim}
# ---- VHS documentation start (DO NOT CHANGE) ----
# Require:
#   Require <string>
# Sleep:
#   Sleep <time>
# Type:
#   Type[@<time>] "<characters>"
# Keys:
#   Escape[@<time>] [number]
#   Backspace[@<time>] [number]
#   Delete[@<time>] [number]
#   Insert[@<time>] [number]
#   Down[@<time>] [number]
#   Enter[@<time>] [number]
#   Space[@<time>] [number]
#   Tab[@<time>] [number]
#   Left[@<time>] [number]
#   Right[@<time>] [number]
#   Up[@<time>] [number]
#   PageUp[@<time>] [number]
#   PageDown[@<time>] [number]
#   ctrl+<key>
# Display:
#   Hide
#   Show
# ---- VHS documentation end (DO NOT CHANGE) ----

# ID: vhs_example
# INSTRUCTION: Runs `uname -s` repeatedly as a basic shell exercise, then hides the prompt.
# LEVEL: 1
# EVENTS: 23
# VISUAL_COMPLEXITY: 45

# ---- Theme setting start (DO NOT CHANGE) ----
Output vhs_example.mp4

Set Shell "bash"

Set Theme {
  "name": "Catppuccin Mocha (Pure White, Warm Pink Cursor)",
  "background": "#1e1e2e",
  "foreground": "#ffffff",
  "black": "#45475a",
  "red": "#f38ba8",
  "green": "#a6e3a1",
  "yellow": "#f9e2af",
  "blue": "#89b4fa",
  "purple": "#cba6f7",
  "cyan": "#94e2d5",
  "white": "#ffffff",
  "brightBlack": "#585b70",
  "brightRed": "#f38ba8",
  "brightGreen": "#a6e3a1",
  "brightYellow": "#f9e2af",
  "brightBlue": "#89b4fa",
  "brightPurple": "#cba6f7",
  "brightCyan": "#89dceb",
  "brightWhite": "#ffffff",
  "cursor": "#f5c2e7",
  "cursorAccent": "#1e1e2e",
  "selectionBackground": "#585b70"
}

Set FontSize 40
Set Width 1600
Set Height 900
Set TypingSpeed 300ms
Set PlaybackSpeed 1
Set Margin 28
Set MarginFill "#0091FF"
Set BorderRadius 25
Set Padding 18
Set LineHeight 1.2
Set LetterSpacing 0.8
# ---- Theme setting end (DO NOT CHANGE) ----

Sleep 800ms
Sleep 180ms
Type "uname -s"
Sleep 120ms
Enter
Sleep 400ms
Type "uname -s"
Sleep 120ms
Enter
Sleep 400ms
Type "uname -s"
Sleep 120ms
Enter
Sleep 400ms
Type "uname -s"
Sleep 120ms
Enter
Sleep 400ms
Type "uname -s"
Sleep 120ms
Enter
Sleep 400ms
Sleep 400ms
Sleep 600ms
Hide
\end{verbatim}

%% file: section/appendix_gui.tex
\noindent This appendix provides additional technical details on GUIWorld action representation, temporal alignment, and conditioning used in Section~\ref{section:impl-guiworld}.
Additional visualization pages are collected in Appendix~\ref{appendix:vis}; evaluation metrics and protocols are summarized in Appendix~\ref{appendix:gui-metrics}.

\subsection{Action schema}
\ncguiworld{} represents actions as a structured stream, enabling the NC to condition on both cursor movements and key presses.

At each timestep, we log absolute cursor coordinates, button up/down transitions, scroll deltas, and keyboard events.
Keyboard inputs are split into two types: typed characters (e.g., \texttt{ls -l}) and shortcut-style chords (e.g., \texttt{ctrl+v}).
We also track state flags such as whether a drag is currently active.
This lets us represent extended interactions like click-drag or press-hold as short labeled segments rather than isolated spikes.
The meta-action encoder described in Section~\ref{section:impl-guiworld} compresses this stream into a small typed schema.
In all reported v2 experiments, we use $S{=}2$ action slots per frame; empty slots are padded with type~0.
Each action has a type (e.g., \texttt{mouse click} or \texttt{keyboard type}) plus parameters.
Table~\ref{tab:meta-action-schema} summarizes the types and fields.
Type~0 corresponds to the absence of an action.
Type~1 encodes mouse clicks and drags via button identity, click count, and a drag flag.
Type~2 captures scrolls with a direction and scalar amount.
Type~3 packages free-form keyboard text (such as \texttt{ls -l}) embedded by the shared text encoder.
Type~4 records shortcuts such as \texttt{ctrl+v} via a small shortcut vocabulary.
This representation resembles a tool API while remaining recoverable from raw logs.

\begin{table}[h]
  \centering
  \small
  \setlength{\tabcolsep}{6pt}
  \renewcommand{\arraystretch}{1.05}
  \caption{Meta-action schema for GUIWorld (per action slot).}
  \label{tab:meta-action-schema}
  \rowcolors{2}{tablewarmrow}{white}
  \begin{tabularx}{\linewidth}{@{}c l >{\raggedright\arraybackslash}X@{}}
    \toprule
    \rowcolor{tablewarmhead}\textbf{Type id} & \textbf{Action} & \textbf{Parameter fields} \\
    0 & None & -- \\
    1 & Mouse Click/Drag & \texttt{button}, \texttt{click\_count}, \texttt{drag\_flag} \\
    2 & Mouse Scroll & \texttt{direction}, \texttt{amount} \\
    3 & Keyboard Type & \texttt{text} (e.g., \texttt{ls -l}) $\rightarrow$ shared text encoder \\
    4 & Shortcut & \texttt{shortcut\_id} (e.g., \texttt{ctrl+v}) \\
    \bottomrule
  \end{tabularx}
\end{table}

\subsection{Conditioning: encoders and injection}
The main text considers two encoders for this stream.
A \emph{raw-action encoder (v1)} keeps fine-grained mouse and key events in a multi-hot representation that closely mirrors real cursor and typing behavior.
A complementary \emph{meta-action encoder (v2)} compresses events into a small typed schema (Table~\ref{tab:meta-action-schema}) and embeds any free-form text with a shared text encoder.
Both encoders produce per-frame action features that undergo temporal windowing and alignment (described below).
These embeddings support four injection modes, summarized below:
\begin{tcolorbox}[
  colback=apricotglaze!40!white,
  colframe=metablue!40,
  interior style={shade,shading angle=315,left color=white,right color=apricotglaze!55!white},
  boxrule=0pt,
  borderline west={1pt}{0pt}{gray!60},
  left=6pt,right=4pt,top=3pt,bottom=3pt]
\small
\begin{itemize}
  \setlength{\itemsep}{2pt}
  \setlength{\topsep}{2pt}
  \item \texttt{external}: fuse actions at the VAE input.
  \item \texttt{contextual}: mix actions and frames as tokens in one sequence.
  \item \texttt{internal}: inject actions inside transformer blocks.
  \item \texttt{residual}: add lightweight action deltas to hidden states.
\end{itemize}
\end{tcolorbox}

\begin{figure}[t]
  \centering
  \includegraphics[width=0.5\linewidth]{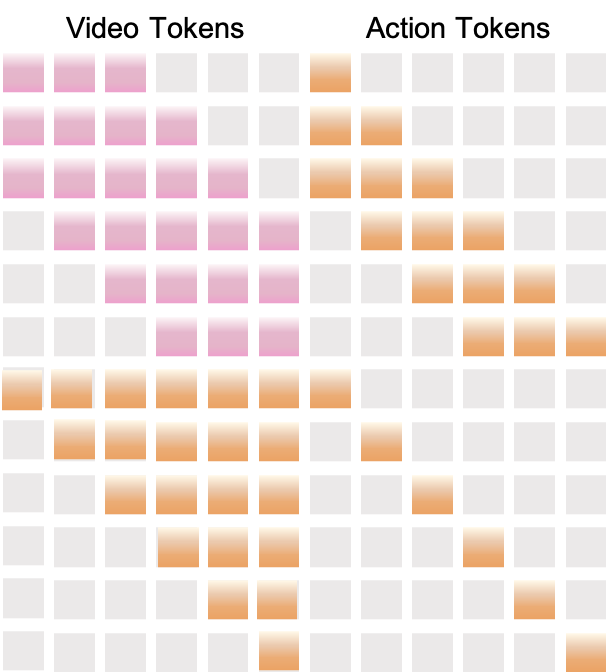}
  \caption{Contextual attention mask for GUIWorld.
  Video tokens (pink) and action tokens (orange) share one sequence.
  The mask restricts attention to a short temporal window so each frame attends only to nearby frames and temporally aligned actions.}
  \label{fig:contextual-mask}
\end{figure}

\noindent\textbf{Injection-mode definitions (schematic)}
\label{appendix:gui-injection-equations}
Below we give compact schematic definitions for the three formula-based modes (\texttt{external}, \texttt{residual}, \texttt{internal}); \texttt{contextual} is specified by the structured attention mask in Figure~\ref{fig:contextual-mask}.
\textbf{External.} Given VAE latents $z_{1:T}$ and temporally aligned action features $u_{1:T}$, an external action module produces a residual update $\Delta z_{1:T}(u_{1:T})$ and forms modified latents
\[
  z'_{1:T} = z_{1:T} + \Delta z_{1:T}(u_{1:T}).
\]
The diffusion backbone then operates on $z'_{1:T}$ (actions do not appear as explicit tokens inside the transformer).
\textbf{Residual.} At selected transformer layers $l$, an auxiliary action module takes block hidden states $h^{(l)}$ together with local action/mouse features and outputs a residual update $\Delta h^{(l)}(a,\text{mouse})$.
The updated hidden states are
\[
  \tilde{h}^{(l)} = h^{(l)} + \Delta h^{(l)}(a,\text{mouse}),
\]
which are passed to the next block.

\textbf{Internal.} At selected blocks, action conditioning is inserted as an additional cross-attention sub-layer inside the standard attention stack.
With self-attention $\mathrm{SA}$, text/reference cross-attention $\mathrm{CA}_{\text{text}}$, and action cross-attention $\mathrm{CA}_{\text{action}}$, a schematic update is
\[
  h' = \mathrm{FFN}\Big(
    h + \mathrm{CA}_{\text{text}}\big(\mathrm{SA}(h), c\big)
      + \mathrm{CA}_{\text{action}}(h, a)
  \Big).
\]

\subsection{Temporal alignment and attention}
\noindent\textbf{Temporal alignment and windows.}
The GUI backbone processes a compressed latent video at stride $c$ (every $c$ pixel frames correspond to one latent frame).
For a pixel sequence of length $F$ and latent sequence of length $T$, we approximately have $F \approx (T-1)c + 1$ under uniform sampling.
Exact indexing follows the dataloader timestamp mapping and boundary handling.
Anchor frame $a_t = t \cdot c$ marks the pixel frame corresponding to latent step $t$.

Mouse and keyboard logs start as per-frame features $r_f \in \mathbb{R}^D$ at the pixel rate.
A windowed encoder aggregates them around each anchor over $p = c \cdot w$ frames.
Here $w$ controls the window width, and a lag $\ell$ accounts for GUI response delay (actions precede their visual effects).
We use zero-padding outside the valid range, i.e., $\tilde{r}_f=r_f$ for $0\le f<F$ and $\tilde{r}_f=0$ otherwise, and form a lag-shifted window that ends at $a_t-\ell$:
\[
  W_{t,k} = \tilde{r}_{\,a_t - (p-1+\ell) + k}, \quad k\in\{0,\dots,p-1\}, \quad
  a^{\text{act}}_t = \frac{1}{p} \sum_{k=0}^{p-1} W_{t,k}.
\]
This shared action encoder yields one latent-aligned action embedding $a^{\text{act}}_t$ per step.
It summarizes a short, lagged history of cursor motion and key events and is reused across all injection modes.

\noindent\textbf{Contextual attention mask.}
In the \texttt{contextual} mode, video and action tokens are concatenated into a single sequence and processed under a structured lag-aware local attention mask.
Appendix Figure~\ref{fig:contextual-mask} can be read as a query--key matrix: rows are queries and columns are keys.

The upper-left block (\emph{V2V}) restricts each frame $V_i$ to attend only to neighboring frames within a window of $\pm w$ steps, so very distant frames cannot interfere.
The upper-right block (\emph{V2A}) lets frame $V_i$ see only recent actions in a lag-bounded recent-action range.
In implementation, this window is $j \in [\max(0, i-\ell), \min(i, A-1)]$, where $\ell$ is the action lag and $A$ is the action-token length.
This way, frame conditioning stays focused on recent operations and excludes future actions.
In the lower-left block (\emph{A2V}), an action $A_i$ can attend to frames $V_t$ that occur after it has had time to take effect ($t \ge i+\ell$, with boundary clipping), but not to earlier frames.
This path is representation-only and does not expose future frame information to frame prediction.
The lower-right block (\emph{A2A}) is strict diagonal: each action token attends to itself.

In practice $(w,\ell)$ act as fixed hyperparameters that trade off temporal coverage and cost.
Together, these choices implement a structured lag-aware temporal prior: actions do not explain past frames, and each frame conditions on recent operations that could plausibly have shaped its pixels.

\noindent\textbf{Design insights.} Two insights from the GUI experiments motivate this schema.
First, raw action streams are bursty and high-dimensional.
Cursor and key events arrive in short spikes, and simple smoothing or full-history attention can cause false interpolated motion and underestimated typing speed.
Using short, lagged windows and local attention bands makes credit assignment more intuitive: each frame connects to the few operations that could have produced it.
Second, in our experiments, with the visual backbone fixed, control fidelity improved more from conditioning design than from encoder choice.
Clean, well-paced supervision and mid- or deep-level action injection improve cursor accuracy and hover timing, while different encodings of the same stream perform similarly.
This action schema and mask implement these principles: keep pixels and actions aligned in time, prioritize recent operations over distant ones, and use attention structure rather than~capacity~alone.

\subsection{Cursor rendering and supervision}
\noindent\textbf{Cursor rendering and reference construction.}
\label{appendix:cursor-rendering} 
The cursor pipeline applies the same design principles on the visual side.
Instead of relying on the global diffusion loss to recover a small, high-frequency visual target, we render the cursor explicitly.
We treat it as a first-class conditioning signal.

\noindent\textbf{From logs to normalized trajectories.}
\label{appendix:mouse-traj-normalization}
Desktop logs provide per-frame cursor positions in screen coordinates $(x_\text{screen},y_\text{screen})$ at the native GUI resolution.
We align these with sampled video frames using the same letterbox mapping as the RGB stream.
Given source and target resolutions $(w_\text{src}, h_\text{src})$ and $(w_\text{dst}, h_\text{dst})$, we compute a uniform scale $s$ and padding offsets $(p_x,p_y)$.
Each coordinate is then mapped to normalized positions $(x_t,y_t)\in[0,1]^2$ as
\[
  x_t = \frac{s\,x_{\text{screen},t} + p_x}{w_\text{dst}-1},\qquad
  y_t = \frac{s\,y_{\text{screen},t} + p_y}{h_\text{dst}-1}.
\]
Stacking these over time yields the trajectory tensor \texttt{mouse\_trajectories} used across rendering and action encoding.

\begin{table}[h!]
  \centering
  \small
  \setlength{\tabcolsep}{4pt}
  \renewcommand{\arraystretch}{1.12}
  \caption{Raw-action versus meta-action encoders in GUIWorld.}
  \label{tab:action-encoder-comparison}
  \begin{tabularx}{\linewidth}{p{2.9cm} >{\raggedright\arraybackslash}X >{\raggedright\arraybackslash}X}
    \toprule
    \textbf{Aspect} &
    \textbf{Raw-action encoder (v1)} &
    \textbf{Meta-action encoder (v2)} \\
    \midrule
    \textbf{Action schema} &
    Per-frame multi-hot vector composed of
    13 mouse actions and 169 keyboard actions;
    no explicit type hierarchy. &
    Hierarchical schema with explicit action types and parameters:
    \texttt{action\_types} $\in \{0,1,2,3,4\}$ for
    \{None, Mouse Click/Drag, Mouse Scroll, Keyboard Type, Keyboard Shortcut\},
    plus typed parameter fields (e.g., button, scroll amount, shortcut ID, text). \\[2pt]
    \textbf{Mouse representation} &
    Two signals per frame:
    (1) continuous cursor trajectory
    $\texttt{mouse\_trajectories}_{t} \in \mathbb{R}^2$;
    (2) discrete mouse events
    $\texttt{mouse\_action\_events}_{t} \in \{0,1\}^{13}$ (multi-hot). &
    Per-frame mouse trajectory
    $\texttt{mouse\_trajectories}_{t}$
    is encoded by a shared mouse-trajectory encoder and the temporal windowing module into
    dense mouse embeddings
    $\texttt{mouse\_latent}_{t} \in \mathbb{R}^{d_{\text{mouse}}}$,
    optionally fused with action embeddings. \\[2pt]
    \textbf{Keyboard representation} &
    Per-frame keyboard state
    $\texttt{keyboard\_action\_events}_{t} \in \{0,1\}^{169}$,
    where each dimension corresponds to a key or shortcut;
    only multi-hot activation is available, no text semantics. &
    Two complementary forms:
    keyboard shortcuts as
    \texttt{keyboard\_shortcut} IDs embedded via an embedding table, and
    free-form text \texttt{keyboard\_text} (e.g.\ ``ls -l'')
    encoded by a shared text encoder (e.g., T5) and projected to the
    action embedding dimension. \\[2pt]
    \textbf{Per-frame representation} &
    Mouse and keyboard events are concatenated:
    $\texttt{raw\_actions}_{t}
      = [\,\texttt{mouse\_events}_{t},\,
           \texttt{keyboard\_events}_{t}\,]
      \in \mathbb{R}^{182}$. &
    For each frame $t$ and slot $s$,
    type plus parameters are embedded into
    \texttt{slot\_embeds}$_{t,s} \in \mathbb{R}^{D}$.
    Valid slots are aggregated (mean or attention) into a single
    per-frame action embedding
    $\texttt{action\_embeds}_{t} \in \mathbb{R}^{D}$. \\[2pt]
    \textbf{Multi-slot support} &
    No explicit notion of multiple action slots per frame;
    all mouse and keyboard signals are collapsed into one
    182-d multi-hot vector. &
    Explicit multi-slot design:
    \texttt{action\_types} and all parameter tensors have shape
    $(B, T, S)$ with $S=2$ slots per frame,
    enabling, e.g., concurrent ``mouse click + keyboard shortcut''
    in the same frame. \\[2pt]
    \textbf{Latent-aligned representation} &
    Temporal alignment is implemented ad-hoc inside each
    v1 action module (e.g., sliding windows over the
    182-d per-frame vector),
    with custom logic per mode (external, internal, residual). &
    A unified temporal alignment module takes per-frame action embeddings
    $(B, F, D)$ and VAE compression ratio $c$,
    constructs lag-aware temporal windows of size
    $c \cdot w$,
    and outputs latent-aligned action features
    $\texttt{action\_latent} \in \mathbb{R}^{B \times T \times D}$,
    with analogous $\texttt{mouse\_latent}$ when mouse features are enabled. \\[2pt]
    \textbf{Lag modeling} &
    GUI latency / response lag is implicitly encoded
    by where and how the sliding window is applied
    inside each v1 action module; there is no shared
    lag configuration across modes. &
    Lag is an explicit hyperparameter
    \texttt{action\_lag} in the temporal alignment and in
    the contextual attention mask:
    it shifts temporal windows and defines which frames
    an action can attend to, giving a consistent lag interpretation across modes. \\[2pt]
    \textbf{Downstream usage} &
    Raw 182-d vectors (plus trajectories) are fed directly into the v1 action modules (external, internal, residual) without a shared temporal encoder. &
    The v2 encoder outputs provide a shared conditioning stream for all v2 modes and also supply features for the temporal contrastive loss. \\[2pt]
    \textbf{Contrastive supervision} &
    No dedicated temporal contrastive loss between frames
    and actions; supervision comes only from the generative objective. &
    Action and mouse latents are coupled to frame features via
    an InfoNCE-style temporal contrastive loss, encouraging time-aligned action embeddings. \\
    \bottomrule
  \end{tabularx}
\end{table}

\subsection{Training signals and encoder design}

\noindent\textbf{From trajectories to cursor layers.}
Starting from normalized $(x_t,y_t)$ coordinates, a cursor-layer module renders a fixed SVG arrow template into RGB and alpha channels.
The template is reused across frames.
For each timestep $t$, it is positioned so the hotspot (arrow tip) aligns with $(x_t,y_t)$, clipped to screen bounds, and alpha-blended over a neutral background.
This produces two tensors at video frame rate: a cursor-only foreground image $f_t \in [-1,1]^{3\times H \times W}$ and a soft mask $m_t \in [0,1]^{1\times H \times W}$ isolating the arrow pixels.
Invalid or missing coordinates zero the mask and leave the foreground unchanged, so frames without visible cursors do not add spurious supervision.

\noindent\textbf{Reference images and masks.}
These cursor layers become reference conditions for the image-to-video (I2V) model.
For each clip we form reference images $\mathrm{ref\_img}_{0:T-1}$ and masks $\mathrm{ref\_mask}_{0:T-1}$:
\begin{itemize}
  \item At $t{=}0$, $\mathrm{ref\_img}_0$ is the full desktop frame and $\mathrm{ref\_mask}_0$ is all ones, anchoring the static layout and theme;
  \item For $t{>}0$, $\mathrm{ref\_img}_t$ is the cursor foreground $f_t$ and $\mathrm{ref\_mask}_t$ is the cursor mask $m_t$, so only the arrow region is supervised while the background remains free.
\end{itemize}
The model encodes these references with the same VAE as target frames and concatenates their latents and masks into the diffusion input.
It enforces masked, reference-consistent reconstruction inside the cursor region while relying on learned dynamics elsewhere.
This makes cursor supervision a pixel-level constraint rather than a side effect of the global loss.

\noindent\textbf{Fourier mouse encoding.}
The same $(x_t,y_t)$ trajectories serve as a continuous control signal.
We apply a Fourier position module: clamp coordinates to $[0,1]^2$, map them to $[-1,1]^2$, and compute random Fourier features via a fixed Gaussian projection followed by sine/cosine.
A small MLP maps these features to per-frame mouse embeddings.
The GUIWorld action encoder then aggregates them with lag-aware, stride-aligned windows to produce latent-aligned mouse features.
These features condition the \texttt{external}/\texttt{contextual}/\texttt{residual}/\texttt{internal} modes and participate in the temporal contrastive loss.

\noindent\textbf{Cursor-aware losses.}
Section~\ref{section:impl-guiworld} introduces cursor-aware losses that use this construction.
A basic variant penalizes position error only in $(x,y)$.
Richer variants add Fourier features of the trajectory and, most importantly, an $\ell_2$ loss on the reconstructed cursor patch under $\mathrm{ref\_mask}_t$.
Table~\ref{tab:cursor-loss} shows that position-only objectives yield low cursor hit rate and visibly jittery arrows even when videos look plausible.
Adding the explicit cursor reference stream together with the masked patch loss substantially improves control, reaching 98.7\% cursor accuracy.
This confirms that explicit cursor rendering plus localized supervision effectively separates ``where the arrow is'' from ``what the rest of the frame should look like''.

\noindent\textbf{Temporal contrastive alignment.}
To strengthen learning signals for the action pathway, we add a lightweight temporal contrastive loss that operates on the same latent timeline as the diffusion model.
For each sequence we take per-step frame features $F_{t} \in \mathbb{R}^{d_f}$ pooled from the latent video.
We also take per-step action features $A_{t} \in \mathbb{R}^{d_a}$ and (optionally) mouse features $M_{t} \in \mathbb{R}^{d_m}$ produced by the action encoder.
Linear projections map these into a common space and the resulting vectors are $\ell_2$-normalized.
An InfoNCE-style objective brings matching pairs $(F_t,A_t)$ (and, when present, $(F_t,M_t)$) from the same timestep together.
In implementation, matching is lag-aware: frame $t$ is aligned with action/mouse features at $t-\ell$, where $\ell$ is the configured \texttt{action\_lag}.
Similarities are scaled by a temperature $\tau$.
It pushes matched pairs away from other timesteps of the \emph{same} sequence.
We use frame and action masks to ignore positions without actions.
A symmetric variant averages frame-to-action and action-to-frame directions.

When enabled, a small future-prediction head adds a second term.
Action features at time $t$ are mapped to a prediction of the frame feature at a slightly later step $t+\ell$.
A mean-squared error encourages consistency between the prediction and the projected future frame.
Together, these terms give the action encoder direct gradients tied to specific frames rather than relying solely on the pixel diffusion loss.
The contrastive term enforces tight temporal alignment between actions and the frames they co-occur with.
The future head encourages actions to anticipate the visual consequences that appear shortly after they are issued.

\noindent\textbf{Encoder comparison.}
Table~\ref{tab:action-encoder-comparison} contrasts the two encoders used in GUIWorld.
The key theme is moving from a high-dimensional, bursty event vector toward a typed, API-like schema with explicit lag handling.
This makes the action stream easier to align with the latent video timeline and to reuse across injection modes.

\noindent\textbf{Injection schemes.}
The GUI experiments explore four conditioning schemes built on this encoder: \texttt{external}, \texttt{contextual}, \texttt{internal}, and \texttt{residual} (Figure~\ref{fig:action-injection}).
Representative action-driven metrics comparing these modes are reported in Table~\ref{tab:gui-action-ssim}.
Full metric definitions and the evaluation protocol are provided in Appendix~\ref{appendix:gui-metrics}.

%% file: section/appendix_vis.tex
\subsection{CLIGen Visualizations}
\noindent This subsection consolidates \emph{all} CLIGen visualization pages referenced in the paper.
The main text keeps section-local thumbnail panels at the end of each visualization subsection, while full-size pages are collected here.

\noindent\textbf{(1) \cligenGeneralData{} visualizations.}
Qualitative samples highlight the breadth of real-world terminal dynamics captured in CLIGen (General): ANSI escape sequences that repaint regions with changing foreground/background colors, incremental command entry with syntax highlighting and cursor edits, classic shell prompts and system outputs, long-running jobs with rapidly scrolling and color-coded package logs, full-screen TUIs, and progress dashboards.

\noindent\textbf{(2) \cligenCleanData{} REPL visualizations.}
In contrast to open-world traces, CLIGen (Clean) REPL samples are scripted and temporally well-paced (Figures~\ref{fig:cligen-clean-viz-a}--\ref{fig:cligen-clean-viz-d}; additional format examples are in Appendix~\ref{appendix:cligen-samples}). Each sample includes an explicit action trace (e.g., \texttt{Sleep}, \texttt{Type}, \texttt{Enter}, arrow keys, \texttt{Hide}) alongside rendered terminal frames, making action-to-pixel causality easy to inspect.

\noindent\textbf{(3) \cligenCleanData{} math visualizations.}
Figures~\ref{fig:cligen-clean-math-comp-a}--\ref{fig:cligen-clean-math-comp-c} compare model rollouts on CLIGen (Clean) math REPL prompts. Figures~\ref{fig:cligen-clean-math-rep-a}--\ref{fig:cligen-clean-math-rep-c} show reprompting cases and highlight why these probes should separate native computation from answer-conditioned rendering.

\subsection{GUIWorld Visualizations}
\noindent This subsection consolidates \emph{all} GUIWorld rollout visualizations (Figures~\ref{fig:guiworld-sample-1}--\ref{fig:guiworld-sample-14}).
CUA-based pages overlay the CUA trace (natural-language rationale when available, e.g., in a \texttt{thinking} field, plus structured \emph{action} fields such as \texttt{left\_click}, \texttt{double\_click}, \texttt{left\_click\_drag}, and \texttt{type}).
It contrasts the \emph{Ground Truth} trajectory (top) with a \emph{Generation} conditioned on the first frame and the action sequence (bottom), making state drift easy to spot.

Figures~\ref{fig:guiworld-sample-6}--\ref{fig:guiworld-sample-8} emphasize compounding low-level deviations; Figures~\ref{fig:guiworld-sample-9}--\ref{fig:guiworld-sample-11} focus on numeric/UI fidelity and interaction semantics; and Figures~\ref{fig:guiworld-sample-12}--\ref{fig:guiworld-sample-14} provide additional stress cases where correctness hinges on precise field edits and page state.

\newcommand{\BackToMainThumbs}[1]{%
  \hyperlink{#1}{%
    {\setlength{\fboxsep}{8.5pt}%
    \fcolorbox{metablue!55}{white}{\Large\bfseries\textsf{\strut Back to Main Thumbnails}}}%
  }%
}

\newcommand{\VisPage}[4]{%
  \clearpage
  \newgeometry{top=1.2cm,bottom=1.6cm,left=1.2cm,right=1.2cm,includefoot,footskip=22pt}
  \begin{landscape}
  \thispagestyle{plain}
  \noindent\makebox[\linewidth][r]{\BackToMainThumbs{#4}}\par
  \centering
  \vspace*{\fill}
  \includegraphics[width=0.98\linewidth,height=0.84\textheight,keepaspectratio]{#1}\par
  {\captionsetup{type=figure,skip=2pt}\captionof{figure}{#2}\label{#3}}
  \vspace*{\fill}
  \end{landscape}
  \restoregeometry
}

\VisPage{assets/cligen_general_1.pdf}{\cligenGeneralData{} visualization samples (A).}{fig:cligen-general-viz-1}{main-cligen-vis-thumbs-p1}
\VisPage{assets/cligen_general_2.pdf}{\cligenGeneralData{} visualization samples (B).}{fig:cligen-general-viz-2}{main-cligen-vis-thumbs-p1}
\VisPage{assets/cligen_general_3.pdf}{\cligenGeneralData{} visualization samples (C).}{fig:cligen-general-viz-3}{main-cligen-vis-thumbs-p1}

\VisPage{assets/cligen_clean_1.pdf}{\cligenCleanData{} REPL visualization samples (A).}{fig:cligen-clean-viz-a}{main-cligen-vis-thumbs-p2}
\VisPage{assets/cligen_clean_2.pdf}{\cligenCleanData{} REPL visualization samples (B).}{fig:cligen-clean-viz-b}{main-cligen-vis-thumbs-p2}
\VisPage{assets/cligen_clean_3.pdf}{\cligenCleanData{} REPL visualization samples (C).}{fig:cligen-clean-viz-c}{main-cligen-vis-thumbs-p2}
\VisPage{assets/cligen_clean_4.pdf}{\cligenCleanData{} REPL visualization samples (D).}{fig:cligen-clean-viz-d}{main-cligen-vis-thumbs-p2}

\VisPage{assets/cligen_math_comp1.pdf}{\cligenCleanData{} math comparison samples (A).}{fig:cligen-clean-math-comp-a}{main-cligen-vis-thumbs-p3}
\VisPage{assets/cligen_math_comp2.pdf}{\cligenCleanData{} math comparison samples (B).}{fig:cligen-clean-math-comp-b}{main-cligen-vis-thumbs-p3}
\VisPage{assets/cligen_math_comp3.pdf}{\cligenCleanData{} math comparison samples (C).}{fig:cligen-clean-math-comp-c}{main-cligen-vis-thumbs-p3}

\VisPage{assets/cligen_math_rep1.pdf}{\cligenCleanData{} math reprompting samples (A).}{fig:cligen-clean-math-rep-a}{main-cligen-vis-thumbs-p4}
\VisPage{assets/cligen_math_rep2.pdf}{\cligenCleanData{} math reprompting samples (B).}{fig:cligen-clean-math-rep-b}{main-cligen-vis-thumbs-p4}
\VisPage{assets/cligen_math_rep3.pdf}{\cligenCleanData{} math reprompting samples (C).}{fig:cligen-clean-math-rep-c}{main-cligen-vis-thumbs-p4}

\VisPage{assets/guiworld_sample_1.pdf}{\guiworldLogo{}\,GUIWorld visualization sample (1).}{fig:guiworld-sample-1}{main-gui-vis-thumbs-p1}
\VisPage{assets/guiworld_sample_2.pdf}{\guiworldLogo{}\,GUIWorld visualization sample (2).}{fig:guiworld-sample-2}{main-gui-vis-thumbs-p1}
\VisPage{assets/guiworld_sample_3.pdf}{\guiworldLogo{}\,GUIWorld visualization sample (3).}{fig:guiworld-sample-3}{main-gui-vis-thumbs-p1}
\VisPage{assets/guiworld_sample_4.pdf}{\guiworldLogo{}\,GUIWorld visualization sample (4).}{fig:guiworld-sample-4}{main-gui-vis-thumbs-p1}
\VisPage{assets/guiworld_sample_5.pdf}{\guiworldLogo{}\,GUIWorld visualization sample (5).}{fig:guiworld-sample-5}{main-gui-vis-thumbs-p1}
\VisPage{assets/guiworld_sample_6.pdf}{\guiworldLogo{}\,GUIWorld visualization sample (6).}{fig:guiworld-sample-6}{main-gui-vis-thumbs-p2}
\VisPage{assets/guiworld_sample_7.pdf}{\guiworldLogo{}\,GUIWorld visualization sample (7).}{fig:guiworld-sample-7}{main-gui-vis-thumbs-p2}
\VisPage{assets/guiworld_sample_8.pdf}{\guiworldLogo{}\,GUIWorld visualization sample (8).}{fig:guiworld-sample-8}{main-gui-vis-thumbs-p2}
\VisPage{assets/guiworld_sample_9.pdf}{\guiworldLogo{}\,GUIWorld visualization sample (9).}{fig:guiworld-sample-9}{main-gui-vis-thumbs-p2}
\VisPage{assets/guiworld_sample_10.pdf}{\guiworldLogo{}\,GUIWorld visualization sample (10).}{fig:guiworld-sample-10}{main-gui-vis-thumbs-p2}
\VisPage{assets/guiworld_sample_11.pdf}{\guiworldLogo{}\,GUIWorld visualization sample (11).}{fig:guiworld-sample-11}{main-gui-vis-thumbs-p3}
\VisPage{assets/guiworld_sample_12.pdf}{\guiworldLogo{}\,GUIWorld visualization sample (12).}{fig:guiworld-sample-12}{main-gui-vis-thumbs-p3}
\VisPage{assets/guiworld_sample_13.pdf}{\guiworldLogo{}\,GUIWorld visualization sample (13).}{fig:guiworld-sample-13}{main-gui-vis-thumbs-p4}
\VisPage{assets/guiworld_sample_14.pdf}{\guiworldLogo{}\,GUIWorld visualization sample (14).}{fig:guiworld-sample-14}{main-gui-vis-thumbs-p5}

%% file: paper.bib
@inproceedings{zhuge2024gptswarm_icml,
  title={Gptswarm: Language agents as optimizable graphs},
  author={Zhuge, Mingchen and Wang, Wenyi and Kirsch, Louis and Faccio, Francesco and Khizbullin, Dmitrii and Schmidhuber, J{\"u}rgen},
  booktitle={Forty-first International Conference on Machine Learning},
  year={2024}
}

@article{fritzke1994growing,
  title={A growing neural gas network learns topologies},
  author={Fritzke, Bernd},
  journal={Advances in neural information processing systems},
  volume={7},
  year={1994}
}

@article{zhuge2024agent_judge,
  title={Agent-as-a-judge: Evaluate agents with agents},
  author={Zhuge, Mingchen and Zhao, Changsheng and Ashley, Dylan and Wang, Wenyi and Khizbullin, Dmitrii and Xiong, Yunyang and Liu, Zechun and Chang, Ernie and Krishnamoorthi, Raghuraman and Tian, Yuandong and others},
  journal={arXiv preprint arXiv:2410.10934},
  year={2024}
}

@inproceedings{zhuge2026ai_recursive,
  title={AI with Recursive Self-Improvement},
  author={Zhuge, Mingchen and Zeng, Ailing and Zhu, Deyao and Yang, Sherry and Chandra, Vikas and Schmidhuber, J{\"u}rgen},
  booktitle={ICLR 2026 Workshop Proposals},
  year={2026}
}

@misc{claudeComputerTool,
		author = {Anthropic},
		title = {{C}omputer use tool --- platform.claude.com},
	howpublished = {\url{https://platform.claude.com/docs/en/agents-and-tools/tool-use/computer-use-tool}},
	year = {},
	note = {[Accessed 02-02-2026]},
}

@inproceedings{werbos1987learning,
  title={Learning how the world works: Specifications for predictive networks in robots and brains},
  author={Werbos, Paul J},
  booktitle={Proceedings of IEEE International Conference on Systems, Man and Cybernetics, NY},
  year={1987}
}

@inproceedings{munro1987dual,
  title={A dual back-propagation scheme for scalar reward learning},
  author={Munro, Paul},
  booktitle={Ninth Annual Conference of the Cognitive Science Society},
  pages={165--176},
  year={1987},
  organization={Hillsdale, NJ. Cognitive Science Society Lawrence Erlbaum}
}

@incollection{nguyen1990truck,
  title={The truck backer-upper: An example of self-learning in neural networks},
  author={Nguyen, Derrick and Widrow, Bernard},
  booktitle={Advanced neural computers},
  pages={11--19},
  year={1990},
  publisher={Elsevier}
}

@inproceedings{hartmanis1974power,
  title={On the power of multiplication in random access machines},
  author={Hartmanis, Juris and Simon, Janos},
  booktitle={15th Annual Symposium on Switching and Automata Theory (swat 1974)},
  pages={13--23},
  year={1974},
  organization={IEEE}
}

@inproceedings{backus1957fortran,
  title={The FORTRAN automatic coding system},
  author={Backus, John W and Beeber, Robert J and Best, Sheldon and Goldberg, Richard and Haibt, Lois M and Herrick, Harlan L and Nelson, Robert A and Sayre, David and Sheridan, Peter B and Stern, Harold and others},
  booktitle={Papers presented at the February 26-28, 1957, western joint computer conference: Techniques for reliability},
  pages={188--198},
  year={1957}
}

@article{backus1963revised,
  title={Revised report on the algorithmic language Algol 60},
  author={Backus, John W and Bauer, Friedrich L and Green, Julien and Katz, Charles and McCarthy, John and Perlis, Alan J and Rutishauser, Heinz and Samelson, Klaus and Vauquois, Bernard and Wegstein, Joseph Henry and others},
  journal={Communications of the ACM},
  volume={6},
  number={1},
  pages={1--17},
  year={1963},
  publisher={ACM New York, NY, USA}
}

@inproceedings{anagnostopoulos1973computer,
  title={Computer architecture and instruction set design},
  author={Anagnostopoulos, Paul Constantine and Michel, MJ and Sockut, Gary H and Stabler, George M and van Dam, Andries},
  booktitle={Proceedings of the June 4-8, 1973, national computer conference and exposition},
  pages={519--527},
  year={1973}
}

@article{raffel2020exploring,
  title={Exploring the limits of transfer learning with a unified text-to-text transformer},
  author={Raffel, Colin and Shazeer, Noam and Roberts, Adam and Lee, Katherine and Narang, Sharan and Matena, Michael and Zhou, Yanqi and Li, Wei and Liu, Peter J},
  journal={Journal of machine learning research},
  volume={21},
  number={140},
  pages={1--67},
  year={2020}
}

@article{ha2018world,
  title={World models},
  author={Ha, David and Schmidhuber, J{\"u}rgen},
  journal={arXiv preprint arXiv:1803.10122},
  year={2018}
}

@article{dubey2024llama,
  title={The llama 3 herd of models},
  author={Dubey, Abhimanyu and Jauhri, Abhinav and Pandey, Abhinav and Kadian, Abhishek and Al-Dahle, Ahmad and Letman, Aiesha and Mathur, Akhil and Schelten, Alan and Yang, Amy and Fan, Angela and others},
  journal={arXiv preprint arXiv:2407.21783},
  year={2024}
}

@misc{openai2024sora,
  author       = {OpenAI},
  title        = {Sora by OpenAI},
  year         = {2024},
  howpublished = {\url{https://openai.com/sora/}},
  note         = {Accessed: 2025-07-14}
}

@online{deepmind_veo_2025,
  title        = {Veo},
  author       = {{Google DeepMind}},
  year         = {2025},
  month        = may,
  howpublished = {\url{https://deepmind.google/models/veo/}},
  urldate      = {2025-07-14}
}

@article{polyak2024movie,
  title={Movie gen: A cast of media foundation models},
  author={Polyak, Adam and Zohar, Amit and Brown, Andrew and Tjandra, Andros and Sinha, Animesh and Lee, Ann and Vyas, Apoorv and Shi, Bowen and Ma, Chih-Yao and Chuang, Ching-Yao and others},
  journal={arXiv preprint arXiv:2410.13720},
  year={2024}
}

@article{rivard2025neuralos,
  title={NeuralOS: Towards Simulating Operating Systems via Neural Generative Models},
  author={Rivard, Luke and Sun, Sun and Guo, Hongyu and Chen, Wenhu and Deng, Yuntian},
  journal={arXiv preprint arXiv:2507.08800},
  year={2025}
}

@article{hochreiter1997long,
  title={Long short-term memory},
  author={Hochreiter, Sepp and Schmidhuber, J{\"u}rgen},
  journal={Neural computation},
  volume={9},
  number={8},
  pages={1735--1780},
  year={1997},
  publisher={MIT press}
}

@article{he2025matrix,
  title={{Matrix-game 2.0}: An open-source, real-time, and streaming interactive world model},
  author={He, Xianglong and Peng, Chunli and Liu, Zexiang and Wang, Boyang and Zhang, Yifan and Cui, Qi and Kang, Fei and Jiang, Biao and An, Mengyin and Ren, Yangyang and others},
  journal={arXiv preprint arXiv:2508.13009},
  year={2025}
}

@misc{anthropic_claude_sonnet4_5_2025,
  author       = {Anthropic},
  title        = {Introducing Claude Sonnet 4.5},
  year         = {2025},
  month        = sep,
  day          = {29},
  howpublished = {\url{https://www.anthropic.com/news/claude-sonnet-4-5}},
}

@inproceedings{hong2023metagpt,
  title={MetaGPT: Meta programming for a multi-agent collaborative framework},
  author={Hong, Sirui and Zhuge, Mingchen and Chen, Jonathan and Zheng, Xiawu and Cheng, Yuheng and Wang, Jinlin and Zhang, Ceyao and Wang, Zili and Yau, Steven Ka Shing and Lin, Zijuan and others},
  booktitle={The twelfth international conference on learning representations},
  year={2023}
}

@inproceedings{zhuge2026ai,
  title={AI with Recursive Self-Improvement},
  author={Zhuge, Mingchen and Zeng, Ailing and Zhu, Deyao and Yang, Sherry and Chandra, Vikas and Schmidhuber, J{\"u}rgen},
  booktitle={ICLR 2026 Workshop Proposals},
  year={2026}
}

@misc{google_veo3_1_2025,
  author       = {Google},
  title        = {Break the Silence with Veo 3.1},
  year         = {2025},
  month        = nov,
  howpublished = {\url{https://gemini.google/overview/video-generation/}},
}

@inproceedings{bruce2024genie,
  title={Genie: Generative interactive environments},
  author={Bruce, Jake and Dennis, Michael D and Edwards, Ashley and Parker-Holder, Jack and Shi, Yuge and Hughes, Edward and Lai, Matthew and Mavalankar, Aditi and Steigerwald, Richie and Apps, Chris and others},
  booktitle={Forty-first International Conference on Machine Learning},
  year={2024}
}

@misc{openai_sora2_2025,
  author       = {OpenAI},
  title        = {Sora 2 is here},
  year         = {2025},
  month        = sep,
  day          = {30},
  howpublished = {\url{https://openai.com/index/sora-2/}},
}

@article{graves2014neural,
  title={Neural Turing Machines},
  author={Graves, Alex and Wayne, Greg and Danihelka, Ivo},
  journal={arXiv preprint arXiv:1410.5401},
  year={2014}
}

@article{graves2016hybrid,
  title={Hybrid computing using a neural network with dynamic external memory},
  author={Graves, Alex and Wayne, Greg and Reynolds, Malcolm and Harley, Tim and Danihelka, Ivo and Grabska-Barwi{\'n}ska, Agnieszka and G{\'o}mez Colmenarejo, Sergio and Grefenstette, Edward and Ramalho, Tiago and Agapiou, John and others},
  journal={Nature},
  volume={538},
  number={7626},
  pages={471--476},
  year={2016}
}

@article{wan2025wan,
  title={Wan: Open and advanced large-scale video generative models},
  author={Wan, Team and Wang, Ang and Ai, Baole and Wen, Bin and Mao, Chaojie and Xie, Chen-Wei and Chen, Di and Yu, Feiwu and Zhao, Haiming and Yang, Jianxiao and others},
  journal={arXiv preprint arXiv:2503.20314},
  year={2025}
}

@book{mead2012analog,
  title={Analog VLSI implementation of neural systems},
  author={Mead, Carver and Ismail, Mohammed},
  volume={80},
  year={2012},
  publisher={Springer Science \& Business Media}
}

@article{reed2015neural,
  title={Neural programmer-interpreters},
  author={Reed, Scott and De Freitas, Nando},
  journal={arXiv preprint arXiv:1511.06279},
  year={2015}
}

@inproceedings{hafner2019learning,
  title={Learning latent dynamics for planning from pixels},
  author={Hafner, Danijar and Lillicrap, Timothy and Fischer, Ian and Villegas, Ruben and Ha, David and Lee, Honglak and Davidson, James},
  booktitle={International conference on machine learning},
  pages={2555--2565},
  year={2019},
  organization={PMLR}
}

@article{hafner2019dream,
  title={Dream to control: Learning behaviors by latent imagination},
  author={Hafner, Danijar and Lillicrap, Timothy and Ba, Jimmy and Norouzi, Mohammad},
  journal={arXiv preprint arXiv:1912.01603},
  year={2019}
}

@article{katz2019programmable,
  title={A programmable neural virtual machine based on a fast store-erase learning rule},
  author={Katz, Garrett E and Davis, Gregory P and Gentili, Rodolphe J and Reggia, James A},
  journal={Neural Networks},
  volume={119},
  pages={10--30},
  year={2019},
  publisher={Elsevier}
}

@article{Smolensky1988-SMOOTP-2,
	author = {Paul Smolensky},
	doi = {10.1017/s0140525x00052432},
	journal = {Behavioral and Brain Sciences},
	number = {1},
	pages = {1--23},
	title = {On the Proper Treatment of Connectionism},
	volume = {11},
	year = {1988}
}

@inproceedings{Reynolds2021,
author = {Reynolds, Laria and McDonell, Kyle},
title = {Prompt Programming for Large Language Models: Beyond the Few-Shot Paradigm},
year = {2021},
isbn = {9781450380959},
publisher = {Association for Computing Machinery},
address = {New York, NY, USA},
url = {https://doi.org/10.1145/3411763.3451760},
doi = {10.1145/3411763.3451760},
booktitle = {Extended Abstracts of the 2021 CHI Conference on Human Factors in Computing Systems},
articleno = {314},
numpages = {7},
location = {Yokohama, Japan},
series = {CHI EA '21}
}

@misc{wei2023chainofthoughtpromptingelicitsreasoning,
      title={Chain-of-Thought Prompting Elicits Reasoning in Large Language Models}, 
      author={Jason Wei and Xuezhi Wang and Dale Schuurmans and Maarten Bosma and Brian Ichter and Fei Xia and Ed Chi and Quoc Le and Denny Zhou},
      year={2023},
      eprint={2201.11903},
      archivePrefix={arXiv},
      primaryClass={cs.CL},
      url={https://arxiv.org/abs/2201.11903}, 
}

@Book{jm3,
  author =       "Daniel Jurafsky and James H. Martin",
  title =        "Speech and Language Processing: An Introduction to Natural Language Processing, 
  		  Computational Linguistics, and Speech Recognition,
		   with Language Models",
  year =         "2026",
  url = {https://web.stanford.edu/~jurafsky/slp3/},
  note = "Online manuscript released January 6, 2026",
  edition =         "3rd",
  }

@book{Bishop2006,
  archiveprefix = {arXiv},
  arxivid = {0-387-31073-8},
  author = {Bishop, Christopher M},
  booktitle = {Pattern Recognition},
  chapter = {Graphical},
  doi = {10.1117/1.2819119},
  editor = {Jordan, M and Kleinberg, J and Sch\"{o}lkopf, B},
  eprint = {0-387-31073-8},
  isbn = {9780387310732},
  issn = {10179909},
  number = 4,
  pages = 738,
  pmid = {8943268},
  publisher = {Springer},
  series = {Information science and statistics},
  title = {{Pattern Recognition and Machine Learning}},
  url = {http://www.library.wisc.edu/selectedtocs/bg0137.pdf},
  volume = 4,
  year = 2006
}

@book{tanenbaum2013structured,
  title={Structured Computer Organization},
  author={Tanenbaum, A.S. and Austin, T.},
  isbn={9780132916523},
  lccn={2012021627},
  url={https://books.google.com.sa/books?id=m0HHygAACAAJ},
  year={2013},
  publisher={Pearson}
}

@article{vasilache2018tensor,
  title={Tensor comprehensions: Framework-agnostic high-performance machine learning abstractions},
  author={Vasilache, Nicolas and Zinenko, Oleksandr and Theodoridis, Theodoros and Goyal, Priya and DeVito, Zachary and Moses, William S and Verdoolaege, Sven and Adams, Andrew and Cohen, Albert},
  journal={arXiv preprint arXiv:1802.04730},
  year={2018}
}

@book{gregg2014systems,
  title={Systems performance: enterprise and the cloud},
  author={Gregg, Brendan},
  year={2014},
  publisher={Pearson Education}
}

@book{silberschatz2019operating,
  title={Operating system concepts},
  author={Silberschatz, Abraham and Galvin, Peter B and Gagne, Greg},
  year={2019},
  publisher={John Wiley \& Sons}
}

@article{myers2002demonstrational,
  title={Demonstrational interfaces: A step beyond direct manipulation},
  author={Myers, Brad A},
  journal={Computer},
  volume={25},
  number={8},
  pages={61--73},
  year={2002},
  publisher={IEEE}
}

@inbook{Hinton1986,
author = {Hinton, G. E. and McClelland, J. L. and Rumelhart, D. E.},
title = {Distributed representations},
year = {1986},
isbn = {026268053X},
publisher = {MIT Press},
address = {Cambridge, MA, USA},
booktitle = {Parallel Distributed Processing: Explorations in the Microstructure of Cognition, Vol. 1: Foundations},
pages = {77–109},
numpages = {33}
}

@article{Newell1976,
author = {Newell, Allen and Simon, Herbert A.},
title = {Computer science as empirical inquiry: symbols and search},
year = {1976},
issue_date = {March 1976},
publisher = {Association for Computing Machinery},
address = {New York, NY, USA},
volume = {19},
number = {3},
issn = {0001-0782},
url = {https://doi.org/10.1145/360018.360022},
doi = {10.1145/360018.360022},
journal = {Commun. ACM},
month = mar,
pages = {113–126},
numpages = {14}
}

@article{BROOKS1991139,
title = {Intelligence without representation},
journal = {Artificial Intelligence},
volume = {47},
number = {1},
pages = {139-159},
year = {1991},
issn = {0004-3702},
doi = {https://doi.org/10.1016/0004-3702(91)90053-M},
url = {https://www.sciencedirect.com/science/article/pii/000437029190053M},
author = {Rodney A. Brooks}
}

@misc{sager2025comprehensivesurveyagentscomputer,
      title={A Comprehensive Survey of Agents for Computer Use: Foundations, Challenges, and Future Directions}, 
      author={Pascal J. Sager and Benjamin Meyer and Peng Yan and Rebekka von Wartburg-Kottler and Layan Etaiwi and Aref Enayati and Gabriel Nobel and Ahmed Abdulkadir and Benjamin F. Grewe and Thilo Stadelmann},
      year={2025},
      eprint={2501.16150},
      archivePrefix={arXiv},
      primaryClass={cs.AI},
      url={https://arxiv.org/abs/2501.16150}, 
}

@article{feng2023finetuning,
  title={Finetuning offline world models in the real world},
  author={Feng, Yunhai and Hansen, Nicklas and Xiong, Ziyan and Rajagopalan, Chandramouli and Wang, Xiaolong},
  journal={arXiv preprint arXiv:2310.16029},
  year={2023}
}

@misc{openaiComputerUsingAgent,
	author = {OpenAI},
	title = {{C}omputer-{U}sing {A}gent --- openai.com},
	howpublished = {\url{https://openai.com/index/computer-using-agent/}},
	year = {2023},
	note = {[Accessed 07-02-2026]},
}

@inproceedings{garcia2023robust,
  title={Robust visual sim-to-real transfer for robotic manipulation},
  author={Garcia, Ricardo and Strudel, Robin and Chen, Shizhe and Arlaud, Etienne and Laptev, Ivan and Schmid, Cordelia},
  booktitle={2023 IEEE/RSJ International Conference on Intelligent Robots and Systems (IROS)},
  pages={992--999},
  year={2023},
  organization={ieee}
}

@inproceedings{dosovitskiy2017carla,
  title={CARLA: An open urban driving simulator},
  author={Dosovitskiy, Alexey and Ros, German and Codevilla, Felipe and Lopez, Antonio and Koltun, Vladlen},
  booktitle={Conference on robot learning},
  pages={1--16},
  year={2017},
  organization={PMLR}
}

@inproceedings{richter2016playing,
  title={Playing for data: Ground truth from computer games},
  author={Richter, Stephan R and Vineet, Vibhav and Roth, Stefan and Koltun, Vladlen},
  booktitle={European conference on computer vision},
  pages={102--118},
  year={2016},
  organization={Springer}
}

@inproceedings{tobin2017domain,
  title={Domain randomization for transferring deep neural networks from simulation to the real world},
  author={Tobin, Josh and Fong, Rachel and Ray, Alex and Schneider, Jonas and Zaremba, Wojciech and Abbeel, Pieter},
  booktitle={2017 IEEE/RSJ international conference on intelligent robots and systems (IROS)},
  pages={23--30},
  year={2017},
  organization={IEEE}
}

@inproceedings{yao2022react,
  title={React: Synergizing reasoning and acting in language models},
  author={Yao, Shunyu and Zhao, Jeffrey and Yu, Dian and Du, Nan and Shafran, Izhak and Narasimhan, Karthik R and Cao, Yuan},
  booktitle={The eleventh international conference on learning representations},
  year={2022}
}

@article{yao2022webshop,
  title={Webshop: Towards scalable real-world web interaction with grounded language agents},
  author={Yao, Shunyu and Chen, Howard and Yang, John and Narasimhan, Karthik},
  journal={Advances in Neural Information Processing Systems},
  volume={35},
  pages={20744--20757},
  year={2022}
}

@book{cypher1993watch,
  title={Watch what I do: programming by demonstration},
  author={Cypher, Allen and Halbert, Daniel Conrad},
  year={1993},
  publisher={MIT press}
}

@article{flener2008introduction,
  title={An introduction to inductive programming},
  author={Flener, Pierre and Schmid, Ute},
  journal={Artificial Intelligence Review},
  volume={29},
  number={1},
  pages={45--62},
  year={2008},
  publisher={Springer}
}

@article{austin2021program,
  title={Program synthesis with large language models},
  author={Austin, Jacob and Odena, Augustus and Nye, Maxwell and Bosma, Maarten and Michalewski, Henryk and Dohan, David and Jiang, Ellen and Cai, Carrie and Terry, Michael and Le, Quoc and others},
  journal={arXiv preprint arXiv:2108.07732},
  year={2021}
}

@article{ouyang2022training,
  title={Training language models to follow instructions with human feedback},
  author={Ouyang, Long and Wu, Jeffrey and Jiang, Xu and Almeida, Diogo and Wainwright, Carroll and Mishkin, Pamela and Zhang, Chong and Agarwal, Sandhini and Slama, Katarina and Ray, Alex and others},
  journal={Advances in neural information processing systems},
  volume={35},
  pages={27730--27744},
  year={2022}
}

@article{radford2019language,
  title={Language models are unsupervised multitask learners},
  author={Radford, Alec and Wu, Jeffrey and Child, Rewon and Luan, David and Amodei, Dario and Sutskever, Ilya and others},
  journal={OpenAI blog},
  volume={1},
  number={8},
  pages={9},
  year={2019}
}

@article{bommasani2021opportunities,
  title={On the opportunities and risks of foundation models},
  author={Bommasani, Rishi},
  journal={arXiv preprint arXiv:2108.07258},
  year={2021}
}

@article{wei2022chain,
  title={Chain-of-thought prompting elicits reasoning in large language models},
  author={Wei, Jason and Wang, Xuezhi and Schuurmans, Dale and Bosma, Maarten and Xia, Fei and Chi, Ed and Le, Quoc V and Zhou, Denny and others},
  journal={Advances in neural information processing systems},
  volume={35},
  pages={24824--24837},
  year={2022}
}

@article{brown2020language,
  title={Language models are few-shot learners},
  author={Brown, Tom and Mann, Benjamin and Ryder, Nick and Subbiah, Melanie and Kaplan, Jared D and Dhariwal, Prafulla and Neelakantan, Arvind and Shyam, Pranav and Sastry, Girish and Askell, Amanda and others},
  journal={Advances in neural information processing systems},
  volume={33},
  pages={1877--1901},
  year={2020}
}

@article{ramachandram2017deep,
  title={Deep multimodal learning: A survey on recent advances and trends},
  author={Ramachandram, Dhanesh and Taylor, Graham W},
  journal={IEEE signal processing magazine},
  volume={34},
  number={6},
  pages={96--108},
  year={2017},
  publisher={IEEE}
}

@article{pierrot2019learning,
  title={Learning compositional neural programs with recursive tree search and planning},
  author={Pierrot, Thomas and Ligner, Guillaume and Reed, Scott E and Sigaud, Olivier and Perrin, Nicolas and Laterre, Alexandre and Kas, David and Beguir, Karim and de Freitas, Nando},
  journal={Advances in Neural Information Processing Systems},
  volume={32},
  year={2019}
}

@article{sze2017efficient,
  title={Efficient processing of deep neural networks: A tutorial and survey},
  author={Sze, Vivienne and Chen, Yu-Hsin and Yang, Tien-Ju and Emer, Joel S},
  journal={Proceedings of the IEEE},
  volume={105},
  number={12},
  pages={2295--2329},
  year={2017},
  publisher={Ieee}
}

@article{paszke2019pytorch,
  title={Pytorch: An imperative style, high-performance deep learning library},
  author={Paszke, Adam and Gross, Sam and Massa, Francisco and Lerer, Adam and Bradbury, James and Chanan, Gregory and Killeen, Trevor and Lin, Zeming and Gimelshein, Natalia and Antiga, Luca and others},
  journal={Advances in neural information processing systems},
  volume={32},
  year={2019}
}

@article{bengio2013representation,
  title={Representation learning: A review and new perspectives},
  author={Bengio, Yoshua and Courville, Aaron and Vincent, Pascal},
  journal={IEEE transactions on pattern analysis and machine intelligence},
  volume={35},
  number={8},
  pages={1798--1828},
  year={2013},
  publisher={IEEE}
}

@article{kingma2013auto,
  title={Auto-encoding variational bayes},
  author={Kingma, Diederik P and Welling, Max},
  journal={arXiv preprint arXiv:1312.6114},
  year={2013}
}

@article{marcus2018deep,
  title={Deep learning: A critical appraisal},
  author={Marcus, Gary},
  journal={arXiv preprint arXiv:1801.00631},
  year={2018}
}

@article{silver2017mastering,
  title={Mastering the game of go without human knowledge},
  author={Silver, David and Schrittwieser, Julian and Simonyan, Karen and Antonoglou, Ioannis and Huang, Aja and Guez, Arthur and Hubert, Thomas and Baker, Lucas and Lai, Matthew and Bolton, Adrian and others},
  journal={nature},
  volume={550},
  number={7676},
  pages={354--359},
  year={2017},
  publisher={Nature Publishing Group UK London}
}

@article{clements2019estimating,
  title={Estimating risk and uncertainty in deep reinforcement learning},
  author={Clements, William R and Van Delft, Bastien and Robaglia, Beno{\^\i}t-Marie and Slaoui, Reda Bahi and Toth, S{\'e}bastien},
  journal={arXiv preprint arXiv:1905.09638},
  year={2019}
}

@article{parisi2019continual,
  title={Continual lifelong learning with neural networks: A review},
  author={Parisi, German I and Kemker, Ronald and Part, Jose L and Kanan, Christopher and Wermter, Stefan},
  journal={Neural networks},
  volume={113},
  pages={54--71},
  year={2019},
  publisher={Elsevier}
}

@book{ivakhnenko1965,
  title={Cybernetic Predicting Devices},
  author={Ivakhnenko, Alekse{\u\i} Grigorʹevich and Lapa, Valentin Grigor{\'e}vich},
  year={1965},
  publisher={CCM Information Corporation}
}

@book{ivakhnenko1967,
  title={Cybernetics and forecasting techniques},
  author={Ivakhnenko, Aleksey Grigorievitch and Lapa, Valentin Grigorievitch and McDonough, Robert N},
  year={1967},
  publisher={American Elsevier, NY}
}

@article{ivakhnenko1968,
  title={The group method of data handling -- a rival of the method of stochastic approximation},
  author={Ivakhnenko, Aleksey Grigorievitch},
  journal={Soviet Automatic Control},
  volume={13},
  number={3},
  pages={43--55},
  year={1968}
}

@article{ivakhnenko1971,
  title={Polynomial theory of complex systems},
  author={Ivakhnenko, Aleksey Grigorievitch},
  journal={IEEE Transactions on Systems, Man and Cybernetics},
  number={4},
  pages={364--378},
  year={1971},
  publisher={IEEE}
}

@article{Goedel:31,
author = {K. G\"{o}del},
title ={\"{U}ber formal unentscheidbare {S\"{a}tze der 
Principia Mathematica und verwandter Systeme I}}, 
journal={Monatshefte f\"{u}r Mathematik und Physik},
volume = {38},
pages={173--198},
year   = {1931}}

@article{Church:35,
author = {A. Church},
title ={An unsolvable problem of elementary number theory},
journal={Bulletin of the American Mathematical Society},
volume = {41},
pages={332-333},
year   = {1935}}

@inproceedings{lake2018generalization,
  title={Generalization without systematicity: On the compositional skills of sequence-to-sequence recurrent networks},
  author={Lake, Brenden and Baroni, Marco},
  booktitle={International conference on machine learning},
  pages={2873--2882},
  year={2018},
  organization={PMLR}
}

@article{poggio2019theoretical,
  title={Theoretical Issues in Deep Networks: Approximation},
  author={Poggio, Tomaso and Banburski, Andrzej and Liao, Qianli},
  journal={Optimization and Generalization},
  year={2019}
}

@article{ha2016hypernetworks,
  title={Hypernetworks},
  author={Ha, David and Dai, Andrew and Le, Quoc V},
  journal={arXiv preprint arXiv:1609.09106},
  year={2016}
}

@article{rusu2016progressive,
  title={Progressive neural networks},
  author={Rusu, Andrei A and Rabinowitz, Neil C and Desjardins, Guillaume and Soyer, Hubert and Kirkpatrick, James and Kavukcuoglu, Koray and Pascanu, Razvan and Hadsell, Raia},
  journal={arXiv preprint arXiv:1606.04671},
  year={2016}
}

@article{davis2022neurolisp,
  title={NeuroLISP: High-level symbolic programming with attractor neural networks},
  author={Davis, Gregory P and Katz, Garrett E and Gentili, Rodolphe J and Reggia, James A},
  journal={Neural Networks},
  volume={146},
  pages={200--219},
  year={2022},
  publisher={Elsevier}
}

@book{schmidhuber1990making,
  title={Making the world differentiable: on using self supervised fully recurrent neural networks for dynamic reinforcement learning and planning in non-stationary environments},
  author={Schmidhuber, J{\"u}rgen},
  volume={126},
  year={1990},
  publisher={Inst. f{\"u}r Informatik}
}

@article{schmidhuber92fastweights,
author = {J. Schmidhuber},
title =  { Learning to Control Fast-Weight Memories: An Alternative to
Recurrent Nets},
journal={Neural Computation},
volume = {4},
number = {1},
pages={131-139},
year   = {1992}}

@inproceedings{schmidhuber93selfref,
author = {J. Schmidhuber},
title =  {A self-referential  weight matrix},
booktitle = {Proceedings of the International Conference on
Artificial Neural Networks, Amsterdam},
pages={446-451},
publisher = {Springer},
year   = {1993}}

@inproceedings{schmidhuber93ratio,
author = {J. Schmidhuber},
title =  {On decreasing the ratio between learning complexity 
and number of time-varying variables in fully recurrent nets},
booktitle = {Proceedings of the International Conference on
Artificial Neural Networks, Amsterdam},
pages={460-463},
publisher = {Springer},
year   = {1993}}

@article{schmidhuber2015learning,
  title={On learning to think: Algorithmic information theory for novel combinations of reinforcement learning controllers and recurrent neural world models},
  author={Schmidhuber, J{\"u}rgen},
  journal={arXiv preprint arXiv:1511.09249},
  year={2015}
}

@article{schmidhuber2018one,
  title={One Big Net For Everything},
  author={Schmidhuber, J{\"u}rgen},
  journal={arXiv preprint arXiv:1802.08864},
  year={2018}
}

@article{kanervisto2025world,
  title={World and human action models towards gameplay ideation},
  author={Kanervisto, Anssi and Bignell, Dave and Wen, Linda Yilin and Grayson, Martin and Georgescu, Raluca and Valcarcel Macua, Sergio and Tan, Shan Zheng and Rashid, Tabish and Pearce, Tim and Cao, Yuhan and others},
  journal={Nature},
  volume={638},
  number={8051},
  pages={656--663},
  year={2025},
  publisher={Nature Publishing Group UK London}
}

@article{reed2022generalist,
  title={A generalist agent},
  author={Reed, Scott and Zolna, Konrad and Parisotto, Emilio and Colmenarejo, Sergio Gomez and Novikov, Alexander and Barth-Maron, Gabriel and Gimenez, Mai and Sulsky, Yury and Kay, Jackie and Springenberg, Jost Tobias and others},
  journal={arXiv preprint arXiv:2205.06175},
  year={2022}
}

@inproceedings{zhang2023adding,
  title={Adding conditional control to text-to-image diffusion models},
  author={Zhang, Lvmin and Rao, Anyi and Agrawala, Maneesh},
  booktitle={Proceedings of the IEEE/CVF international conference on computer vision},
  pages={3836--3847},
  year={2023}
}

@inproceedings{radford2021learning,
  title={Learning transferable visual models from natural language supervision},
  author={Radford, Alec and Kim, Jong Wook and Hallacy, Chris and Ramesh, Aditya and Goh, Gabriel and Agarwal, Sandhini and Sastry, Girish and Askell, Amanda and Mishkin, Pamela and Clark, Jack and others},
  booktitle={International conference on machine learning},
  pages={8748--8763},
  year={2021},
  organization={PmLR}
}

@article{turing1936computable,
  title={On computable numbers, with an application to the Entscheidungsproblem},
  author={Turing, Alan Mathison and others},
  journal={J. of Math},
  volume={58},
  number={345-363},
  pages={5},
  year={1936},
  publisher={Wiley Online Library}
}

@inproceedings{siegelmann1992computational,
  title={On the computational power of neural nets},
  author={Siegelmann, Hava T and Sontag, Eduardo D},
  booktitle={Proceedings of the fifth annual workshop on Computational learning theory},
  pages={440--449},
  year={1992}
}

@article{perez2021attention,
  title={Attention is turing-complete},
  author={P{\'e}rez, Jorge and Barcel{\'o}, Pablo and Marinkovic, Javier},
  journal={Journal of Machine Learning Research},
  volume={22},
  number={75},
  pages={1--35},
  year={2021}
}

@article{von1993first,
  title={First Draft of a Report on the EDVAC},
  author={Von Neumann, John},
  journal={IEEE Annals of the History of Computing},
  volume={15},
  number={4},
  pages={27--75},
  year={1993},
  publisher={IEEE}
}

@article{wilkes1981best,
  title={The best way to design an automatic calculating machine},
  author={Wilkes, Maurice},
  year={1981}
}

@article{queloz2025explainability,
  title={Explainability through systematicity: The hard systematicity challenge for artificial intelligence},
  author={Queloz, Matthieu},
  journal={Minds and machines},
  volume={35},
  number={3},
  pages={35},
  year={2025},
  publisher={Springer}
}

@inproceedings{calanzonelogically,
  title={Logically Consistent Language Models via Neuro-Symbolic Integration},
  author={Calanzone, Diego and Teso, Stefano and Vergari, Antonio},
  booktitle={The Thirteenth International Conference on Learning Representations},
  year={2025}
}

@article{kirkpatrick2017overcoming,
  title={Overcoming catastrophic forgetting in neural networks},
  author={Kirkpatrick, James and Pascanu, Razvan and Rabinowitz, Neil and Veness, Joel and Desjardins, Guillaume and Rusu, Andrei A and Milan, Kieran and Quan, John and Ramalho, Tiago and Grabska-Barwinska, Agnieszka and others},
  journal={Proceedings of the national academy of sciences},
  volume={114},
  number={13},
  pages={3521--3526},
  year={2017},
  publisher={National Academy of Sciences}
}

@article{adam2014method,
  title={A method for stochastic optimization},
  author={Adam, Kingma DP Ba J and others},
  journal={arXiv preprint arXiv:1412.6980},
  volume={1412},
  number={6},
  year={2014}
}

@article{wierstra2014natural,
  title={Natural evolution strategies},
  author={Wierstra, Daan and Schaul, Tom and Glasmachers, Tobias and Sun, Yi and Peters, Jan and Schmidhuber, J{\"u}rgen},
  journal={The Journal of Machine Learning Research},
  volume={15},
  number={1},
  pages={949--980},
  year={2014},
  publisher={JMLR. org}
}

@article{innes2019differentiable,
  title={A differentiable programming system to bridge machine learning and scientific computing},
  author={Innes, Mike and Edelman, Alan and Fischer, Keno and Rackauckas, Chris and Saba, Elliot and Shah, Viral B and Tebbutt, Will},
  journal={arXiv preprint arXiv:1907.07587},
  year={2019}
}

@article{fukushima1980neocognitron,
  title={Neocognitron: A self-organizing neural network model for a mechanism of pattern recognition unaffected by shift in position},
  author={Fukushima, Kunihiko},
  journal={Biological cybernetics},
  volume={36},
  number={4},
  pages={193--202},
  year={1980},
  publisher={Springer}
}

@article{vaswani2017attention,
  title={Attention is all you need},
  author={Vaswani, Ashish and Shazeer, Noam and Parmar, Niki and Uszkoreit, Jakob and Jones, Llion and Gomez, Aidan N and Kaiser, {\L}ukasz and Polosukhin, Illia},
  journal={Advances in neural information processing systems},
  volume={30},
  year={2017}
}

@article{wiedemer2025video,
  title={Video models are zero-shot learners and reasoners},
  author={Wiedemer, Thadd{\"a}us and Li, Yuxuan and Vicol, Paul and Gu, Shixiang Shane and Matarese, Nick and Swersky, Kevin and Kim, Been and Jaini, Priyank and Geirhos, Robert},
  journal={arXiv preprint arXiv:2509.20328},
  year={2025}
}
